\documentclass[accepted]{uai2021} 

\usepackage[american]{babel}
\usepackage{amssymb,amsmath,bm}
\usepackage{float}
\newtheorem{prop}{Proposition}
\newtheorem{corollary}{Corollary}

\usepackage{bbm}

\usepackage{subfig}
\DeclareMathAlphabet{\mathcal}{OMS}{cmsy}{m}{n}

\usepackage[]{color}

\usepackage{natbib} 
\usepackage{mathtools} 
\usepackage{booktabs} 
\usepackage{tikz} 
\DeclareUnicodeCharacter{2212}{-} 

\title{Dimension Reduction for Data with Heterogeneous Missingness}

\author[1]{\href{mailto:Yurong Ling <yurong.ling.16@ucl.ac.uk>?Subject=Your UAI 2021 paper}{Yurong Ling}{}} 
\author[2]{Zijing Liu}
\author[1]{Jing-Hao Xue}
\affil[1]{%
    Department of Statistical Science\\
    University College London\\
    London, UK
}
\affil[2]{%
    Department of Brain Sciences\\
    Imperial College London\\
    London, UK
}
  
\begin{document}
\maketitle

\begin{abstract}
  Dimension reduction plays a pivotal role in analysing high-dimensional data. However, observations with missing values present serious difficulties in directly applying standard dimension reduction techniques. As a large number of dimension reduction approaches are based on the Gram matrix, we first investigate the effects of missingness on dimension reduction by studying the statistical properties of the Gram matrix with or without missingness, and then we present a bias-corrected Gram matrix with nice statistical properties under heterogeneous missingness. Extensive empirical results, on both simulated and publicly available real datasets, show that the proposed unbiased Gram matrix can significantly improve a broad spectrum of representative dimension reduction approaches.
\end{abstract}

\section{Introduction}\label{sec:intro}
Dimension reduction (DR) is important for analysing high-dimensional data as it helps reveal underlying structures of the data. A large number of DR methods have been successfully applied to real data, such as principal components analysis (PCA)~\citep{pca}, dual probabilistic PCA and its non-linear variant Gaussian process latent variable model (GPLVM)~\citep{gplvm}. However, missing data arise in many applications~\citep{cfmissing,remotesensing,genomics}, making it infeasible standard DR methods, which are usually designed for complete data. To address the problems posed by missing data, a broad spectrum of methods have been proposed, including the expectation-maximisation approaches~\citep{little}, the direct imputation of observations via either matrix completion~\citep{matrix1,matrix2,softimpute} or by chained equations~\citep{ice1}, and the implicit imputation of covariance matrix~\citep{svdpca,primePCA}. 

In this work, we focus on the implicit imputation of the Gram matrix and show that an unbiased estimator of it, in the presence of missing data, offers a significant prospect for enhancing the reliability of many DR procedures. Specifically, a large number of widely used DR methods obtain the low-dimensional projections via the distance matrix or the Gram matrix rather than the data matrix. 
For example, multidimensional scaling (MDS) seeks to find an embedded low-dimensional structure, of which the distance matrix is as close to the high-dimensional distance matrix as possible~\citep{mds}. In addition, the objective of preserving the distance relationship between data points is shared by many dimension reduction algorithms. Algorithms such as the t-distributed stochastic neighbor embedding (tSNE)~\citep{tsne} and the uniform manifold approximation and projection (UMAP)~\citep{umap}, two favoured visualisation tools in data analysis, build the stochastic relationship between data points in the low-dimensional space based on their original Euclidean distances. Similarly, dual probabilistic PCA and its non-linear variant GPLVM also seek to find the low-dimensional embedding using the Gram matrix~\citep{gplvm}. It is adequate for performing the aforementioned methods through precise calculation of either the Gram matrix or the distance matrix, due to the linear transformations between these two matrices: the Gram matrix can be obtained by doubly centering the squared Euclidean distance matrix~\citep{van2009dimensionality}, while there also exists a linear transformation for converting the Gram matrix to the distance matrix, as shown by~\eqref{eq:gtoe}). Consequently, for the relevant DR approaches, we do not need to impute the missing values as long as we can estimate the distance or Gram matrix reliably in such cases. 

Although \citet{svdpca} and \citet{primePCA} studied the eigenvectors and eigenvalues for homogeneous and heterogeneous missing data, respectively, the effect of missing data on the techniques beyond PCA remains unclear. Moreover, the consequences of neglecting missing observations are not thoroughly studied from a statistical perspective. Therefore, this paper aims to fill in this critical gap by making the following contributions. First, we elucidate how a reliable Gram matrix can ensure the powerful representation using GPLVM, a generalised framework for DR where the Gram matrix can be seen as an estimator of the covariance matrix~\citep{gplvm} (sec.\ref{ss:method:gplvm}). Secondly, we show that, owing to missing data, the original computation of the Gram matrix by an inner product matrix is a biased estimator of the covariance matrix with a larger variance under the framework of GPLVM, and we propose unbiased estimators in the cases of homogeneous missingness (sec.\ref{ss:method:homo}) and heterogeneous mechanism (sec.\ref{ss:method:hetero}). In addition, we clarify the role of input dimension in the relevant dimension reduction methods, based on its relationship with the variances of the estimators (sec.\ref{ss:method:dim}).

The data illustrated in this paper include image data and single-cell RNA sequencing (scRNA-seq) data, which measure gene expression at a single-cell level and offer a way to investigate the stochastic heterogeneity of complex issues on a near-genome-wide scale~\citep{gku555,shapiro2013single,molecularcell}.  The comparison is conducted in two aspects: visualisation and clustering results, on both simulated and real datasets (sec.\ref{ss:results:vis} and sec.\ref{ss:results:clustering}). Moreover, we empirically verify that the impact of input dimension is consistent to the results from our theoretical analysis (sec.\ref{ss:results:dim}). 

\section{Proposed unbiased estimators}\label{s:method}

In this section, we first show that the Gram matrix of high-dimensional data is an unbiased estimator of a covariance matrix when there is no missing observation and we clarify the importance of accurately computing the Gram matrix under the framework of GPLVM. The missing data model is then introduced and leads to bias in the estimator, and an unbiased estimator is derived in the presence of missing observations. Finally, we elucidate the role of input dimension on DR.

\subsection{For complete data}
\label{ss:method:gplvm}

Consider a dataset of $N$ observations and $D$ features represented as an $N\times D$ matrix $Y = [\mathbf{y}_1,\ldots,\mathbf{y}_N]^T $, where $\mathbf{y}_i \in \mathbb{R}^D$ is a $D$-dimensional observation. Under the assumption of GPLVM, every dimension is a realisation of a Gaussian process (GP) indexed by the latent variables $X = [\mathbf{x}_1,\ldots,\mathbf{x}_N]^T $, where $\mathbf{x_i} \in \mathbb{R}^d$ and $d$ is the dimension of the latent space (normally $d\ll D$). Let the GP have a mean function $m(\mathbf{x})$ and a covariance function $k(\mathbf{x},\mathbf{x}')$.
For simplicity, $m(\mathbf{x})$ is taken to be the zero function ($m(\mathbf{x})=0$). Let $\mathbf{y}_{:,i}$ denote the $i$-th column of the data matrix $Y$, GPLVM assumes that $\mathbf{y}_{:,i} \sim \mathcal{N}(\mathbf{0},K)$, where $K$ is the covariance matrix with $K_{ij} = k(\mathbf{x}_i,\mathbf{x}_j)$. GPLVM then aims to find the latent variables by maximising the marginal likelihood of the data:
\begin{equation}\label{eq:gplvm}
    \mathrm{log}~p({Y}|{X},\theta) = \sum_{s=1}^D \log~p(\mathbf{y}_{:,s}| {X},\theta),
\end{equation}
where 
\begin{equation*}
     \mathrm{log}~ p(\mathbf{y}_{:,s}|{X},\theta) = -\frac{1}{2} \mathbf{y}_{:,s}^T {K^{-1}} \mathbf{y}_{:,s}- \frac{N}{2} \mathrm{log}2\pi - \frac{1}{2} \mathrm{log} \left| {K} \right|.
\end{equation*}
The above formulation provides a probabilistic interpretation of dual PCA in the case of the linear covariance function $k(\mathbf{x}_{i},\mathbf{x}_{j}) =\mathbf{x}_{i}^T\mathbf{x}_{j}$. Notably, GPLVM would be a non-linear model as long as $K$ is obtained by a non-linear covariance function. Here we denote the latent points found by GPLVM by $\hat{X}$ and denote the corresponding covariance matrix by $\hat{K}$ to differentiate them from the true latent points $X$ and true covariance matrix $K$, respectively. 

The following Kullback-Leibler (KL) divergence, between the two Gaussians~\citep{kullback} equivalent to~\eqref{eq:gplvm} up to a constant independent of $X$, clarifies the objective of GPLVM:
 \begin{equation}
   \begin{split}
     &\mathbf{KL}(\mathcal{N}(z \mid \mathbf{0}, \frac{1}{D}G) \mid \mid \mathcal{N}(z \mid \mathbf{0}, K )
       = \frac{1}{2} \mathrm{log}\left | K \right | -\\ &\frac{1}{2}\mathrm{log}\left | \frac{1}{D}G \right | 
       + \frac{1}{2}\mathrm{tr}(\frac{1}{D}GK^{-1}) - \frac{N}{2},
   \end{split}
 \end{equation}
where $G = YY^{T}$ is the Gram matrix. Thus, GPLVM seeks a matrix of latent points $\hat{X}$, which generate the covariance matrix $\hat{K}$, with $\hat{K}_{ij} = k(\mathbf{\hat{x}}_{i}, \mathbf{\hat{x}}_{j})$, as close to $\frac{1}{D}G$ as possible in terms of KL divergence.
Moreover, $\frac{1}{D}G$ can be regarded as an estimator of the covariance matrix in $\mathcal{N}(\mathbf{0}, K)$.
Since $G_{ij} = \sum_{s=1}^D y_{is}y_{js}$, under the assumption that each column of $Y$ follows $\mathcal{N}(\mathbf{0},K)$, the asymptotic properties of $\frac{1}{D}G$ can be summarised by the Lindeberg–Lévy central limit theorem (CLT) as~\citep{elements}
\begin{equation}\label{eq:gplvm_asym}
   \begin{aligned}
   \sqrt{D}( \frac{{G}_{ij}}{D}  - K_{ij}) 
    &\overset{dist.}{\rightarrow}
    \mathcal{N}(\mathbf{0} ,{K}_{ii} {K}_{jj} + {K}_{ij}^2), \,i\neq j, \\
     \sqrt{D}(\frac{{G}_{ii}}{D}  - K_{ii})
    &\overset{dist.}{\rightarrow}
    \mathcal{N}(\mathbf{0} ,2K^2_{ii}),i= j, \text{ as } D \rightarrow \infty.
    \end{aligned}
\end{equation}
That is, when there is no event of missingness, not only $\frac{1}{D}G$ is an unbiased estimator of the covariance matrix $K$, but also,
with a higher dimension $D$, $\frac{1}{D}G$ is an estimator of higher accuracy because the variance shrinks with $D$.

\subsection{For data with homogeneous missingness}\label{ss:method:homo}

The nice asymptotic properties in~\eqref{eq:gplvm_asym} may not hold as missing data exist. It is hence necessary to investigate the consequences of directly using the Gram matrix obtained from data with missing values. Let $\tilde{Y}$ denote the complete data matrix without missing entries and $Y$ the partially observed data matrix. We define a $N\times D$ binary revelation matrix $\Omega$ with $1$ representing the corresponding entry in $\tilde{Y}$ being observed and $0$ for the missing entry. Here, we assign a value of $0$ to missing entries, as \citet{svdpca,primePCA} did in their work, to calculate the Gram matrix with missing observations by $YY^{T}$. Therefore, we have $Y = \tilde{Y} \circ \Omega$, where $\circ$ denotes the element-wise product.

We first study a simple case where values in a dataset are missing independently and completely at random (MCAR) with the homogeneous probability. Under the framework of GPLVM, we assume that the partially observed data matrix $Y$ is generated column-wise from $\mathcal{N}(\mathbf{0},K)$, followed by the event of homogeneous missingness. The occurrences of missing observations are assumed to follow independent Bernoulli distributions with the homogeneous probability $1-p ~(0<p\leq1)$. We also assume that the missing observations are independent of the Gaussian processes given $p$. The detailed statistical model is
\begin{equation}
    \begin{aligned}
    \mathbf{\tilde{y}}_{:,s} &\sim \mathcal{N}(\mathbf{0},K), \\
    h_{is} \mid \tilde{y}_{is} &\sim \mathrm{Bernoulli}(p), \\
    y_{is} & =
    \left\{
        \begin{array}{cc}
                \tilde{y}_{is} & \mathrm{if\ } h_{is}=1, \\
                0  & \mathrm{if\ } h_{is}=0,  \\
        \end{array} 
    \right. 
    \end{aligned}
\end{equation}
where $\tilde{y}_{is}$ is the true value in the $i$-th observation and $s$-th feature $(s=1,\ldots,D; \ i=1,\ldots,N)$. 

From the above model, we get $\mathbb{E}({y}_{is}{y}_{js}) = p^2{K}_{ij}$ and $\mathrm{Var}({y}_{is}{y}_{js}) =  p^2{K}_{ii}{K}_{jj}+{K}_{ij}^2(2p^2-p^4)$ for $i\neq j$; and $\mathbb{E}({y}_{is}^2) =p{K}_{ii}$ and $\mathrm{Var}({y}_{is}^2) ={K}_{ii}^2(3p-p^2) $. 
With the CLT, the asymptotic properties of $\frac{1}{D}G$, as $D \rightarrow \infty$, are
\begin{equation}\label{eq:gplcm_asym}
   \begin{aligned}
    &\sqrt{D}( \frac{{G}_{ij}}{D}  - p^2K_{ij})
    \overset{dist.}{\rightarrow}
    \mathcal{N}(\mathbf{0},p^2{K}_{ii}{K}_{jj}+{K}_{ij}^2(2p^2-p^4) ), \\
    &i\neq j; \\
    &\sqrt{D}(\frac{{G}_{ii}}{D}  - p K_{ii})
    \overset{dist.}{\rightarrow}
    \mathcal{N}(\mathbf{0},{K}_{ii}^2(3p-p^2)),i= j.
    \end{aligned}
\end{equation}
Therefore, $\frac{1}{D}G$ is no longer an unbiased estimator of $K$ for data with missing values.
As we mentioned in sec.\ref{ss:method:gplvm}, the Gram matrix is of importance to dual PCA and GPLVM.
Hence the latent points found by GPLVM with the original Gram matrix $G$ can be misleading. A straightforward solution to the problem is to use an unbiased estimator instead of $\frac{1}{D}G$.
Based on the above analysis in~\eqref{eq:gplcm_asym}, it is easy to see that an unbiased estimator for $K$ is a matrix $\tilde{G}$ where
\[
                \tilde{G}_{ij}= \frac{G_{ij}}{Dp^2}\, ( i\neq j)
\,\mathrm{and}\, \tilde{G}_{ii}=\frac{G_{ii}}{Dp}.
\]

\subsection{For data with heterogeneous missingness}\label{ss:method:hetero}

In spite of the simplicity of the assumption presented in sec.\ref{ss:method:homo}, the homogeneous missing data model conflicts with the fact that the missingness probability is often linked to some internal or external factors in reality. For instance, there usually exists an inverse relationship between the true values and the corresponding missingness probabilities in scRNA-seq data~\citep{zifa}; in the recommendation system, whether a user rates a movie is determined by their preference and the movie's genre. We therefore consider the case where the missingness probabilities are heterogeneous; that is, values are missing not at random (MNAR). Further, we propose an unbiased estimator in such a situation. Now the statistical model is
\begin{equation}\label{eq:model}
    \begin{aligned}
    \mathbf{\tilde{y}}_{:,s} &\sim \mathcal{N}(\mathbf{0},K), \\
    h_{is} \mid \tilde{y}_{is} &\sim \mathrm{Bernoulli}(p_{is}), \\
    y_{is} &=
    \left\{
        \begin{array}{cc}
                \tilde{y}_{is} & \mathrm{if\ } h_{is}=1, \\
                0  & \mathrm{if\ } h_{is}=0, \\
        \end{array} 
    \right.
    \end{aligned}
\end{equation}
where $1-p_{is}$ denotes the probability of missingness for the $i$-th observation and $s$-th feature, and $0<p_{is} \leq  1$ ($ s=1,\ldots,D; \ i=1,\ldots,N$). 

By using the statistical model in~\eqref{eq:model}, we get the following two propositions regarding the estimator of $K$ in such case.
\begin{prop}\label{prop1}
Let the probabilities of missingness for the $s$-th feature in the $i$-th and $j$-th observations to be $1-p_{is}$ and $1-p_{js}$ respectively, where $i=1,\ldots,N,\, s=1,\ldots,D.$ By assuming that the observed data matrix $Y$ is generated according to the model in~\eqref{eq:model}, we have, for $i \neq j$,  
\begin{equation}
    \begin{aligned}
    \mathbb{E}[{y}_{is}{y}_{js}] & = p_{is}p_{js}{K}_{ij}, \\
    \mathrm{Var}[{y}_{is}{y}_{js}] & = p_{is}p_{js}{K}_{ii}{K}_{jj} +{K}_{ij}^2(2p_{is}p_{js}-p_{is}^2p_{js}^2);
    \end{aligned}
\end{equation}
and for $i  = j$,
 \begin{equation}
    \begin{aligned}
    \mathbb{E}[{y}_{is}{y}_{js}] & = \mathbb{E}[{y}_{is}^{2}] = p_{is}{K}_{ii}, \\
    \mathrm{Var}[{y}_{is}{y}_{js}] & =\mathrm{Var}[{y}_{is}^{2}] = {K}_{ii}^2(3p_{is} - p_{is}^2). 
    \end{aligned}
\end{equation}
\end{prop}

Based on Proposition~\ref{prop1}, it is straightforward to conclude that $\frac{1}{D}G$ is a biased estimator of $K$. Consequently, it is necessary to correct the bias so as to get reliable $\hat{K}$ and $\hat{X}$. 

\begin{prop}\label{prop2}
By adopting the same assumption and notation as those in Proposition~\ref{prop1}, we obtain an unbiased estimator $\tilde{G}$ of $K$ with bounded variances. Specifically, for $i \neq j$ we have $\tilde{G}_{ij} = \frac{{G}_{ij}}{\sum_{s=1}^D p_{is}p_{js} }$, and $\mathrm{Var}[\tilde{G}_{ij}]$ is given by 
\begin{equation}
       \frac{ {K}_{ii}{K}_{jj}\sum_{s=1}^D p_{is}p_{js} + {K}_{ij}^2 \sum_{s=1}^D (2p_{is}p_{js}-p_{is}^2p_{js}^2)}
     {\left( \sum_{s=1}^D p_{is}p_{js} \right)^2}.
\end{equation}
The bounds of $\mathrm{Var}[\tilde{G}_{ij}]$ are given by
\begin{equation}
    \frac{{K}_{ii}{K}_{jj}+{K}_{ij}^2}{D \bar{p}_{ij}}\leq \mathrm{Var}(\tilde{G}_{ij}) \leq \frac{{K}_{ii}{K}_{jj}}{\bar{p}_{ij} D} +
     \frac{{K}_{ij}^2}{D}\left( \frac{2}{\bar{p}_{ij}} - 1\right),
\end{equation}
where $0<\bar{p}_{ij} = \frac{1}{D}\sum_{s=1}^D p_{is}p_{js} \leq 1$. Note that the equality holds if and only if $p_{is}=1, \text{for all } i \text{ and } s$, which means no event of missing observations.

For the diagonal entries, $\tilde{G}_{ii} =\frac{{G}_{ii}}{\sum_{s=1}^D p_{is} }$, and $\mathrm{Var}[\tilde{G}_{ii}]$ is
\begin{equation}
       \frac{ {K}_{ii}^2 \sum_{s=1}^D p_{is}(3 - p_{is})}
     {\left( \sum_{s=1}^D p_{is} \right)^2}.
\end{equation}
The bounds of $\mathrm{Var}[\tilde{G}_{ii}]$ are given by
\begin{equation}
    \frac{2K_{ii}^2}{D \bar{p}_{i}}\leq \mathrm{Var}(\tilde{G}_{ii}) \leq \frac{{K}_{ii}^2}{D}\left( \frac{3}{\bar{p}_{i}} - 1\right),
\end{equation}
where $ \bar{p}_{i} = \frac{1}{D} \sum_{s=1}^D p_{is}$. Again, the equality holds if and only if $p_{is}=1, \, \text{for all } i \text{ and } s.$
\end{prop}

Based on Proposition~\ref{prop2}, we conclude that $\tilde{G}$ is bias-corrected with the bounds of variance decreasing with $D$. Furthermore, under a mild condition, $\tilde{G}$ is a consistent estimator.
\begin{prop}\label{prop3}
Let $x_{ij, s} = y_{is}y_{js}$ and $\mu_{ij,s} = \mathbb{E}(x_{ij,s}) = p_{is}p_{js}K_{ij}$. 
If $\sum\limits_{s=1}^D p_{is}p_{js}\asymp D$, then 
\begin{equation}
Z_{ij,D} = \frac{\sum_{s=1}^{D} \left( x_{ij,s} - \mu_{ij,s}\right)}{ \left[\sum_{s=1}^{D}  \mathrm{Var} (x_{ij,s})\right]^{\frac{1}{2}}}
\overset{dist.}{\rightarrow}
    \mathcal{N} (0,1),
 \end{equation} 
as $D$ approaches infinity. Here we define $\sum\limits_{s=1}^D p_{is}p_{js}\asymp D$ if there exist constants $0<m<M<\infty$, and an integer $n_0$ such that $m<\frac{\sum_{s=1}^D p_{is}p_{js}}{D} <M$, for all $D>n_0$.
\end{prop}
\begin{corollary}\label{collary1}
If $\sum\limits_{s=1}^D p_{is}p_{js}\asymp D$, 
the proposed unbiased estimator $\tilde{G}$ converges in probability to the true covariance matrix $K$ as $D$ approaches infinity.
\end{corollary}

Proposition~\ref{prop3} and Corollary~\ref{collary1} suggest that the unbiased estimator $\tilde{G}$ of $K$ could be beneficial for a method using the Gram matrix as input since it converges to the ground-truth covariance matrix in the presence of missing observations if $\sum\limits_{s=1}^D p_{is}p_{js}\asymp D$. 
In reality, $\frac{\sum_{s=1}^D p_{is}p_{js}}{D} <M$ for any constant $M>1$, since $p_{ij}$'s $\leq 1$. Furthermore, there exists a constant $m>0$ such that $m < \frac{\sum_{s=1}^D p_{is}p_{js}}{D}$ as long as the probabilities of non-missingness $p_{ij}$'s are bounded from below. Thus, the condition $\sum\limits_{s=1}^D p_{is}p_{js}\asymp D$ is readily satisfied in practice.
When applying the proposed estimator to DR methods, we use $\tilde{G}$ rather than $\frac{1}{D}G$ to improve the performance. 

\begin{corollary}\label{collary2}
If $\sum\limits_{s=1}^D p_{is}p_{js}\asymp D$, 
the estimator $G/D$ converges in probability to the true covariance matrix $K$ if and only if $\lim\limits_{D \rightarrow \infty} \frac{\sum\limits_{s=1}^{D} p_{is}p_{js}}{D} = 1$.
\end{corollary}
Corollary~\ref{collary2} implies that the estimator $G/D$ would still converge to $K$ if the fraction of missing values is small enough as compared to $1$. All proofs in this section are provided in the supplementary material. Note that the mathematical terms in the bias and the variances presented in Proposition~\ref{prop1} and Proposition~\ref{prop2}, respectively, would be more involved if we set the missing observations to a non-zero constant, but the statistical properties remain the same. 


\subsection{Impact of the input dimension $D$}\label{ss:method:dim}
As shown in~\eqref{eq:gplvm_asym}, \eqref{eq:gplcm_asym} and Proposition~\ref{prop2}, the variance of $\frac{1}{D}G$ and the bounds of $\mathrm{Var}(\tilde{G})$ are inversely proportional to the input dimension $D$. 
Hence, higher input dimension can lead to more accurate results of dimension reduction, from decreasing the variances of the estimators.
Moreover, $\frac{1}{D}G$ would be close enough to $K$ when there exists no missing entries in $Y$, as long as the dimension is high enough such that the corresponding variances approach zero, so is $\tilde{G}$ for data with missingness. 
Although $\tilde{G}$ is an unbiased estimator, as shown by Proposition~\ref{prop2}, the variance of $\tilde{G}$ are greater than that of $\frac{1}{D}G$ in the presence of missing observations. 
In other words, in order to reach the same accuracy, more dimensions are required in the presence of missing data, compared with the case of complete data.

\section{Application of the proposed unbiased estimator}\label{s:algorithm}
In practice, the heterogeneous probabilities of missingness are unknown. Hence, we need to estimate them before applying the proposed estimator to real datasets. The procedure of estimation is proposed as follows: first compute $\mathbf{p_{F}} \in \mathbb{R}^D$ and $\mathbf{p_{S}} \in \mathbb{R}^N$, which are the vectors containing the proportion of non-missing observations for each feature and for each sample, respectively; then, the entries in the outer product of two vectors $\mathbf{p_{S}} \otimes \mathbf{p_{F}}$ scaled by a constant are treated as the matrix of estimated non-missingness probabilities for the data matrix. The detail of estimators is provided in sec.\ref{secs:dropout_prob} of the supplementary material. 

Once all the $p_{ij}$ are estimated, we compute $\tilde{G}$ and then substitute $\frac{G}{D}$ in the relevant Gram-matrix-based dimension reduction methods, such as PCA and GPLVM, to correct the bias. In addition, $\tilde{G}$ can benefit the approaches designed taking advantage of the distance structure, owing to the linear transformation between the squared Euclidean distance matrix $E^2$ and $G$: 
\begin{equation}\label{eq:gtoe}
    E^2 = \text{diag}(G)\mathbf{1}^T + \mathbf{1}\text{diag}(G)^T - 2G, 
\end{equation}
where $\text{diag} (G)$ is a column vector of the diagonal elements in $G$. Considering that the bias corrected $\tilde{G}$ could result in negative values via~\eqref{eq:gtoe}, we propose an alternative way to enhance the distance-matrix-based methods such as tSNE and UMAP: first do PCA with the bias-corrected Gram matrix $\tilde{G}$, and then calculate the distance matrix in the PC space. To ensure a good estimate of the distance, we keep all the PCs with non-negative eigenvalues.

\begin{figure}[!t]
 \centering
 \includegraphics[width=1\linewidth]{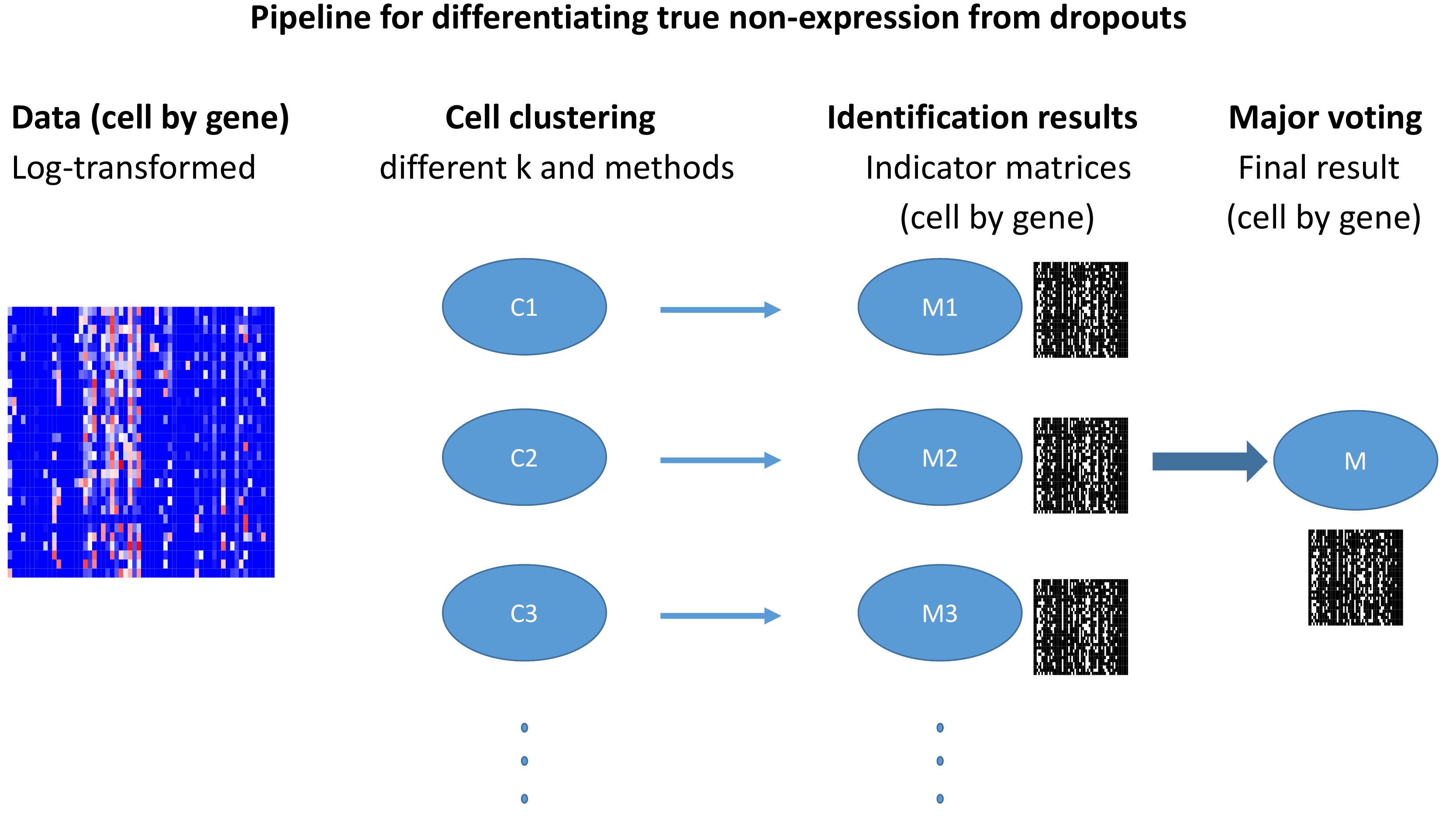}
  \caption{Pipeline for identifying dropout events. From left to right: 1) log-transformed data matrix; 2) clustering with different clustering methods and different numbers of clusters ($k$); 3) indicator matrices from different clustering results; and 4) combining all indicator matrices (results) by the majority voting.}
  \label{fig:pipeline}
\end{figure}

There is an additional challenge when handling real scRNA-seq data: the positions of missing entries remain unknown. There exist highly-frequent zero expression values in scRNA-seq data. Some zeros indicate the true biological non-expression while others are due to the corresponding missing values, which are called dropouts~\citep{genomics,scimpute}. To address the mentioned problem, we propose a simple yet reliable ensemble-learning strategy to infer the positions of missing entries (dropouts) in data matrix, as illustrated in Figure~\ref{fig:pipeline}. Specifically, we first identify similar cells via clustering. A zero count is then regarded as true biological non-expression if most values of the same gene in the corresponding cluster are zero, otherwise it is taken to be a missing value. This identification procedure is performed multiple times using different clustering methods and different numbers of clusters to ensure reliable results. We reach the final decision by the majority voting: a zero count is considered as the true non-expression if more than half results confirm this. The proposal of this procedure is inspired by the principle introduced in scImpute that a zero count may reflect real biological variability if the corresponding gene has constantly low expression in similar cells~\citep{scimpute}. After the true non-expression or dropout events are identified, we compute the probability of being a dropout across both observations (cells) and features (genes) as we mentioned before. 

\section{Experimental results and analysis}\label{s:results}

In this section, we show the superiority of the proposed unbiased estimator in terms of the clustering accuracy and visualisation quality on both simulated and real datasets. 
The results in sec.\ref{ss:method:dim} regarding the role of the input dimension are empirically verified with the simulated data. The code to reproduce these experiments is available at \url{https://github.com/yurongling/DR-for-Data-with-Missingness}

\begin{table}[t]
\caption{Real datasets used in this paper.\label{tb:data}}
\resizebox{\linewidth}{!}{
\begin{tabular}{lllll}
\hline
Dataset  & \# clusters/classes & N & D & Ref\\
\hline
Pollen (scRNA-seq) & 11 & 301 &21045 & \citep{pollen} \\
Deng (scRNA-seq)& 10  & 286   &20484 & \citep{Deng}\\
Treutlein (scRNA-seq)   & 8  & 405 &15893  & \citep{Treutlein} \\
Koh (scRNA-seq)  & 10 & 651 &41594 & \citep{Koh}  \\
Usoskin (scRNA-seq) & 4 & 622 &17571 & \citep{usoskin}                \\
Kumar (scRNA-seq) & 3 & 361 &16092 & \citep{Kumar}  \\
Olivetti faces (image) & 40 & 400 &4096 & \citep{olivetti} \\
fashion MNIST (image) & 10 & 1000 &784 &\citep{MNIST} \\
wine (UCI) & 3 & 178 &13 & \citep{UCI} \\
\hline
\end{tabular}}
\end{table}

\subsection{Datasets}
Nine publicly available real datasets from different domains are selected for comparing different methods: $6$ scRNA-seq datasets, $2$ image datasets, and $1$ dataset from the UCI repository. The characteristics of each dataset are provided in Table~\ref{tb:data}; the data pre-processing and availability are provided in sec.\ref{secs:data_preprocessing} and sec.\ref{secs:download} of the supplementary material, respectively. Clusters in each real scRNA-seq dataset are for different cell types. We sample $1000$ images from the test set of the fashion MNIST dataset for comparison by preserving the percentage of samples of each class to reduce the computational complexity of some benchmark imputation methods. Note that the wine dataset possesses only a small number of features ($13$).

In addition to real datasets, we also simulate a dataset with $3$ clusters for investigating the role of the input dimension. The complete dataset is first simulated with the Probabilistic PCA (PPCA)~\citep{ppca}, which can be regarded as a GPLVM with a linear kernel. Then, the data matrix with missing observations is generated with a missingness mechanism mentioned below.

\textbf{Missing value generation mechanism.} Apart from scRNA-seq datasets, all datasets we employ are complete. We therefore adopt a missingness mechanism to generate missing positions. Specifically, P($\Omega_{ij}=0) = b_i q_j$, for $i \in [N]$, $j \in [D]$, where iid $b_1,\ldots,b_N \sim U[0.4,0.6]$,  and iid $q_1,\ldots,q_D \sim U[0.7,0.9]$. The fraction of missingness is around $0.4$.

\begin{figure}[t]
    \centering
    \includegraphics[width=1\linewidth]{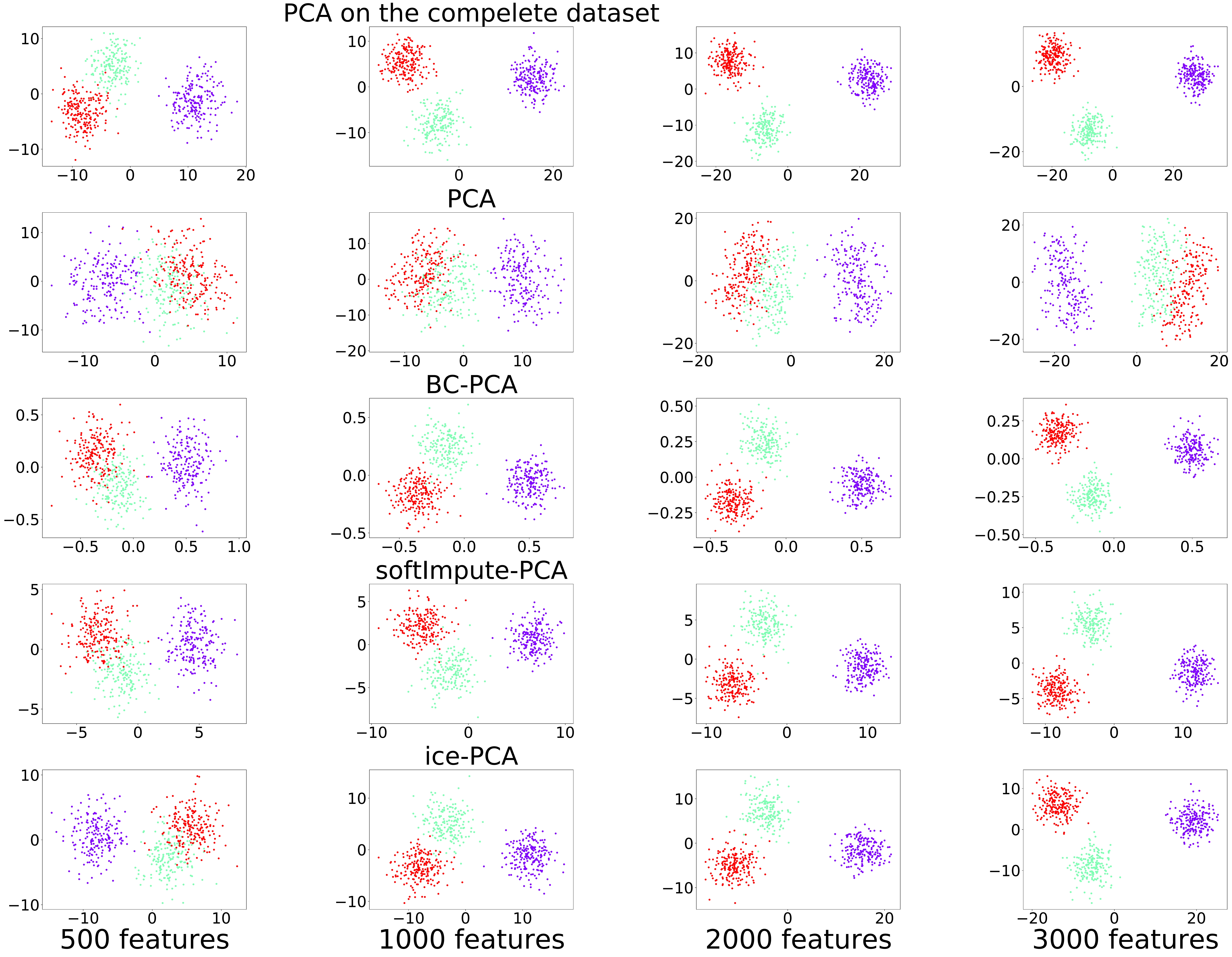}
    \caption{2D scatter plots of the simulated data with or without missing observations, obtained by four different methods (from top to bottom: PCA on the complete dataset, PCA, BC-PCA, softImpute-PCA, and ice-PCA on the dataset with missing values), and with different numbers of genes (from left to right: 500, 1000, 2000 and all 3000 features). Colours indicate the cluster labels. Comparing columns: generally more input features, better separation between different clusters. Comparing rows: BC-PCA, softImpute-PCA, and ice-PCA lead to much more distinct clusters on the dataset with heterogeneous missingness.}
    \label{fig:sim_vis} 
\end{figure}

\subsection{Benchmarks}
\textbf{DR methods.} To demonstrate the applicability and the effectiveness of the proposed estimator, we consider four DR methods: PCA, GPLVM, tSNE, and UMAP. PCA and GPLVM are representative Gram-matrix-based methods, while tSNE and UMAP are widely-used approaches depending on the distance matrix. In both the simulated and real experiments, we compare them with their bias-corrected variants proposed in this paper, where the Gram matrix is replaced by $\tilde{G}$ or the distance matrix is calculated in the PCA space obtained from $\tilde{G}$ as discussed in sec.\ref{s:algorithm}. The prefix \textit{BC-} (bias-corrected) of each method is to denote its bias-corrected variant. 

\textbf{Imputation methods.} The widely-used imputation methods softImpute~\citep{softimpute} and imputation by chained equations (ice)~\citep{ice1} are applied to the datasets, followed by performing the DR methods on the imputed data matrices. 
softImpute is proposed with the low-rank assumption and performs missing values imputation using iterative soft-thresholded SVD’s, while ice uses a strategy that models each feature with missing values as a function of other features in a round-robin fashion. 
We use the prefix \textit{softImpute-}/\textit{soft-} and \textit{ice-} to represent the corresponding DR methods applied to the data matrix imputed by softImpute and ice, respectively. When these two imputation approaches are applied to real scRNA-seq data where the missing positions are unidentified, we input the missing positions inferred by the proposed pipeline. 

\textbf{Approaches designed specifically for scRNA-seq data} For real scRNA-seq data, we also integrate two approaches that deal with the missing data problem of scRNA-seq data: CIDR~\citep{cidr} and scImpute~\citep{scimpute}. CIDR tries to recover the Euclidean distance matrix while scImpute aims at imputing the missing values. The imputed data matrix produced by scImpute is directly fed into the mentioned four benchmarks for extracting low-dimensional components. tSNE and UMAP take the imputed Euclidean distance matrix yielded by CIDR as input. On the other hand, we transform the imputed distance matrix to the Gram matrix by doubly-centering, which is then input into GPLVM and PCA, respectively. We use the prefix \textit{CIDR-} and \textit{sc-}/\textit{scImpute-} to denote the corresponding DR methods integrated with CIDR and scImpute, respectively,

\textbf{Evaluation.} We evaluate the performances from two perspectives: clustering and visualisation. For clustering-based evaluation, we use $k$-means clustering in the reduced space and the number of clusters is set to the same as the ground truth.
The clustering results are evaluated in terms of the adjusted rand index (ARI) between the cluster/class labels obtained from the original publication and the inferred clustering labels. Since the missing positions determined by the proposed pipeline could be variable, we replicate the procedure of first performing DR and then applying $k$-means $20$ times on the real scRNA-seq datasets for a more reliable comparison.
Regarding the visualisation-based evaluation, we reduce the input data into two dimensions and visually compare the visualisations. For implementation details, see sec.\ref{secs:implementation} of the supplementary material.

\begin{figure}[t]
    \centering
    \includegraphics[width=0.7\linewidth]{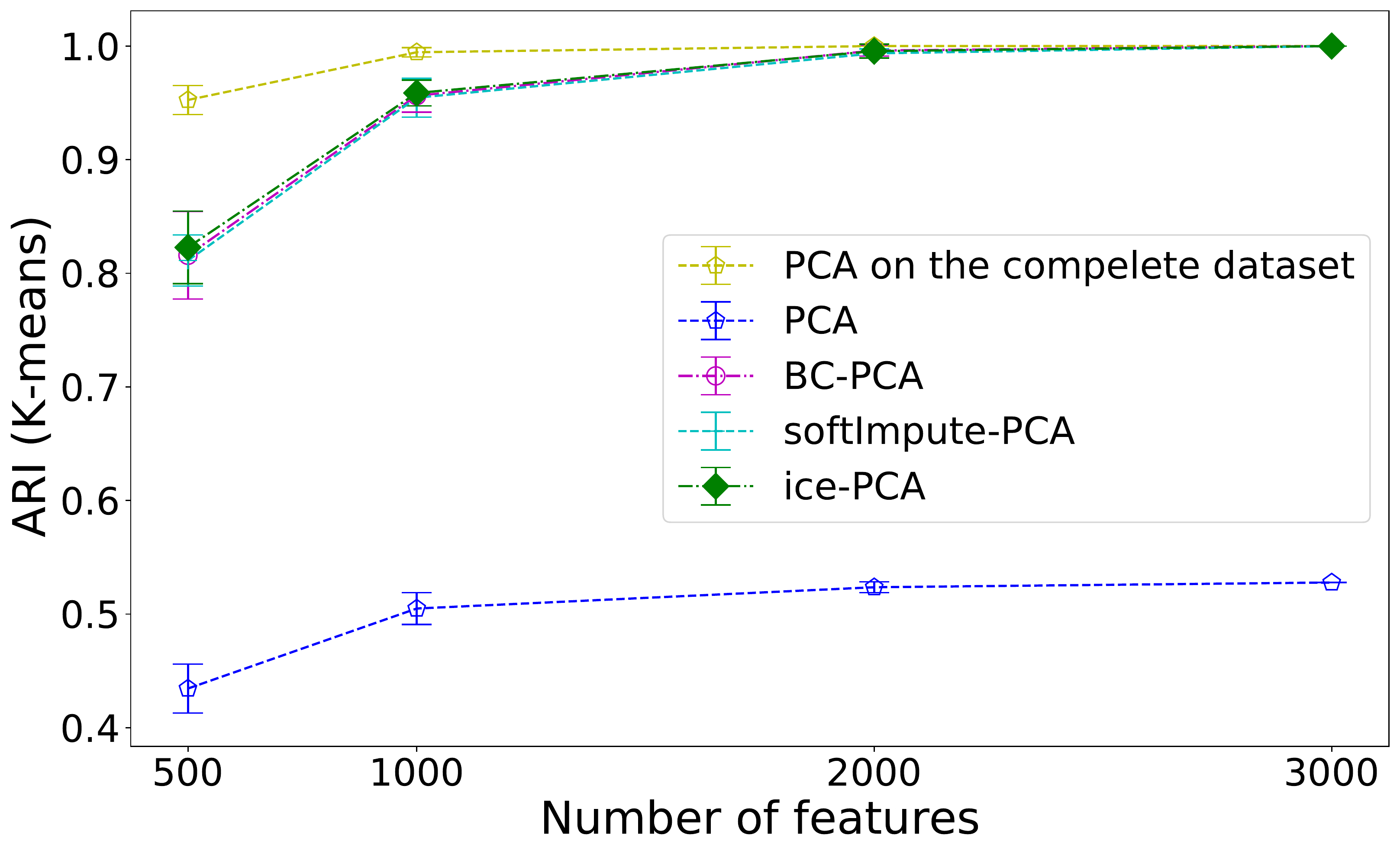}
    \caption{ARI of the $k$-means clustering results with different DR approaches on the simulated dataset with or without missing values.}
    \label{fig:sim_Kmeans}
\end{figure}

\begin{figure}[h]
    \centering
    \includegraphics[width=1\linewidth]{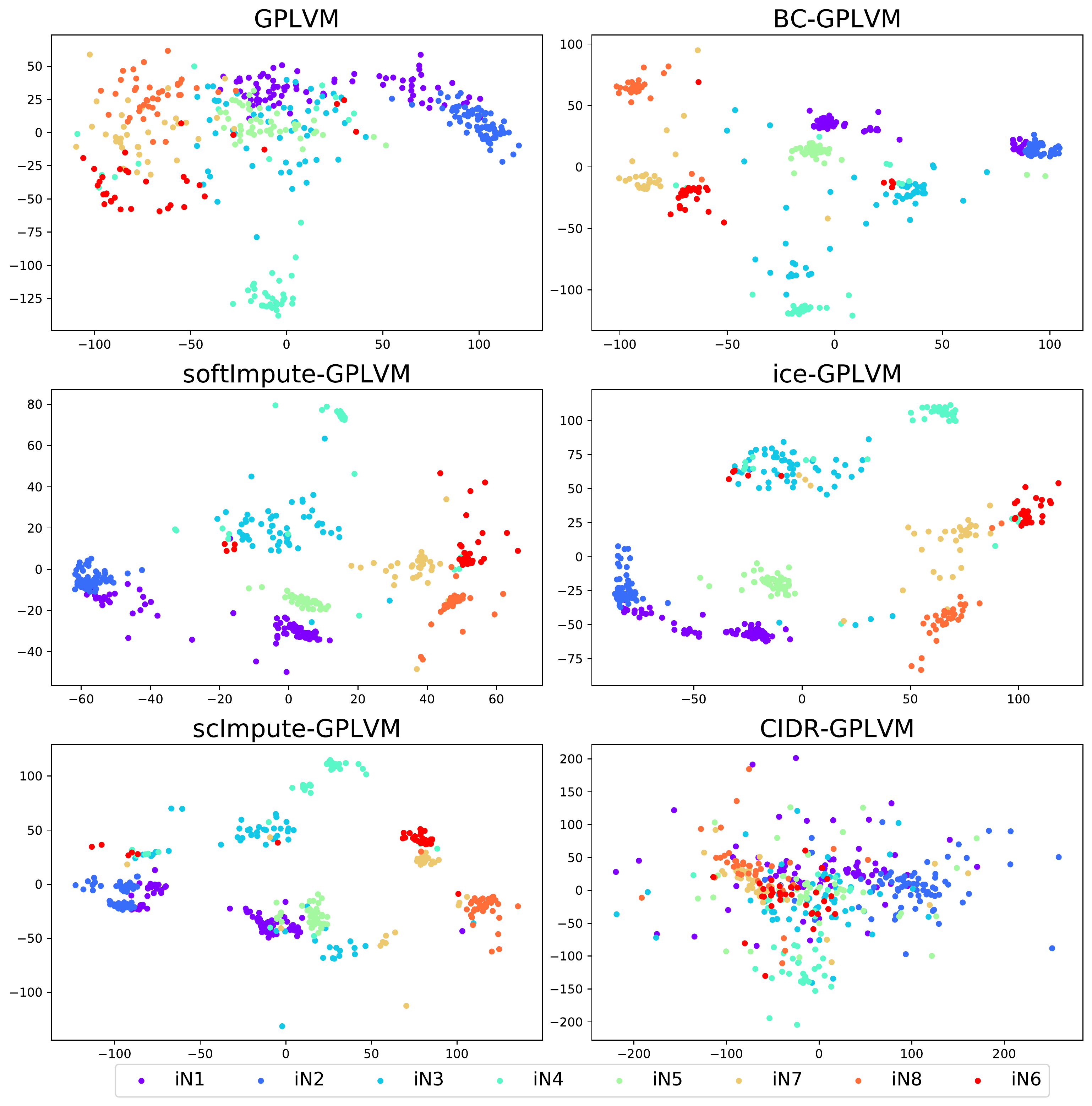}
    \caption{Visualisation of the Treutlein dataset obtained by GPLVM and its variants integrated with the bias correction or imputations.}
    \label{fig:Treutlein_GPLVM_vis}
\end{figure}

\begin{figure}[h]
    \centering
    \includegraphics[width=1\linewidth]{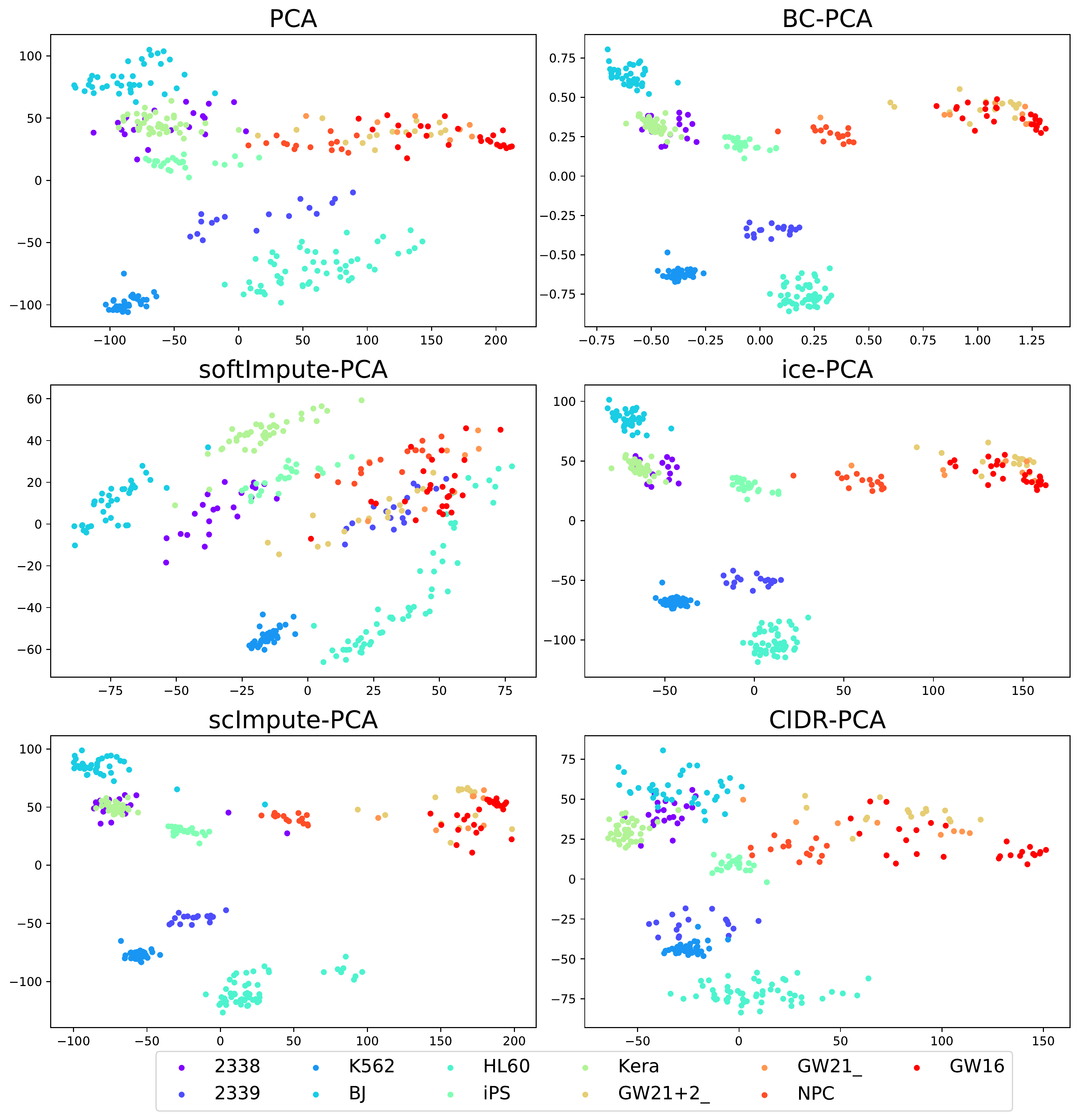}
    \caption{Visualisation of the Pollen dataset obtained by PCA and its variants integrated with the bias correction or imputations.}
    \label{fig:Pollen_PCA_vis}
\end{figure}

\begin{figure*}[t]
    \centering
     \subfloat[]{\includegraphics[width=0.9\linewidth]{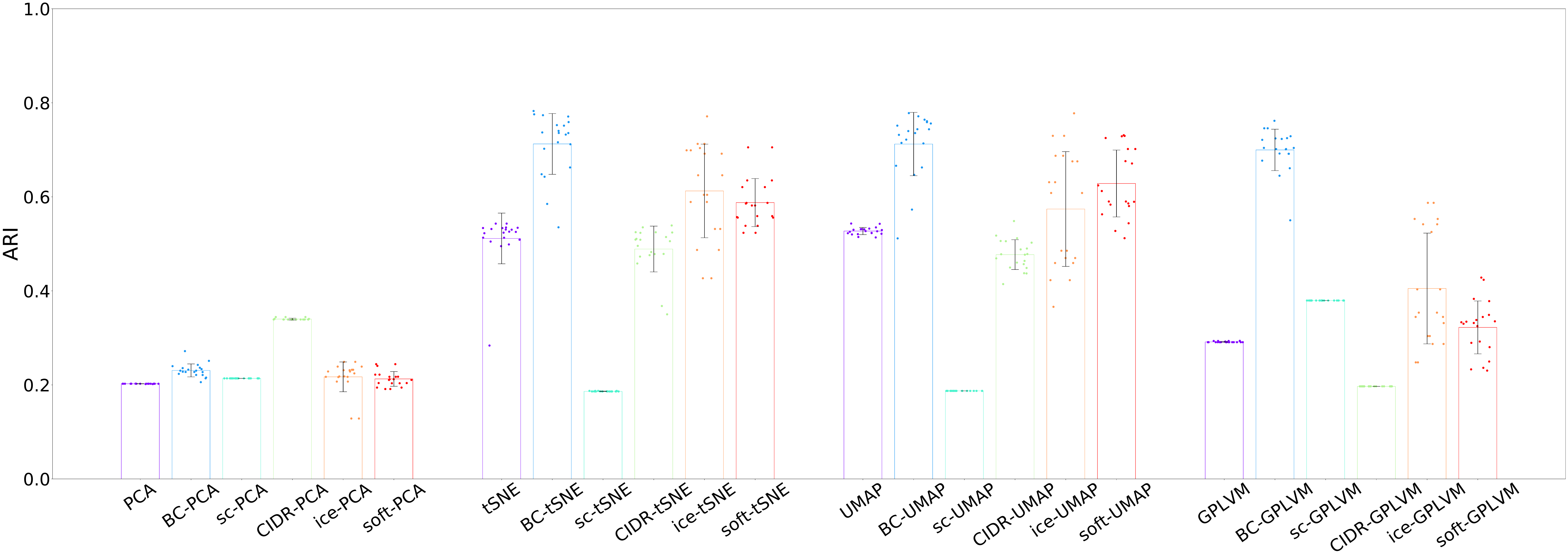}} \\
     \subfloat[]{\includegraphics[width=0.9\linewidth]{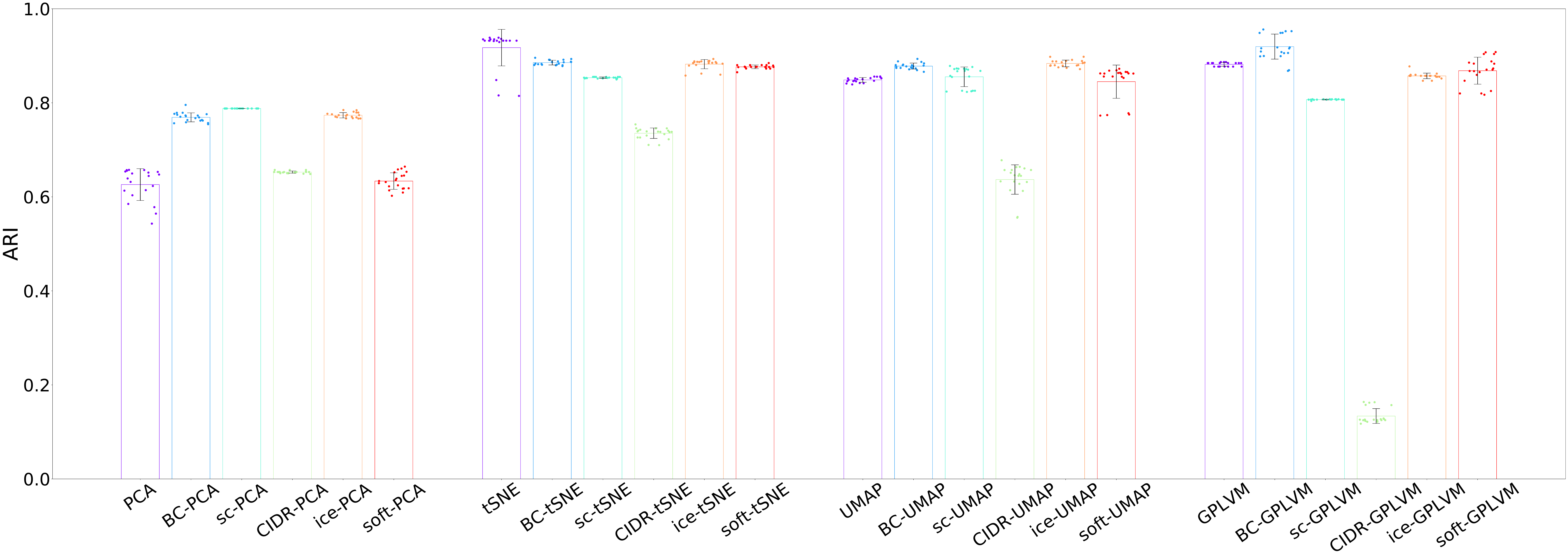}} 
    \caption{ARI of $k$-means with different DR approaches and their variants on the datasets (a) Usoskin, and (b) Pollen.}
    \label{fig:Kmeans_all}
\end{figure*}

\subsection{Input dimension influences the performance of dimension reduction}\label{ss:results:dim}

In order to examine the impact of input dimension on the performance of DR, we randomly select a subset of dimensions (features) from the simulated data. 
We then reduce the selected subset of data into two dimensions (2D). The qualities of the produced 2D projections are assessed according to the visualisation and the clustering accuracy. We repeat the aforementioned procedure $10$ times and average the ARI. The size of the subset features varies across $500$, $1000$, and $2000$. The performance using all features ($3000$) is also provided for comparison. Since the simulated dataset is generated in the context of PPCA, we compare only the DR methods based on PCA. 

First, we find that, on the simulated datasets with and without missing observations, the visualisation (Figure~\ref{fig:sim_vis}) is of a higher quality with more input features, based on the separation between different clusters.
The upward trends of the clustering performances on the simulated datasets shown in Figure~\ref{fig:sim_Kmeans} are consistent with the visualisation. In addition, more input features lead to a smaller deviation of the ARI.

Second, by comparing the performances between the dataset with missing observations and that without missing observations, we find that, with missing data, a higher input dimension is required to reach a performance comparable to that of the complete data. For instance, PCA leads to distinct clusters when the number of input features is $1000$ while BC-PCA renders clusters that are overlapping to some degree in such a case.

Overall, the experimental results on the simulated data offer empirical evidence confirming that the input dimension influences the performance of relevant DR approaches, as discussed in sec.\ref{ss:method:dim}.

\subsection{Bias correction improves visualisation}\label{ss:results:vis}

Now we examine whether the bias correction exploiting the information of missing data can produce better visualisation. First, we visualise the dimension-reduced data of the simulated datasets without and with bias-correction. Figure~\ref{fig:sim_vis} shows that PCA is capable of separating different clusters when no missing entry is present in data matrix. However, for data with missing observations, PCA cannot distinguish the subpopulations very clearly (Figure~\ref{fig:sim_vis}). In contrast, BC-PCA shows much more distinct clusters. Furthermore, BC-PCA has comparable performance to softImpute-PCA and ice-PCA in terms of the separation of clusters.

Next, we focus on the comparison between the benchmark DR methods and their bias-corrected versions in terms of the visualisations displayed by them on a wide spectrum of real datasets. Compared with PCA, BC-PCA presents more divergent clusters on the Pollen dataset and the Kumar dataset (Figure~\ref{fig:Pollen_PCA_vis} and Figure~\ref{figs:Kumar_PCA_vis} of the supplementary material), and it achieves comparable performance on the other datasets. BC-GPLVM succeeds in separating most clusters on the Treutlein dataset (Figure~\ref{fig:Treutlein_GPLVM_vis}), the Usoskin dataset (Figure~\ref{figs:Usoskin_GPLVM_vis} of the supplementary material), and the Koh dataset (Figure~\ref{figs:Koh_GPLVM_vis} of the supplementary material), showing a better performance than BC-PCA.
It may be due to the nonlinearity of data structure, which is difficult to be captured by a linear dimension reduction method like PCA even after the bias correction. Meanwhile, the degrees of overlapping between clusters are reduced greatly by BC-GPLVM on the fashion MNIST dataset (Figure~\ref{figs:fashion_MNIST_GPLVM_vis} of the supplementary material) and the Olivettic faces dataset (Figure~\ref{figs:Olivetti_faces_GPLVM_vis} of the supplementary material). 
Comparing GPLVM with BC-GPLVM, we find that BC-GPLVM often yields a clearer visualisation. 
Regarding the distance-matrix-based methods, both BC-tSNE and BC-UMAP clearly show superior visualisation to tSNE and UMAP, respectively.

\begin{figure}[htbp]
    \centering
    \includegraphics[width=1\linewidth]{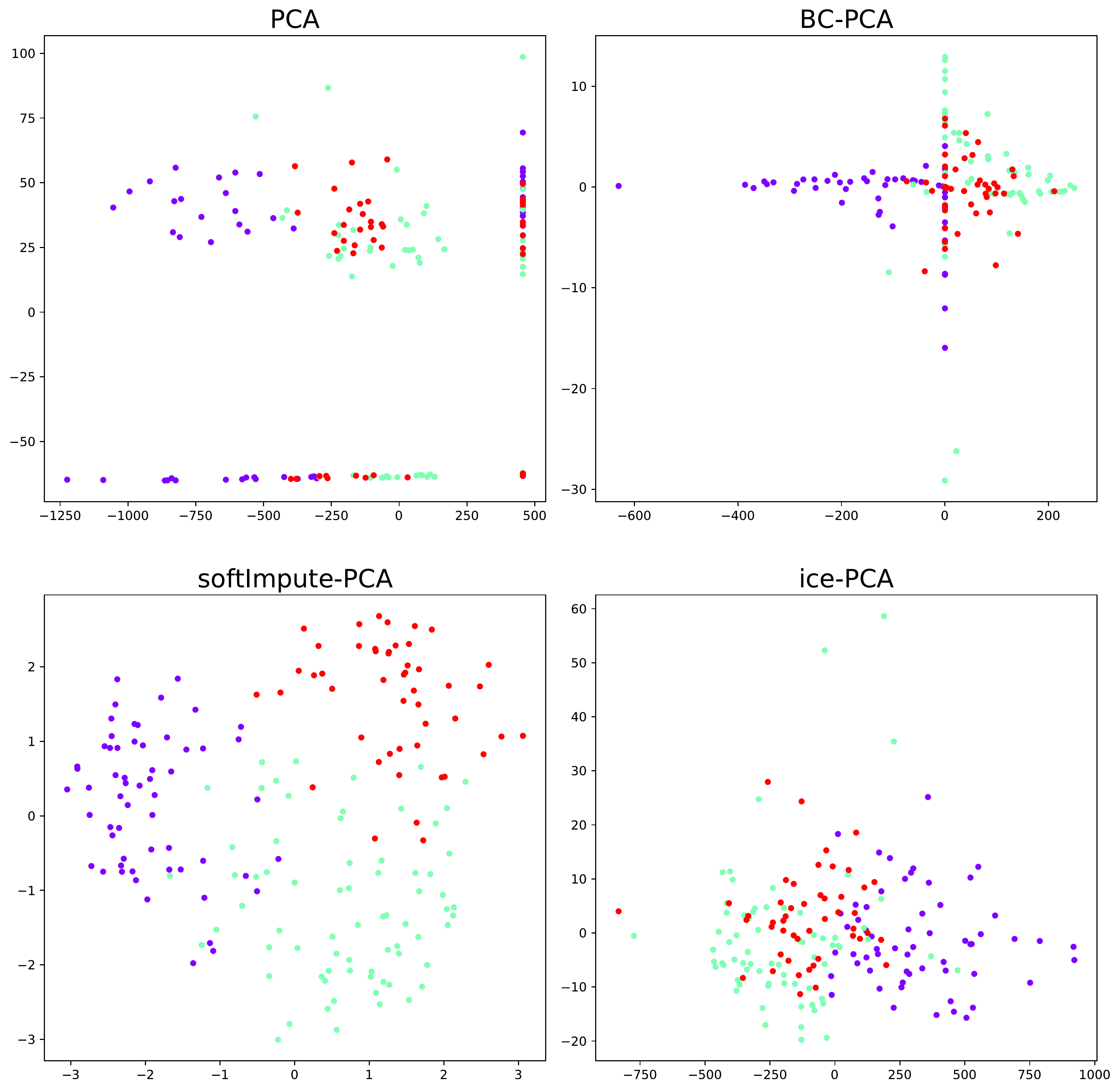}
    \caption{Visualisation of the wine dataset obtained by PCA and its variants integrated with the bias correction or imputations. Different colours represent different classes of wine.}
    \label{figs:wine_PCA_vis}
\end{figure}

\begin{figure}[htbp]
    \centering
    \includegraphics[width=1\linewidth]{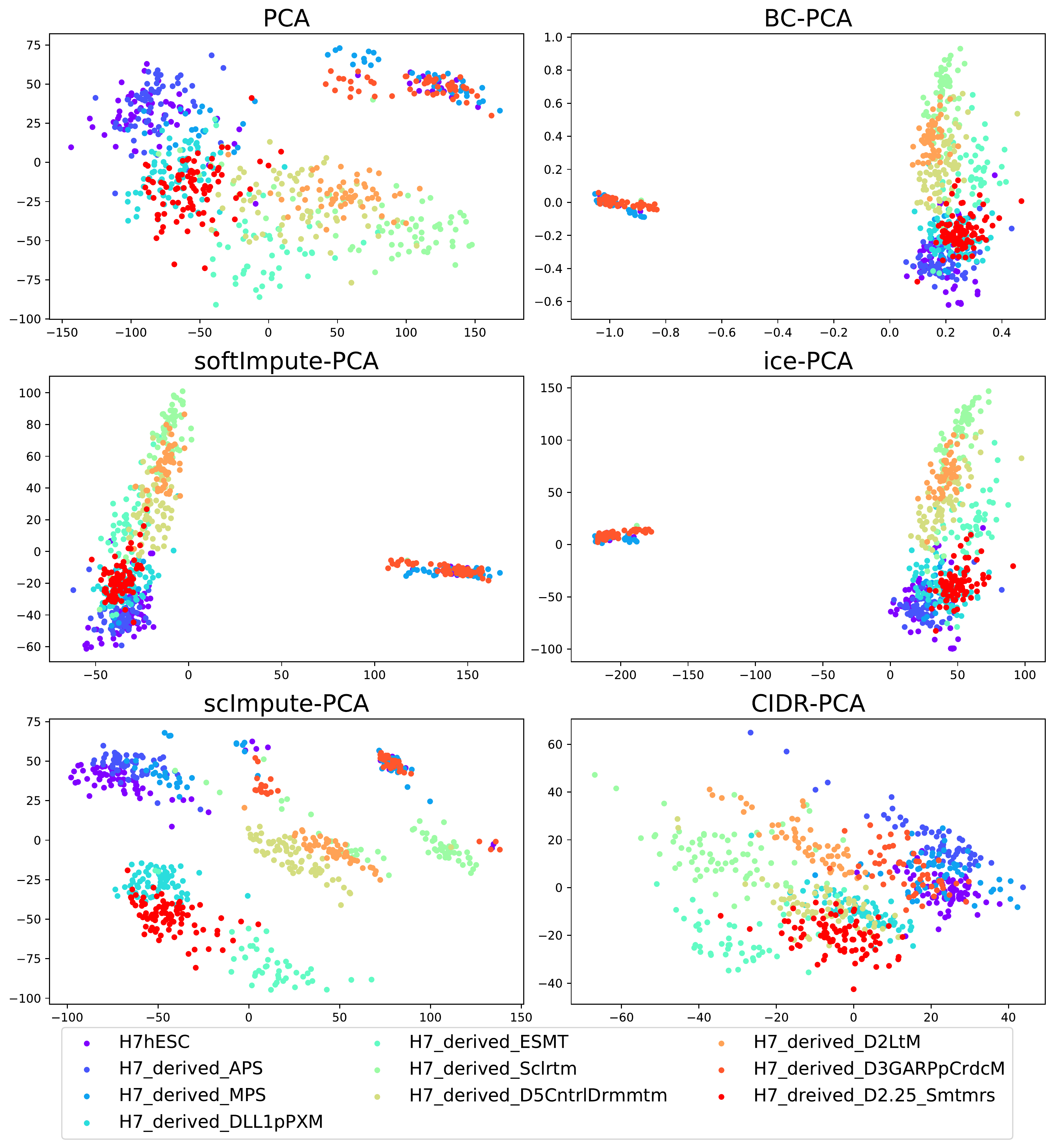}
    \caption{Visualisation of the Koh dataset obtained by PCA and its variants integrated with the bias correction or imputations.}
    \label{figs:Koh_PCA_vis}
\end{figure}

Last, we compare the proposed bias-corrected estimator with other imputation methods and the approaches specifically handling scRNA-seq data. 
The visualisations obtained by CIDR are inferior to those produced by the bias-corrected variants in terms of the separation between different groups of cells. 
Compared with the bias-corrected benchmark methods, scImpute yields more compact clusters on the Koh dataset (Figure~\ref{figs:Koh_PCA_vis}, Figure~\ref{figs:Koh_GPLVM_vis}, Figure~\ref{figs:Koh_tSNE_vis}, and Figure~\ref{figs:Koh_UMAP_vis} of the supplementary material). However, the within-cluster compactness and between-cluster separation are worse than or comparable to the approaches incorporating the unbiased Gram matrix on the other datasets. 
When applied to the fashion MNIST dataset and the Olivetti faces dataset, the proposed bias-corrected estimator is comparable to softImpute and ice; see the visualisations shown in Figure~\ref{figs:fashion_MNIST_GPLVM_vis} and Figure~\ref{figs:Olivetti_faces_GPLVM_vis} of the supplementary material. However, softImpute presents superior visualisations on the wine dataset (Figure~\ref{figs:wine_PCA_vis}, Figure~\ref{figs:wine_GPLVM_vis}, Figure~\ref{figs:wine_tSNE_vis}, and Figure~\ref{figs:wine_UMAP_vis} of the supplementary material). The bias-corrected DR methods fail to separate different clusters in such a case, since our method is proposed to the data that are high-dimensional while the wine dataset has only a small number of features. For the scRNA-seq datasets, the DR methods incorporating the bias correction match or outperform softImpute and ice, respectively. In particular, the visualisations attained by the bias correction on the Usoskin dataset are much more clear than those achieved by softImpute and ice (Figure~\ref{figs:Usoskin_GPLVM_vis}, Figure~\ref{figs:Usoskin_tSNE_vis} and Figure~\ref{figs:Usoskin_UMAP_vis} of the supplementary material).

To sum up, the superiority shown by the bias-corrected variants suggests that the proposed bias correction is beneficial for displaying better separation of clusters in the presence of missing observations. 

\subsection{Bias correction enhances clustering}\label{ss:results:clustering}

In this subsection, we investigate how the proposed bias correction impacts on the clustering applications. To this end, we first apply different DR methods and their variants to the dataset to extract the low-dimensional points, which are then grouped using the $k$-means clustering algorithm. The ARI is then calculated as a measure of clustering performance. For the real datasets, the dimension of latent points extracted from PCA and BC-PCA is chosen in terms of the Cattell–Nelson–Gorsuch scree test~\citep{CNG}, while only two-dimensional projections are produced with the other dimension reduction methods. Note that the dimension determined by the scree test is usually $2$ in our experiments. 

First, we assess the clustering performance obtained from using all features on the simulated data. On the simulated data without missing observations, the inferred labels obtained from PCA match perfectly with the ground truth labels in terms of ARI (Figure~\ref{fig:sim_Kmeans}). On the simulated data with missing observations, the original PCA is unable to provide distinct clusters in the low-dimensional space (Figure~\ref{fig:sim_vis}), and hence hinders the clustering (Figure~\ref{fig:sim_Kmeans}). On the contrary, their bias-corrected PCA presents nearly perfect ARI values, suggesting that the bias-correction significantly improves the clustering accuracy in such a case. 

Next, we compare benchmark DR methods with their variants integrating the bias-correction in terms of the clustering performance on the real datasets, as presented in Figure~\ref{fig:Kmeans_all}, Figure~\ref{fig:Kmeans_all_2} and  Figure~\ref{fig:Kmeans_all_3} of the supplementary material. Consistently with the visualisations, the $k$-means clustering performance of BC-tSNE and BC-UMAP is better than that of tSNE and UMAP on almost all datasets except the Deng dataset and the wine dataset. Similarly, the cluster labels obtained by BC-GPLVM show a much higher agreement with the ground truth labels than GPLVM on the Treutlein dataset, the Usoskin dataset, the Koh dataset, and the Olivetti faces dataset. For the other datasets, BC-GPLVM accomplishes the ARI values comparable to those of GPLVM. BC-PCA surpasses or is comparable to PCA on all the datasets except the Koh dataset which can be due to the nonlinearity possessed by the datasets and the wine dataset which is not suited for being handled by the proposed estimator, as we discussed in sec.\ref{ss:results:vis}.

Last, the bias-corrected estimators are compared with the imputation methods and the approaches handling scRNA-seq data. It is clear that the $k$-means results obtained by integrating CIDR with different DR methods are inferior to those attained by the bias correction, according to ARI.
Bias-corrected DR methods outperform DR approaches integrated with scImpute on the Usoskin, Treutlein and Pollen datasets. Moreover, all the values of ARI achieved by the proposed bias correction are nearly $1$ while the values of ARI of sc-tSNE and sc-UMAP are much lower than $1$ on the Kumar dataset. Although BC-tSNE and BC-UMAP perform slightly worse than sc-tSNE and sc-UMAP on the Koh dataset, BC-GPLVM achieves much higher ARI value.
On the Deng dataset, scImpute achieves higher ARI values when combined with the benchmark methods in comparison with the bias-corrected versions and the original ones. Generally speaking, the bias-corrected DR methods is better than scImpute on most datasets. 
When applied to the fashion MNIST dataset and the Olivetti dataset, the methods based on the bias correction often yield higher ARI compared to those integrating ice (Figure~\ref{fig:Kmeans_all_3} of the supplementary material), while their clustering performances are slightly worse than those attained by the methods inputting the data matrix imputed by softImpute. For the scRNA-seq datasets, the bias-corrected approaches outperforms softImpute on the Usoskin dataset, the Pollen dataset, and the Deng dataset. For the other scRNA-seq datasets, the bias-corrected variants accomplishes the ARI values comparable to or slightly worse than those obtained by softImpute. 


Overall, the clustering results indicate that the bias correction is able to infer the cluster labels that are more consistent with the ground truth and improve the performance of clustering following dimension reduction.

\section{Conclusion}\label{s:conclusion}
This paper proposes an unbiased estimator of the covariance matrix in the presence of missing data. The proposed bias-corrected Gram matrix is able to substantially improve the performance of various DR methods. 
As shown by the theoretical results in this paper, the Gram matrix is a biased estimator in the presence of missing observations and could be adverse for DR, while the proposed unbiased estimator can correct the bias introduced to the Gram matrix by the missingness. Moreover, the bounds of variances ensure the accurate estimation of the ground-truth covariance matrix in the low-dimensional space as long as the input dimension is high enough, and hence guarantees the reliable representation of the high-dimensional data. 
The experimental results on both simulated and real datasets demonstrate that the proposed unbiased estimator is widely applicable and is able to effectively enhance the performance of both the distance-matrix-based and Gram-matrix-based DR methods.

\begin{acknowledgements}
The authors thank the anonymous reviewers for helpful discussions and suggestions.
\end{acknowledgements}

\bibliography{main.bib}

\begin{thebibliography}{40}
\providecommand{\natexlab}[1]{#1}
\providecommand{\url}[1]{\texttt{#1}}
\expandafter\ifx\csname urlstyle\endcsname\relax
  \providecommand{\doi}[1]{doi: #1}\else
  \providecommand{\doi}{doi: \begingroup \urlstyle{rm}\Url}\fi

\bibitem[Candes and Plan(2010)]{matrix2}
Emmanuel~J. Candes and Yaniv Plan.
\newblock Matrix completion with noise.
\newblock \emph{Proceedings of the IEEE}, 98\penalty0 (6):\penalty0 925--936,
  2010.

\bibitem[Cand{\`e}s and Recht(2009)]{matrix1}
Emmanuel~J. Cand{\`e}s and Benjamin Recht.
\newblock Exact matrix completion via convex optimization.
\newblock \emph{Foundations of Computational Mathematics}, 9\penalty0
  (6):\penalty0 717--772, 2009.

\bibitem[Cho et~al.(2017)Cho, Kim, and Rohe]{svdpca}
Juhee Cho, Donggyu Kim, and Karl Rohe.
\newblock Asymptotic theory for estimating the singular vectors and values of a
  partially-observed low rank matrix with noise.
\newblock \emph{Statistica Sinica}, 27\penalty0 (4):\penalty0 1921--1948, 2017.

\bibitem[Deng et~al.(2014)Deng, Ramsk{\"o}ld, Reinius, and Sandberg]{Deng}
Qiaolin Deng, Daniel Ramsk{\"o}ld, Bj{\"o}rn Reinius, and Rickard Sandberg.
\newblock Single-cell {RNA}-seq reveals dynamic, random monoallelic gene
  expression in mammalian cells.
\newblock \emph{Science}, 343\penalty0 (6167):\penalty0 193--196, 2014.

\bibitem[Dua and Graff(2017)]{UCI}
Dheeru Dua and Casey Graff.
\newblock {UCI} machine learning repository, 2017.
\newblock URL \url{http://archive.ics.uci.edu/ml}.

\bibitem[Gorsuch and Nelson(1981)]{CNG}
RL~Gorsuch and J~Nelson.
\newblock {CNG} scree test: an objective procedure for determining the number
  of factors.
\newblock In \emph{Annual Meeting of the Society for Multivariate Experimental
  Psychology}, 1981.

\bibitem[Hastie et~al.(2015)Hastie, Mazumder, Lee, and Zadeh]{softimpute}
Trevor Hastie, Rahul Mazumder, Jason~D. Lee, and Reza Zadeh.
\newblock Matrix completion and low-rank svd via fast alternating least
  squares.
\newblock \emph{Journal of Machine Learning Research}, 16\penalty0
  (104):\penalty0 3367--3402, 2015.

\bibitem[Hicks et~al.(2017)Hicks, Townes, Teng, and Irizarry]{genomics}
Stephanie~C Hicks, F~William Townes, Mingxiang Teng, and Rafael~A Irizarry.
\newblock {Missing data and technical variability in single-cell RNA-sequencing
  experiments}.
\newblock \emph{Biostatistics}, 19\penalty0 (4):\penalty0 562--578, 2017.

\bibitem[Kolodziejczyk et~al.(2015)Kolodziejczyk, Kim, Svensson, Marioni, and
  Teichmann]{molecularcell}
Aleksandra~A. Kolodziejczyk, Jong~Kyoung Kim, Valentine Svensson, John~C.
  Marioni, and Sarah~A. Teichmann.
\newblock The technology and biology of single-cell {RNA} sequencing.
\newblock \emph{Molecular Cell}, 58\penalty0 (4):\penalty0 610 -- 620, 2015.
\newblock ISSN 1097-2765.

\bibitem[Kullback and Leibler(1951)]{kullback}
S.~Kullback and R.~A. Leibler.
\newblock On information and sufficiency.
\newblock \emph{The Annals of Mathematical Statistics}, 22\penalty0
  (1):\penalty0 79--86, 1951.

\bibitem[Kumar et~al.(2014)Kumar, Cahan, Shalek, Satija, Jay~DaleyKeyser, Li,
  Zhang, Pardee, Gennert, Trombetta, Ferrante, Regev, Daley, and
  Collins]{Kumar}
Roshan~M. Kumar, Patrick Cahan, Alex~K. Shalek, Rahul Satija,
  A.~Jay~DaleyKeyser, Hu~Li, Jin Zhang, Keith Pardee, David Gennert, John~J.
  Trombetta, Thomas~C. Ferrante, Aviv Regev, George~Q. Daley, and James~J.
  Collins.
\newblock Deconstructing transcriptional heterogeneity in pluripotent stem
  cells.
\newblock \emph{Nature}, 516\penalty0 (7529):\penalty0 56--61, 2014.

\bibitem[Lawrence(2005)]{gplvm}
Neil Lawrence.
\newblock Probabilistic non-linear principal component analysis with gaussian
  process latent variable models.
\newblock \emph{Journal of Machine Learning Research}, 6\penalty0
  (Nov):\penalty0 1783--1816, 2005.

\bibitem[Lehmann(2004)]{elements}
Erich~Leo Lehmann.
\newblock \emph{Elements of large-sample theory}.
\newblock Springer Science \& Business Media, 2004.

\bibitem[Li and Li(2018)]{scimpute}
Wei~Vivian Li and Jingyi~Jessica Li.
\newblock An accurate and robust imputation method scimpute for single-cell
  {RNA}-seq data.
\newblock \emph{Nature Communications}, 9\penalty0 (1):\penalty0 997, 2018.

\bibitem[Lin et~al.(2017)Lin, Troup, and Ho]{cidr}
Peijie Lin, Michael Troup, and Joshua W.~K. Ho.
\newblock {CIDR}: Ultrafast and accurate clustering through imputation for
  single-cell rna-seq data.
\newblock \emph{Genome Biology}, 18\penalty0 (1):\penalty0 59, 2017.

\bibitem[Little and Rubin(2019)]{little}
Roderick~JA Little and Donald~B Rubin.
\newblock \emph{Statistical analysis with missing data}.
\newblock John Wiley \& Sons, 2019.

\bibitem[Loh et~al.(2016)Loh, Chen, Koh, Deng, Sinha, Tsai, Barkal, Shen, Jain,
  Morganti, Shyh-Chang, Fernhoff, George, Wernig, Salomon, Chen, Vogel,
  Epstein, Kundaje, Talbot, Beachy, Ang, and Weissman]{Koh}
Kyle~M. Loh, Angela Chen, Pang~Wei Koh, Tianda~Z. Deng, Rahul Sinha,
  Jonathan~M. Tsai, Amira~A. Barkal, Kimberle~Y. Shen, Rajan Jain, Rachel~M.
  Morganti, Ng~Shyh-Chang, Nathaniel~B. Fernhoff, Benson~M. George, Gerlinde
  Wernig, Rachel~E.A. Salomon, Zhenghao Chen, Hannes Vogel, Jonathan~A.
  Epstein, Anshul Kundaje, William~S. Talbot, Philip~A. Beachy, Lay~Teng Ang,
  and Irving~L. Weissman.
\newblock Mapping the pairwise choices leading from pluripotency to human bone,
  heart, and other mesoderm cell types.
\newblock \emph{Cell}, 166\penalty0 (2):\penalty0 451 -- 467, 2016.

\bibitem[Maaten and Hinton(2008)]{tsne}
Laurens van~der Maaten and Geoffrey Hinton.
\newblock Visualizing data using {t-SNE}.
\newblock \emph{Journal of Machine Learning Research}, 9:\penalty0 2579--2605,
  2008.

\bibitem[Marlin et~al.(2007)Marlin, Zemel, Roweis, and Slaney]{cfmissing}
Benjamin~M. Marlin, Richard~S. Zemel, Sam Roweis, and Malcolm Slaney.
\newblock Collaborative filtering and the missing at random assumption.
\newblock In \emph{Proceedings of the Twenty-Third Conference on Uncertainty in
  Artificial Intelligence}, page 267–275, Arlington, Virginia, USA, 2007.
  AUAI Press.

\bibitem[Matthews et~al.(2017)Matthews, {van der Wilk}, Nickson, Fujii,
  {Boukouvalas}, {Le{\'o}n-Villagr{\'a}}, Ghahramani, and Hensman]{gpflow}
Alexander G. de~G. Matthews, Mark {van der Wilk}, Tom Nickson, Keisuke. Fujii,
  Alexis {Boukouvalas}, Pablo {Le{\'o}n-Villagr{\'a}}, Zoubin Ghahramani, and
  James Hensman.
\newblock {{GP}flow: A {G}aussian process library using {T}ensor{F}low}.
\newblock \emph{Journal of Machine Learning Research}, 18\penalty0
  (40):\penalty0 1--6, 2017.

\bibitem[McInnes et~al.(2018{\natexlab{a}})McInnes, Healy, and Melville]{umap}
Leland McInnes, John Healy, and James Melville.
\newblock {UMAP}: Uniform manifold approximation and projection for dimension
  reduction.
\newblock \emph{arXiv preprint arXiv:1802.03426}, 2018{\natexlab{a}}.

\bibitem[McInnes et~al.(2018{\natexlab{b}})McInnes, Healy, Saul, and
  Grossberger]{umap-software}
Leland McInnes, John Healy, Nathaniel Saul, and Lukas Grossberger.
\newblock {UMAP}: Uniform manifold approximation and projection.
\newblock \emph{The Journal of Open Source Software}, 3\penalty0 (29):\penalty0
  861, 2018{\natexlab{b}}.

\bibitem[Pearson(1901)]{pca}
Karl Pearson.
\newblock {LIII. O}n lines and planes of closest fit to systems of points in
  space.
\newblock \emph{The London, Edinburgh, and Dublin Philosophical Magazine and
  Journal of Science}, 2\penalty0 (11):\penalty0 559--572, 1901.

\bibitem[Pedregosa et~al.(2011)Pedregosa, Varoquaux, Gramfort, Michel, Thirion,
  Grisel, Blondel, Prettenhofer, Weiss, Dubourg, Vanderplas, Passos,
  Cournapeau, Brucher, Perrot, and Duchesnay]{scikit-learn}
F.~Pedregosa, G.~Varoquaux, A.~Gramfort, V.~Michel, B.~Thirion, O.~Grisel,
  M.~Blondel, P.~Prettenhofer, R.~Weiss, V.~Dubourg, J.~Vanderplas, A.~Passos,
  D.~Cournapeau, M.~Brucher, M.~Perrot, and E.~Duchesnay.
\newblock Scikit-learn: Machine learning in {P}ython.
\newblock \emph{Journal of Machine Learning Research}, 12:\penalty0 2825--2830,
  2011.

\bibitem[Pierson and Yau(2015)]{zifa}
Emma Pierson and Christopher Yau.
\newblock {ZIFA}: Dimensionality reduction for zero-inflated single-cell gene
  expression analysis.
\newblock \emph{Genome Biology}, 16\penalty0 (1):\penalty0 241, 2015.

\bibitem[Pollen et~al.(2014)Pollen, Nowakowski, Shuga, Wang, Leyrat, Lui, Li,
  Szpankowski, Fowler, Chen, et~al.]{pollen}
Alex~A Pollen, Tomasz~J Nowakowski, Joe Shuga, Xiaohui Wang, Anne~A Leyrat,
  Jan~H Lui, Nianzhen Li, Lukasz Szpankowski, Brian Fowler, Peilin Chen, et~al.
\newblock Low-coverage single-cell {mRNA} sequencing reveals cellular
  heterogeneity and activated signaling pathways in developing cerebral cortex.
\newblock \emph{Nature Biotechnology}, 32\penalty0 (10):\penalty0 1053, 2014.

\bibitem[Rubinsteyn and Feldman(2016)]{fancyimpute}
Alex Rubinsteyn and Sergey Feldman.
\newblock fancyimpute: An imputation library for python, 2016.
\newblock URL \url{https://github.com/iskandr/fancyimpute}.

\bibitem[Saliba et~al.(2014)Saliba, Westermann, Gorski, and Vogel]{gku555}
Antoine-Emmanuel Saliba, Alexander~J. Westermann, Stanislaw~A. Gorski, and
  Jörg Vogel.
\newblock {Single-cell {RNA}-seq: advances and future challenges}.
\newblock \emph{Nucleic Acids Research}, 42\penalty0 (14):\penalty0 8845--8860,
  2014.

\bibitem[Samaria and Harter(1994)]{olivetti}
F.S. Samaria and A.C. Harter.
\newblock Parameterisation of a stochastic model for human face identification.
\newblock In \emph{Proceedings of 1994 IEEE Workshop on Applications of
  Computer Vision}, pages 138--142, 1994.

\bibitem[Shapiro et~al.(2013)Shapiro, Biezuner, and
  Linnarsson]{shapiro2013single}
Ehud Shapiro, Tamir Biezuner, and Sten Linnarsson.
\newblock Single-cell sequencing-based technologies will revolutionize
  whole-organism science.
\newblock \emph{Nature Reviews Genetics}, 14\penalty0 (9):\penalty0 618, 2013.

\bibitem[Shen et~al.(2015)Shen, Li, Cheng, Zeng, Yang, Li, and
  Zhang]{remotesensing}
Huanfeng Shen, Xinghua Li, Qing Cheng, Chao Zeng, Gang Yang, Huifang Li, and
  Liangpei Zhang.
\newblock Missing information reconstruction of remote sensing data: A
  technical review.
\newblock \emph{IEEE Geoscience and Remote Sensing Magazine}, 3\penalty0
  (3):\penalty0 61--85, 2015.

\bibitem[Soneson and Robinson(2018)]{conquer}
Charlotte Soneson and Mark~D. Robinson.
\newblock Bias, robustness and scalability in single-cell differential
  expression analysis.
\newblock \emph{Nature Methods}, 15\penalty0 (4):\penalty0 255--261, 2018.

\bibitem[Tipping and Bishop(1999)]{ppca}
Michael~E. Tipping and Christopher~M. Bishop.
\newblock Probabilistic principal component analysis.
\newblock \emph{Journal of the Royal Statistical Society. Series B (Statistical
  Methodology)}, 61\penalty0 (3):\penalty0 611--622, 1999.
\newblock ISSN 13697412, 14679868.

\bibitem[Torgerson(1952)]{mds}
Warren~S Torgerson.
\newblock Multidimensional scaling: I. theory and method.
\newblock \emph{Psychometrika}, 17\penalty0 (4):\penalty0 401--419, 1952.

\bibitem[Treutlein et~al.(2016)Treutlein, Lee, Camp, Mall, Koh, Shariati, Sim,
  Neff, Skotheim, Wernig, and Quake]{Treutlein}
Barbara Treutlein, Qian~Yi Lee, J.~Gray Camp, Moritz Mall, Winston Koh, Seyed
  Ali~Mohammad Shariati, Sopheak Sim, Norma~F. Neff, Jan~M. Skotheim, Marius
  Wernig, and Stephen~R. Quake.
\newblock Dissecting direct reprogramming from fibroblast to neuron using
  single-cell {RNA}-seq.
\newblock \emph{Nature}, 534\penalty0 (7607):\penalty0 391--395, 2016.

\bibitem[Usoskin et~al.(2015)Usoskin, Furlan, Islam, Abdo, L{\"o}nnerberg, Lou,
  Hjerling-Leffler, Haeggstr{\"o}m, Kharchenko, Kharchenko, et~al.]{usoskin}
Dmitry Usoskin, Alessandro Furlan, Saiful Islam, Hind Abdo, Peter
  L{\"o}nnerberg, Daohua Lou, Jens Hjerling-Leffler, Jesper Haeggstr{\"o}m,
  Olga Kharchenko, Peter~V Kharchenko, et~al.
\newblock Unbiased classification of sensory neuron types by large-scale
  single-cell {RNA} sequencing.
\newblock \emph{Nature Neuroscience}, 18\penalty0 (1):\penalty0 145, 2015.

\bibitem[Van~Buuren(2018)]{ice1}
Stef Van~Buuren.
\newblock \emph{Flexible imputation of missing data}.
\newblock CRC press, 2018.

\bibitem[Van Der~Maaten et~al.(2009)]{van2009dimensionality}
L.~Van Der~Maaten et~al.
\newblock Dimensionality reduction: a comparative review.
\newblock \emph{Technical report, Tilburg University, TiCC-TR 2009-005}, 2009.

\bibitem[Xiao et~al.(2017)Xiao, Rasul, and Vollgraf]{MNIST}
Han Xiao, Kashif Rasul, and Roland Vollgraf.
\newblock Fashion-mnist: a novel image dataset for benchmarking machine
  learning algorithms.
\newblock \emph{CoRR}, abs/1708.07747, 2017.

\bibitem[Zhu et~al.(2019)Zhu, Wang, and Samworth]{primePCA}
Ziwei Zhu, Tengyao Wang, and Richard~J Samworth.
\newblock High-dimensional principal component analysis with heterogeneous
  missingness.
\newblock \emph{arXiv preprint arXiv:1906.12125}, 2019.

\end{thebibliography}

\clearpage
\section{Supplementary Material}

\subsection{Dataset pre-processing}\label{secs:data_preprocessing}
For the pre-processing of the scRNA-seq datasets, genes (features) observed in less than $2$ cells (observations) are first discarded, followed by a  log2 transformation with pseudo $1$ count added. For the other real datasets, we keep all features obtained from the original publications. 

\subsection{Data availability}\label{secs:download}
\begin{itemize}
  \item Pollen: cells in this dataset are defined by the human cell lines. We directly download the TPM values from \url{https://s3.amazonaws.com/scrnaseq-public-datasets/manual-data/pollen/NBT_hiseq_linear_tpm_values.txt}.
    \item Deng: this dataset was used to study monoallelic expression at the single cell level. The data are available in the Gene Expression Omnibus (GEO) database under the accession number GSE45719 (\url{https://www.ncbi.nlm.nih.gov/geo/query/acc.cgi?acc=GSE45719}).
    \item Treutlein: this dataset contains cell populations from direct reprogramming from fibroblast to neuron (MEF, day 2, 5, and 22). The data are available in the GEO database under the accession number GSE67310 (\url{https://www.ncbi.nlm.nih.gov/geo/query/acc.cgi?acc=GSE67310}).
    \item Koh: this dataset comprises 9 different cell types that include H7 hESCs, H7-derived anterior primitive streak populations, H7-derived mid primitive streak populations, H7-derived lateral mesoderm, H7-derived FACS-purified GARP+ cardiac mesoderm, H7-derived FACS-purified DLL1+ paraxial mesoderm populations, H7-derived day 3 early somite progenitor populations, H7-derived dermomyotome populations, and H7-derived FACS-purified PDGFR$\alpha$+ sclerotome populations. RNA was extracted from either whole cell populations or, alternatively, cell subsets purified by fluorescence activated cell sorting (FACS). We download the processed data from conquer, a repository of consistently processed, analysis-ready public scRNA-seq datasets~\cite{conquer} (\url{http://imlspenticton.uzh.ch:3838/conquer/}).
    \item Usoskin: this dataset consists of 11 types of mouse lumbar DRG (dorsal root ganglion). We download the normalized data from \url{http://linnarssonlab.org/drg/} (External resource table 1).
    \item Kumar: this dataset contains three populations of mouse embryonic stem cells. The data are available in the GEO database under the accession number GSE60749 (\url{https://www.ncbi.nlm.nih.gov/geo/query/acc.cgi?acc=GSE60749}). 
    \item Olivetti faces: this dataset is loaded using a sklearn function in Python
    (\url{https://scikit-learn.org/stable/modules/generated/sklearn.datasets.fetch_olivetti_faces.html}). 
    \item fashion MNIST: this dataset is loaded using Keras 
    (\url{https://keras.io/api/datasets/fashion_mnist/}). 
    \item wine: this dataset is loaded using a sklearn function in Python
    (\url{https://scikit-learn.org/stable/modules/generated/sklearn.datasets.load_wine.html?highlight=wine%20dataset}).
\end{itemize}

\subsection{Estimation of missingness probabilities}\label{secs:dropout_prob}

Suppose that the probability of non-missingness for the $i$-th cell and the $s$-th gene is given by $p_{is}=c_i g_s$,
where $c_i$ and $g_s$ account for the influence from the $i$-th cell and $s$-th gene, respectively.

By using the method of moments, we get
\begin{align*}
    \mathbb{E}\left [\sum_{s=1}^{D}h_{is} \right] = c_i \sum_{s=1}^{D}g_s
    =M_{i\cdot},\\
    \mathbb{E}\left [\sum_{i=1}^{N}h_{is} \right] =g_s\sum_{i=1}^{N}c_i
    = M_{\cdot s},
\end{align*}
where $ M_{i\cdot} = \sum_{s=1}^{D}M_{is}$, $M_{\cdot s} = \sum_{i=1}^{N}M_{is}$, $M$ is the indicator matrix with 1 representing non-missingness event, and $s=1,\ldots,D; \ i=1,\ldots,N$. It is straightforward to verify that $\bar{c}_i = \frac{M_{i \cdot}}{N} /\sqrt{\frac{m}{DN}}$ and $\bar{g}_{s} = \frac{M_{\cdot s}}{D}/\sqrt{\frac{m}{DN}}$, where $m=\sum_i M_{i\cdot}=\sum_s M_{\cdot s}$, satisfy the above equations. However, it is likely that $\bar{c}_i\bar{g}_{s}>1$. Hence, we normalise the estimator by $\bar{c}_i = \frac{M_{i \cdot}}{N\sqrt{n_c}}$ and $\bar{g}_s = \frac{M_{\cdot s}}{D\sqrt{n_c}}$, where $n_c = \text{max}_{i \in [N], s \in [D]}( \frac{M_{i \cdot}M_{\cdot s}}{DN}, \frac{m}{DN})$.

\subsection{Implementation details}\label{secs:implementation}
There are several hyperparameters in the procedure of distinguishing between dropouts and biological non-expression. We use $k$-means clustering, a key ingredient in analysing scRNA-seq data, and spectral clustering, accounting for the nonlinearity in the data. For each clustering method, a wide range of cluster numbers are assigned: $(4,6,8,10,12)$. The kernel coefficient in the spectral clustering is set to be the average distance to the $7$-th nearest neighbour. All other parameters are used as default. 

A zero count would be deemed to be biological non-expression if the proportion of similar cells showing zero expression in the same gene exceeds the threshold $85\%$. In practice, we found that this threshold should be higher than $0.5$, and we would get better results when it ranges from $0.7$ to $0.95$. 

For each DR benchmark or its variant when applied to the scRNA-seq datasets, we replicate the procedure of first performing DR and then applying $k$-means $20$ times. Each time when performing the $k$-means algorithm on the extracted low-dimensional components, the number of repeats of $k$-means starting with different centroid initialisations is set as $30$ for more reliable results.

For the simulated datasets, we first sample subsets of features, followed by performing DR techniques. Last, we run the $K$-means algorithms on the low-dimensional data points with $30$ different centroid initialisations and obtain the most reliable clustering results in terms of inertia. 

All clustering methods and ice (IterativeImputer) are implemented with scikit-learn v0.22.1 of Python~\citep{scikit-learn}. 
For GPLVM, tSNE and UMAP, we use the GPflow package v1.3.0~\citep{gpflow}, the scikit-learn package v0.21.3~\citep{scikit-learn} and the umap-learn package v0.3.10~\citep{umap-software}, with default settings, respectively. For softImpute, we use the fancyimpute package v0.5.5~\citep{fancyimpute}.

\subsection{Proofs}\label{secs:proofs}
\subsubsection{Proof for Proposition~1}
For $i \neq j$, based on the law of total expectation, we get 
\begin{align*}
    &\mathbb{E}\left [ y_{is}y_{js} \right ]  =  \mathbb{E}\left [\mathbb{E} \left [ y_{is}y_{js}  \mid h_{is}, h_{js}\right ] \right ] \\
    & = \mathbb{E} \left [ y_{is}y_{js}  \mid h_{is}=1 \, \text{and} \, h_{js}=1\right ] p_{is} p_{js} \\
    &+ \mathbb{E} \left [ y_{is}y_{js}  \mid h_{is}=0 \, \text{and} \,h_{js}=0\right ] \left ( 1 - p_{is} \right )\left ( 1 - p_{js} \right ) \\
    & + \mathbb{E} \left [ y_{is}y_{js}  \mid h_{is}=1 \, \text{and} \, h_{js}=0\right ] \left ( 1 - p_{js} \right )p_{is} \\
   &+ \mathbb{E} \left [ y_{is}y_{js}  \mid h_{is}=0 \, \text{and} \,h_{js}=1\right ] \left ( 1 - p_{is} \right )p_{js}\\
    & = \mathbb{E} \left [ y_{is}y_{js}  \mid h_{is}=1 \right ] p_{is} p_{js} = p_{is}p_{js} K_{ij} , 
\end{align*}
\begin{align*}
    \mathrm{Var}\left [ y_{is}y_{js} \right ] & =  
    \mathbb{E}\left [y_{is}^2y_{js}^2 \right ]
    - \mathbb{E}\left [ y_{is}y_{js} \right ] ^2 \\
    & =  \mathbb{E}\left [\mathbb{E} \left [ y_{is}^2y_{js}^2  \mid h_{is},h_{js} \right ] \right ] - \mathbb{E}\left [ y_{is}y_{js} \right ] ^2 \\
    &= \mathbb{E}\left [\mathbb{E} \left [ y_{is}^2y_{js}^2  \mid h_{is},h_{js} \right ] \right ] - p_{is}^2p_{js}^2K_{ij}^2 \\
    & = \left (K_{ii}K_{jj} + 2K_{ij}^2 \right )p_{is}p_{js} - 
    p_{is}^2p_{js}^2K_{ij}^2 \\
    & =  p_{is}p_{js}{K}_{ii}{K}_{jj} +{K}_{ij}^2 \left (2p_{is}p_{js}-p_{is}^2p_{js}^2 \right).
\end{align*}
For $i=j$, we get
\begin{align*}
    \mathbb{E}\left [ y_{is}^2 \right ] & =  \mathbb{E}\left [\mathbb{E} \left [ y_{is}^2  \mid h_{is} \right ] \right ] \\
    & = \mathbb{E} \left [ y_{is}^2  \mid h_{is}=1 \right ] p_{is} + \mathbb{E} \left [ y_{is}^2  \mid h_{is}=0 \right ] \left ( 1 - p_{is} \right ) \\
    & = \mathbb{E} \left [ y_{is}^2  \mid h_{is}=1 \right ] p_{is} = p_{is}K_{ii} , 
\end{align*}
\begin{align*}
    \mathrm{Var}\left [ y_{is}^2\right ] & =  \mathbb{E}\left [y_{is}^4 \right ]
    - \mathbb{E}\left [ y_{is}^2 \right ] ^2 \\
    & = \mathbb{E}\left [\mathbb{E} \left [ y_{is}^4  \mid h_{is} \right ] \right ] - \mathbb{E}\left [ y_{is}^2 \right ] ^2 \\
    & = 3K_{ii}^2p_{is} - K_{ii}^2p_{is}^2 = K_{ii}^2 \left( 3p_{is} - p^2_{is}\right).
\end{align*}

\subsubsection{Proof for Proposition~2}
Based on Proposition~1, it is straightforward to get that the elements in the unbiased estimator $\tilde{G}$ of $K$ are given by 
\begin{align*}
\begin{cases}
 \tilde{G}_{ij} = \frac{G_{ij}}{\sum_{s=1}^D p_{is}p_{js}}, \,\,\text{for}  \, \, i \neq j; 
 \\ 
\tilde{G}_{ii} = \frac{G_{ii}}{\sum_{s=1}^D p_{is}},
\end{cases}
\end{align*}
and the corresponding variances are given by
\begin{align*}
\begin{cases}
\mathrm{Var}\left[ \tilde{G}_{ij}\right] = \frac{ {K}_{ii}{K}_{jj}\sum_{s=1}^D p_{is}p_{js} + {K}_{ij}^2 \sum_{s=1}^D (2p_{is}p_{js}-p_{is}^2p_{js}^2)}
   {\left( \sum_{s=1}^D p_{is}p_{js} \right)^2}, \,\, {\rm for} \,\, i \neq j; \\
  \mathrm{Var}\left[ \tilde{G}_{ii}\right] = \frac{ {K}_{ii}^2 \sum_{s=1}^D p_{is}(3 - p_{is})}
     {\left( \sum_{s=1}^D p_{is} \right)^2}.
\end{cases}
\end{align*}
We first consider the bounds for $\mathrm{Var}\left[ \tilde{G}_{ij}\right] , \,\, {\rm for} \,\, i \neq j.$ $\mathrm{Var}\left[ \tilde{G}_{ij}\right]$ can be re-written as

\begin{align*}
    \mathrm{Var}\left[ \tilde{G}_{ij}\right] & =\frac{ {K}_{ii}{K}_{jj}\sum_{s=1}^D p_{is}p_{js} + {K}_{ij}^2 \sum_{s=1}^D (2p_{is}p_{js}-p_{is}^2p_{js}^2)}  {\left( \sum_{s=1}^D p_{is}p_{js} \right)^2} \\
    & = \frac{{K}_{ii}{K}_{jj}}{\bar{p}_{ij} D} +
     \frac{K_{ij}^2}{D}\frac{\displaystyle \sum_{s=1}^{D}(\frac{2p_{is}p_{js}}{D}  - \frac{ p_{is}^2p_{js}^2}{D})}{\displaystyle(\frac{\sum_{s=1}^{D}p_{is}p_{js}}{D})^2} \\
     & = \frac{{K}_{ii}{K}_{jj}}{\bar{p}_{ij} D} +
     \frac{ K_{ij}^2}{D}\frac{\displaystyle (2\bar{p}_{ij}  - \frac{ \sum_{s=1}^{D}p_{is}^2p_{js}^2}{D})}{\bar{p}_{ij}^2},
\end{align*}
where $0<\bar{p}_{ij} = \frac{1}{D}\sum_{s=1}^D p_{is}p_{js}\leq 1$. Note that $ 2\bar{p}_{ij}  - \frac{\sum_{s=1}^{D}p_{is}^2p_{js}^2}{D} \geq\bar{p}_{ij}$,  We thus get the lower bound for $ \mathrm{Var}\left[ \tilde{G}_{ij}\right] $:
\begin{align*}
 \mathrm{Var}\left[ \tilde{G}_{ij}\right]  \geq \frac{{K}_{ii}{K}_{jj}}{\bar{p}_{ij} D} + \frac{K_{ij}^2}{D}\frac{\bar{p}_{ij}}{\bar{p}_{ij}^2} = 
 \frac{{K}_{ii}{K}_{jj} + K_{ij}^2}{D \bar{p}_{ij}} 
\end{align*}

Meanwhile, since $f(x) = 2x-x^2$ (with $E[f(x)] \leq f[E(x)]$) is a strictly concave function, we have 
\begin{align*}
\mathrm{Var}\left[ \tilde{G}_{ij}\right] & = 
\frac{{K}_{ii}{K}_{jj}}{\bar{p}_{ij} D} +
     K_{ij}^2 D\frac{\displaystyle \sum_{s=1}^{D}(\frac{2p_{is}p_{js}}{D}  - \frac{ p_{is}^2p_{js}^2}{D})}{(\sum_{s=1}^{D}p_{is}p_{js})^2} \\
     &  \leq \frac{{K}_{ii}{K}_{jj}}{\bar{p}_{ij} D}
    + {K}_{ij}^2 D
     \frac
     {\displaystyle \frac{2\sum_{s=1}^D p_{is}p_{js}}{D} - \left(\frac{\sum_{s=1}^D p_{is}p_{js}}{D} \right)^2}
     { \left( \sum_{s=1}^D p_{is}p_{js} \right)^2 } \\
   & = \frac{{K}_{ii}{K}_{jj}}{\bar{p}_{ij} D} +
     \frac{{K}_{ij}^2}{D}\left( \frac{2}{\bar{p}_{ij}} - 1\right).
\end{align*}
Next, we consider the bounds for  $\mathrm{Var}\left[ \tilde{G}_{ii}\right] $. By using the fact that $\sum_{s=1}^{D}p^2_{is} \leq \sum_{s=1}^{D}p_{is} $ we get
\begin{align*}
  \mathrm{Var}\left[ \tilde{G}_{ii}\right] = \frac{3K^2_{ii}}{D \bar{p}_{i}}  - \frac{K_{ii}^2 \sum_{s=1}^{D}p^2_{is} }{(\sum_{s=1}^{D}p_{is})^2}
  & \geq \frac{2K^2_{ii}}{D \bar{p}_{i}} , 
\end{align*}
where $ \bar{p}_{i} = \frac{1}{D} \sum_{s=1}^D p_{is}$. Again, since since $g(x) = 3x-x^2$ is a strictly concave function, we have 
\begin{align*}
  \mathrm{Var}\left[ \tilde{G}_{ii}\right] & = \frac{K_{ii}^2 \sum_{s=1}^{D}( 3p_{is} - p^2_{is} )}{(\sum_{s=1}^{D}p_{is})^2} \\
  & \leq {K}_{ii}^2 D
     \frac
     {\displaystyle \frac{3\sum_{s=1}^D p_{is}}{D} - \left(\frac{\sum_{s=1}^D p_{is}}{D} \right)^2}
     { \left( \sum_{s=1}^D p_{is} \right)^2 } \\
     & = \frac{{K}_{ii}^2}{D}\left( \frac{3}{\bar{p}_{i}} - 1\right).
\end{align*}

\subsubsection{Proof for Proposition~3}
First, we consider the limiting distribution of $Z_{ij,D}$  for $i \neq j$.  We verify that for any $\epsilon >0$ , 
\begin{align*}
\lim_{D\to\infty} \frac{1}{S_{ij,D}^2} \sum_{s=1}^{D} \mathrm{E} \left[   (x_{ij,s} - \mu_{ij,s})^2 \mathbbm{1} _{\left \{ \left | x_{ij,s} - \mu_{ij,s} \right | > \epsilon S_{ij,D} \right \} }\right]
= 0, 
\end{align*}
where 
\begin{align*}
    S_{ij,D}^2 =\sum\limits_{s=1}^{D}  \mathrm{Var} (x_{ij,s}) = {K}_{ii}{K}_{jj}\sum\limits_{s=1}^D p_{is}p_{js} \\
    + {K}_{ij}^2 \sum\limits_{s=1}^D (2p_{is}p_{js}-p_{is}^2p_{js}^2),
\end{align*}
and $\mathbbm{1}_{\left \{ \cdot \right \}}$ is the indicator function.  Based on the law of total expectation, we get 

\begin{equation}\label{eq:1}
\begin{aligned}
   & \frac{1}{S_{ij,D}^2} \sum_{s=1}^{D} \mathrm{E} \left[   (x_{ij,s} - \mu_{ij,s})^2 \mathbbm{1} _{\left   \{ \left | x_{ij,s} - \mu_{ij,s} \right | > \epsilon S_{ij,D} \right \} }\right] \\ & =  
    \frac{1}{S_{ij,D}^2}  \sum_{s=1}^{D}(1 - p_{is}p_{js})    \mu_{ij,s}^2   \mathbbm{1}_{\left \{ \left | \mu_{ij,s} \right |> \epsilon S_{ij,D} \right \}} \\
     & +\frac{1}{S_{ij,D}^2} \sum_{s=1}^{D} p_{is}p_{js} \mathrm{E}\left [ (\tilde{y}_{is} \tilde{y}_{js} - \mu_{ij,s})^2 \mathbbm{1}_{\left \{ \left | \tilde{y}_{is} \tilde{y}_{js}  - \mu_{ij,s}\right | > \epsilon S_{ij,D} \right \}}\right ]  
\end{aligned}
\end{equation}

Since  $\sum\limits_{s=1}^D p_{is}p_{js}\asymp D$, there exist constants $0<m < M< \infty$, and an integer $n_0$ such that 
\begin{align*}
m < \frac{\sum_{s=1}^D p_{is}p_{js}}{D} <M, \,\, \text{for all } D>n_0.
\end{align*}Furthermore, we obtain
\begin{align*}
&\frac{p_{is}^2p_{js}^2K_{ij}^2}{S_{ij,D}^2}  \leq \frac{p_{is}^2p_{js}^2K_{ij}^2}{\left ( K_{ii}K_{jj} + K_{ij}^2 \right ) \sum_{s=1}p_{is}p_{js}} \\
& < \frac{p_{is}^2p_{js}^2K_{ij}^2}{\left ( K_{ii}K_{jj} + K_{ij}^2 \right ) mD}
\leq \frac{1}{mD\left ( \frac{K_{ii}K_{jj}}{K_{ij}^2} + 1 \right )}, \,\, for \,\, D>n_0.
\end{align*}
Let $n = \text{max}\left \{ \frac{1}{\epsilon^2 m\left ( \frac{K_{ii}K_{jj}}{K_{ij}^2} + 1 \right )}, n_0 \right \}$, we get $\frac{p_{is}^2p_{js}^2K_{ij}^2}{S_{ij,D}^2}  < \epsilon^2,\,\, \text{for }D>n$. It is hence clear that 
\begin{equation}\label{eq:2}
\lim_{D\to\infty}\frac{1}{S_{ij,D}^2}  \sum_{s=1}^{D}    (1 - p_{is}p_{js}) \mu_{ij,s}^2   \mathbbm{1}_{\left \{ \left | \mu_{ij,s} \right |> \epsilon S_{ij,D} \right \}}  = 0.
\end{equation}

Without loss of generality, suppose that $K_{ij} \geq 0$, with $0<p_{is} \leq 1$ for any $i$ and $s$, we get $\epsilon S_{ij,D} + p_{is}p_{js}K_{ij} > \epsilon S_{ij,D} $ and  $p_{is}p_{js}K_{ij}  - \epsilon S_{ij,D} \leq K_{ij} - \epsilon S_{ij,D}$.  Since $(\tilde{y}_{is} \tilde{y}_{js} - \mu_{ij,s})^2 \geq 0 $, we obtain that 
\begin{equation}\label{eq:3}
\begin{aligned}
& 0 \leq 
\frac{1}{S_{ij,D}^2} \sum_{s=1}^{D} p_{is}p_{js} \mathrm{E}\left [ (\tilde{y}_{is} \tilde{y}_{js} - \mu_{ij,s})^2 \mathbbm{1}_{\left \{ \left | \tilde{y}_{is} \tilde{y}_{js}  - \mu_{ij,s}\right | > \epsilon S_{ij,D} \right \}}\right ] \\
&\leq
\frac{1}{S_{ij,D}^2} \sum_{s=1}^{D} p_{is}p_{js} \\
&\mathrm{E}\left [ (\tilde{y}_{is} \tilde{y}_{js} - \mu_{ij,s})^2 
\left ( \mathbbm{1}_{\left \{ \tilde{y}_{is} \tilde{y}_{js} > \epsilon S_{ij,D} \right \}}  +
\mathbbm{1}_{\left \{ \tilde{y}_{is} \tilde{y}_{js} < K_{ij} - \epsilon S_{ij,D} \right \}} \right )
\right ]
\end{aligned}
\end{equation}
To investigate the limit of 
\begin{align*}
    \frac{1}{S_{ij,D}^2} \sum\limits_{s=1}^{D} p_{is}p_{js} \mathrm{E}\left [ (\tilde{y}_{is} \tilde{y}_{js} - \mu_{ij,s})^2 \mathbbm{1}_{\left \{ \left | \tilde{y}_{is} \tilde{y}_{js}  - \mu_{ij,s}\right | > \epsilon S_{ij,D} \right \}}\right ]
\end{align*}
in~\eqref{eq:3}, we first investigate the limit of 
\begin{equation}\label{eq:4}
\begin{aligned}
&\frac{1}{S_{ij,D}^2} \sum_{s=1}^{D} p_{is}p_{js} \mathrm{E}\left [ (\tilde{y}_{is} \tilde{y}_{js} - \mu_{ij,s})^2 
 \mathbbm{1}_{\left \{ \tilde{y}_{is} \tilde{y}_{js} > \epsilon S_{ij,D} \right \}}  
\right ] \\
 & =\frac{1}{S_{ij,D}^2}\sum_{s=1}^{D} p_{is}p_{js} \mathrm{E}\left [ (\tilde{y}_{is} \tilde{y}_{js} )^2 
 \mathbbm{1}_{\left \{ \tilde{y}_{is} \tilde{y}_{js} > \epsilon S_{ij,D} \right \}}  
\right ] \\
 &+\frac{1}{S_{ij,D}^2}\sum_{s=1}^{D} p_{is}^3p_{js}^3K_{ij}^2 \mathrm{E}\left [ 
 \mathbbm{1}_{\left \{ \tilde{y}_{is} \tilde{y}_{js} > \epsilon S_{ij,D} \right \}}  
\right ] \\
& - \frac{2}{S_{ij,D}^2}\sum_{s=1}^{D} p_{is} p_{js} K_{ij}\mathrm{E}\left [ \tilde{y}_{is} \tilde{y}_{js} 
 \mathbbm{1}_{\left \{ \tilde{y}_{is} \tilde{y}_{js} > \epsilon S_{ij,D} \right \}}  
\right ].
\end{aligned}
\end{equation}
The limit of above equation can be obtained by showing that the upper bound for each term goes to $0$. Based on the fact that $\tilde{y}_{i1}\tilde{y}_{j1}, \, \, \tilde{y}_{i2}\tilde{y}_{j2},\, \, \ldots$ are iid, the upper bounds for the terms in~\eqref{eq:4} are respectively given by
\begin{equation}\label{eq:5}
\begin{aligned}
&0\leq 
\frac{ \left( \sum_{s=1}^{D} p_{is}p_{js} \right) \mathrm{E}\left [ (\tilde{y}_{is} \tilde{y}_{js} )^2 
 \mathbbm{1}_{\left \{ \tilde{y}_{is} \tilde{y}_{js} > \epsilon S_{ij,D} \right \}}  
\right ] }{S_{ij,D}^2 }  \\
&\leq 
\frac{\mathrm{E}\left [ (\tilde{y}_{is} \tilde{y}_{js} )^2 
 \mathbbm{1}_{\left \{ \tilde{y}_{is} \tilde{y}_{js} > \epsilon S_{ij,D} \right \}}  
\right ]}{K_{ii} K_{jj} + K_{ij}^2}, \\
& 0 \leq 
\frac{K_{ij}^{2} \left(\sum_{s=1}^{D} p_{is}^{3} p_{js}^{3}  \right) \mathrm{E}\left [ 
 \mathbbm{1}_{\left \{ \tilde{y}_{is} \tilde{y}_{js} > \epsilon S_{ij,D} \right \}}  
\right ] }{S_{ij,D}^{2}} \\
&\leq 
K_{ij}^2 \frac{\mathrm{E}\left [  \mathbbm{1}_{\left \{ \tilde{y}_{is} \tilde{y}_{js} > \epsilon S_{ij,D} \right \}}  
\right ]}{K_{ii} K_{jj} + K_{ij}^2}, \\
& 0 \leq
\frac{K_{ij} \left(\sum_{s=1}^{D} p_{is}^{2} p_{js}^{2}  \right) \mathrm{E}\left [ \tilde{y}_{is} \tilde{y}_{js} 
 \mathbbm{1}_{\left \{ \tilde{y}_{is} \tilde{y}_{js} > \epsilon S_{ij,D} \right \}}  
\right ] }{S_{ij,D}^{2}} \\
&\leq 
K_{ij} \frac{\mathrm{E}\left [  \tilde{y}_{is} \tilde{y}_{js} \mathbbm{1}_{\left \{ \tilde{y}_{is} \tilde{y}_{js} > \epsilon S_{ij,D} \right \}}  
\right ]}{K_{ii} K_{jj} + K_{ij}^2}. 
\end{aligned}
\end{equation}
Since 
\begin{align*}
     \left | \tilde{y}_{is}^2 \tilde{y}_{js}^2  \mathbbm{1}_{\left \{ \tilde{y}_{is} \tilde{y}_{js} > \epsilon S_{ij,D} \right \}}   \right | \leq
     \tilde{y}_{is}^2 \tilde{y}_{js}^2, \\
     \left | \tilde{y}_{is} \tilde{y}_{js}  \mathbbm{1}_{\left \{ \tilde{y}_{is} \tilde{y}_{js} > \epsilon S_{ij,D} \right \}}   \right | \leq
     \left | \tilde{y}_{is} \tilde{y}_{js} \right |,
\end{align*}
and 
\begin{align*}
\mathrm{E}\left [\tilde{y}_{is}^2 \tilde{y}_{js}^2  \right ], \mathrm{E}\left [\left | \tilde{y}_{is} \tilde{y}_{js} \right | \right ] < \infty,
\end{align*}
the dominated convergence theorem implies that 
\begin{align*}
\lim_{D\to\infty} \frac{\mathrm{E}\left [ (\tilde{y}_{is} \tilde{y}_{js} )^2 
 \mathbbm{1}_{\left \{ \tilde{y}_{is} \tilde{y}_{js} > \epsilon S_{ij,D} \right \}}  
\right ]}{K_{ii} K_{jj} + K_{ij}^2} = 0,\\
\lim_{D\to\infty} K_{ij} \frac{\mathrm{E}\left [  \tilde{y}_{is} \tilde{y}_{js} \mathbbm{1}_{\left \{ \tilde{y}_{is} \tilde{y}_{js} > \epsilon S_{ij,D} \right \}}  
\right ]}{K_{ii} K_{jj} + K_{ij}^2} = 0.
\end{align*}
Moreover, since $\mathbbm{1}_{\left \{ \tilde{y}_{is} \tilde{y}_{js} > \epsilon S_{ij,D} \right \}}\overset{P}{\rightarrow} 0$, it is clear that $ K_{ij}^2 \frac{\mathrm{E}\left [  \mathbbm{1}_{\left \{ \tilde{y}_{is} \tilde{y}_{js} > \epsilon S_{ij,D} \right \}}  
\right ]}{K_{ii} K_{jj} + K_{ij}^2} $ goes to $0$ based on the bounded convergence theorem. Thus, the limits of all upper bounds shown in~\eqref{eq:5} are $0$ and we get 
\begin{align*}
\begin{split}
    &\lim_{D\to\infty} \frac{1}{S_{ij,D}^2} \sum_{s=1}^{D} p_{is}p_{js} \mathrm{E}\left [ (\tilde{y}_{is} \tilde{y}_{js} - \mu_{ij,s})^2 
    \mathbbm{1}_{\left \{ \tilde{y}_{is} \tilde{y}_{js} > \epsilon S_{ij,D} \right \}}  
    \right ] \\
    &= 0.
\end{split}
\end{align*}
Analogously, we can obtain 
\begin{align*}
\begin{split}
    & \lim_{D\to\infty} \frac{1}{S_{ij,D}^2} \sum_{s=1}^{D} p_{is}p_{js} \mathrm{E}\left [ (\tilde{y}_{is} \tilde{y}_{js} - \mu_{ij,s})^2 
    \mathbbm{1}_{\left \{ \tilde{y}_{is} \tilde{y}_{js} < K_{ij} - \epsilon S_{ij,D} \right \}}
\right ] \\
 & = 0.
\end{split}
\end{align*}
Hence, by taking the limits of both sides of inequality provided in~\eqref{eq:3}, we get 
\begin{align}\label{eq:6}
\begin{split}
    &\lim_{D\to\infty}  \frac{1}{S_{ij,D}^2} \sum_{s=1}^{D} p_{is}p_{js} \mathrm{E}\left [ (\tilde{y}_{is} \tilde{y}_{js} - \mu_{ij,s})^2 \mathbbm{1}_{\left \{ \left | \tilde{y}_{is} \tilde{y}_{js}  - \mu_{ij,s}\right | > \epsilon S_{ij,D} \right \}}\right ] \\
    &=0 .
\end{split}
\end{align}
Furthermore, by taking the limits of both sides of~\eqref{eq:1}, we get 
\begin{align*}
\begin{split}
    &\lim_{D\to\infty} \frac{1}{S_{ij,D}^2} \sum_{s=1}^{D}\mathrm{E} \left[   (x_{ij,s} - \mu_{ij,s})^2 \mathbbm{1} _{\left \{ \left | x_{ij,s} - \mu_{ij,s} \right | > \epsilon S_{ij,D} \right \} }\right]\\
    &= 0,
\end{split}
\end{align*}
based on the limiting results provided in~\eqref{eq:2} and \eqref{eq:6}. The Lindeberg's condition is therefore satisfied and $Z_{ij,D} \overset{dist.}{\rightarrow} \mathcal{N} (0,1)$ for $i \neq j$. For $i=j$,  $Z_{ij,D} \overset{dist.}{\rightarrow} \mathcal{N} (0,1)$ can be proved by following the same logic as that for showing $Z_{ij,D} \overset{dist.}{\rightarrow} \mathcal{N} (0,1)$ for $i \neq j$. 

\subsubsection{Proof for Corollary~1}
For $i \neq j$, $\tilde{G}_{ij} = \frac{Z_{ij,D} S_{ij,D} + \sum\limits_{s=1}^{D} \mu_{ij,s}}{\sum\limits_{s=1}^{D} p_{is}p_{js}} = Z_{ij,D} \frac{S_{ij,D}}{\sum\limits_{s=1}^{D} p_{is}p_{js}}
+ \frac{\sum\limits_{s=1}^{D} \mu_{ij,D}}{\sum\limits_{s=1}^{D} p_{is}p_{js}}$. Since
\begin{align*}
\begin{split}
    & \frac{S^2_{ij,D}}{\left ( \sum\limits_{s=1}^{D} p_{is}p_{js} \right)^2} 
    \leq
    \frac{K_{ii}K_{jj}}{\sum\limits_{s=1}^{D} p_{is}p_{js} } + K_{ij}^2\left ( \frac{2}{\sum\limits_{s=1}^{D} p_{is}p_{js}}  - \frac{1}{D}\right )
    \rightarrow 0,\\
    &\text{as }D\rightarrow \infty; \\
    & \frac{\sum\limits_{s=1}^{D} \mu_{ij,s}}{\sum\limits_{s=1}^{D} p_{is}p_{js}} = K_{ij},
\end{split}
\end{align*}
by the Slutsky's theorem, we obtain that $\tilde{G}_{ij} \overset{P}{\rightarrow} K_{ij}$ for $i \neq j$. Analogously, we get $\tilde{G}_{ii} \overset{P}{\rightarrow} K_{ii}$.

\subsubsection{Proof for Corollary~2}
It is straightforward to get the Corollary by using the Slutsky's theorem and $\frac{G}{D} = \frac{G}{ \sum\limits_{s=1}^{D} p_{is}p_{js}} \frac{ \sum\limits_{s=1}^{D} p_{is}p_{js} }{D}$.

\subsection{More experimental results}
\subsubsection{Visualisation of PCA results on the real datasets}
\clearpage
\begin{figure}[t]
    \centering
    \includegraphics[width=1\linewidth]{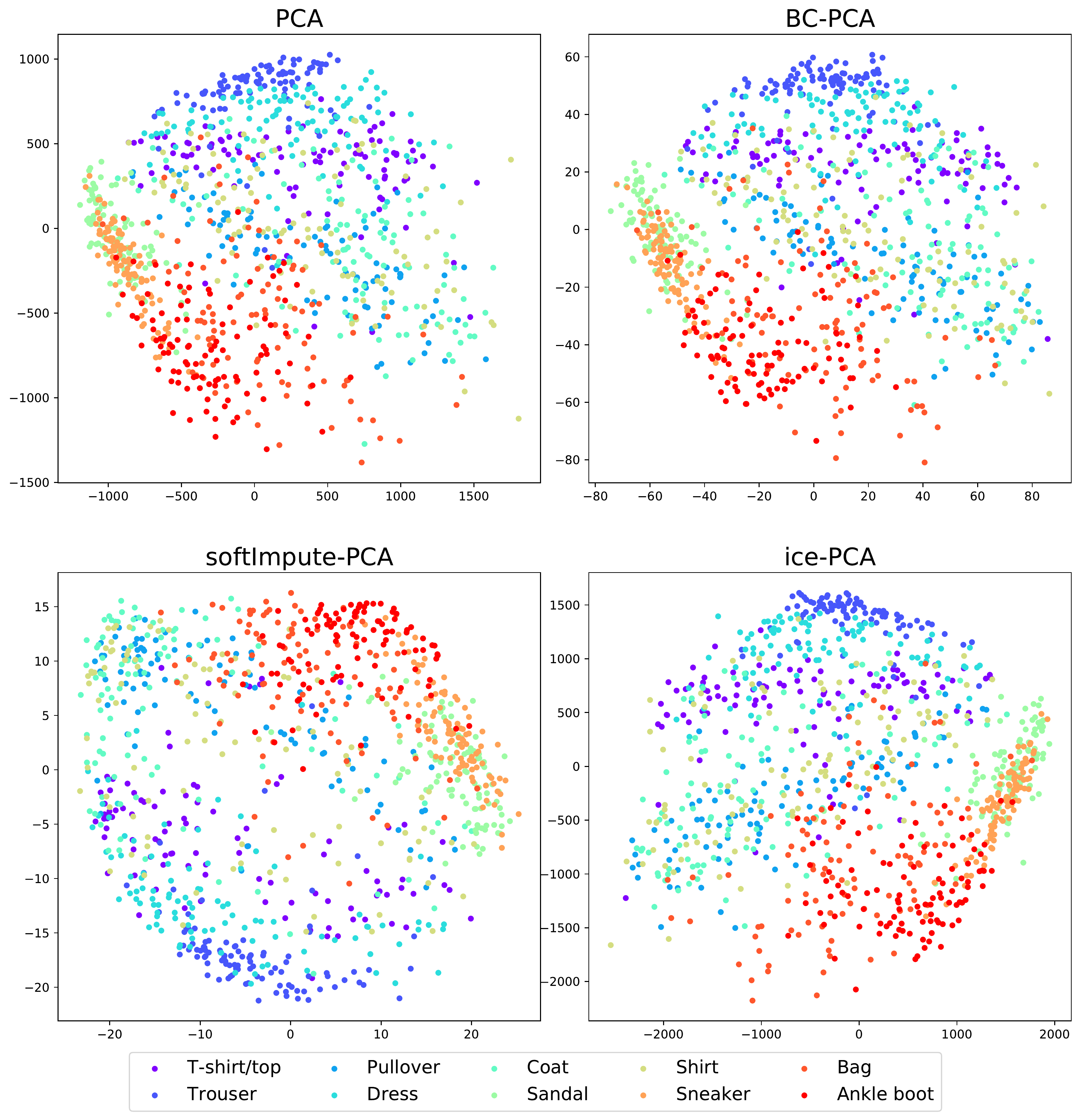}
    \caption{Visualisation of the fashion MNIST dataset obtained by PCA and its variants integrated with the bias correction or imputations.}
    \label{figs:fashion_MNIST_PCA_vis}
\end{figure}

\begin{figure}[h]
    \centering
    \includegraphics[width=1\linewidth]{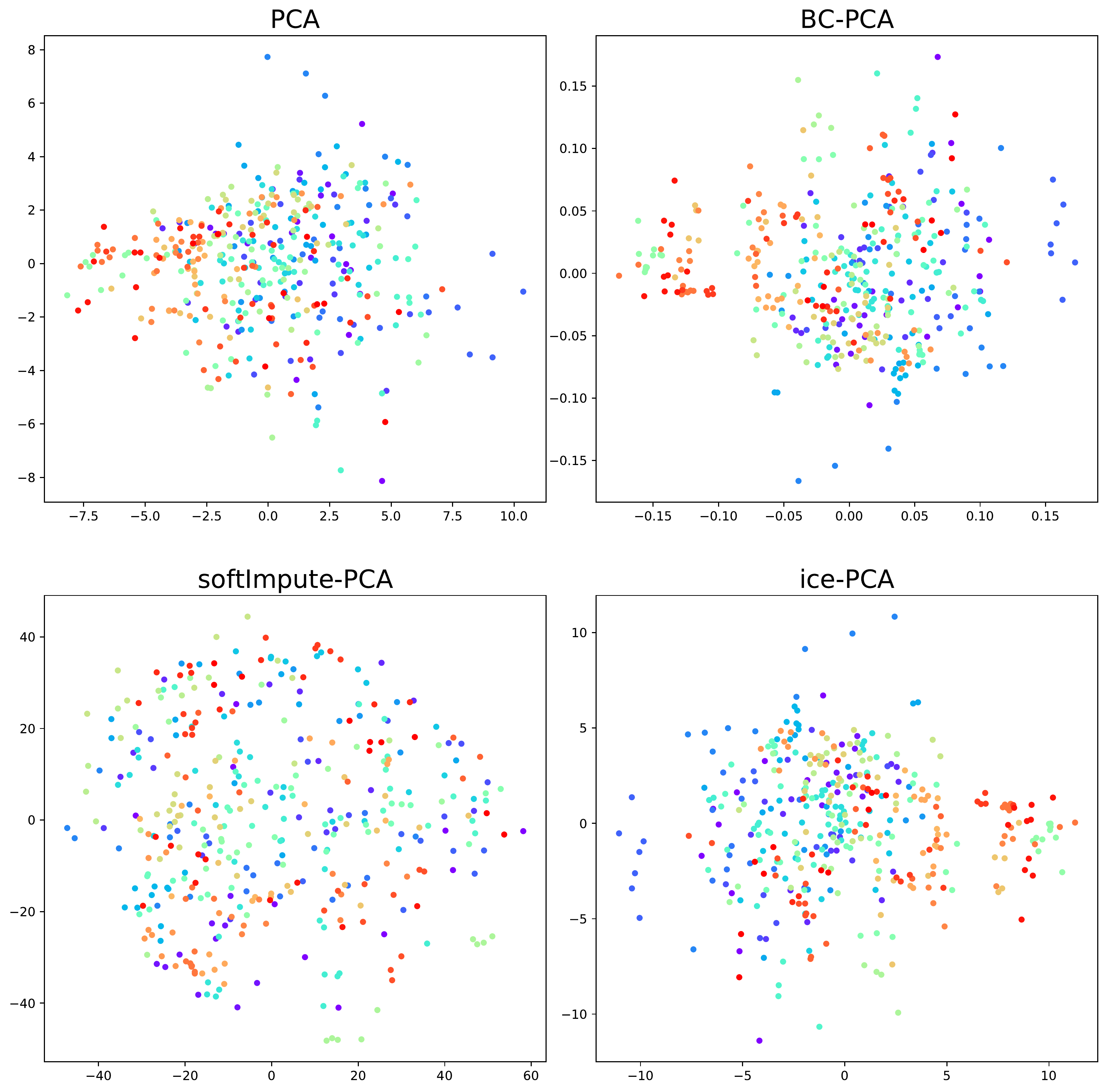}
    \caption{Visualisation of the Olivetti faces dataset obtained by PCA and its variants integrated with the bias correction or imputations. Different colours represent face images of different persons, and there are 40 persons in total.}
    \label{figs:Olivetti_faces_PCA_vis}
\end{figure}

\newpage
\begin{figure}[htbp]
    \centering
    \includegraphics[width=1\linewidth]{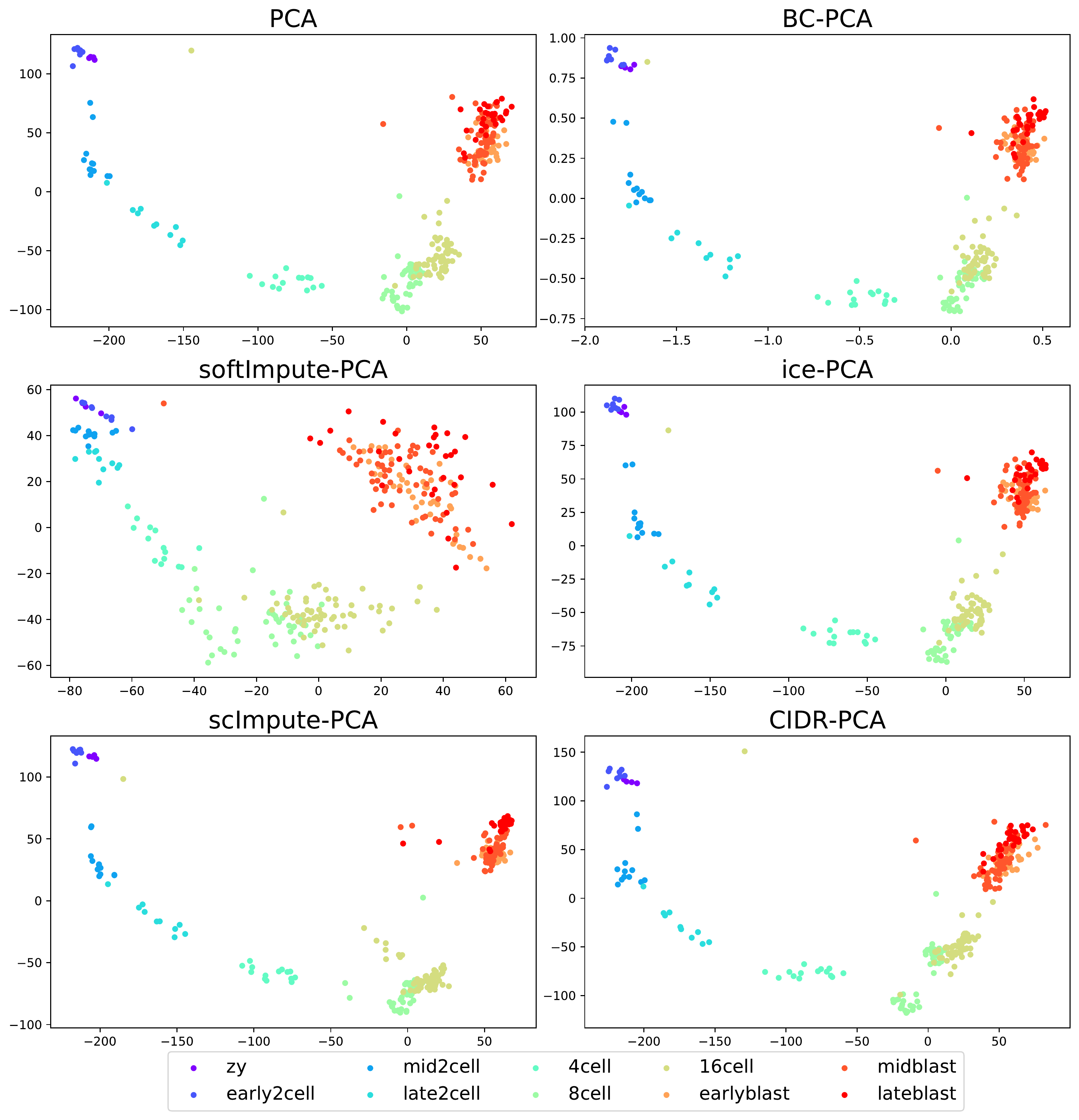}
    \caption{Visualisation of the Deng dataset obtained by PCA and its variants integrated with the bias correction or imputations.}
    \label{figs:Deng_PCA_vis}
\end{figure}

\begin{figure}[htbp]
    \centering
    \includegraphics[width=1\linewidth]{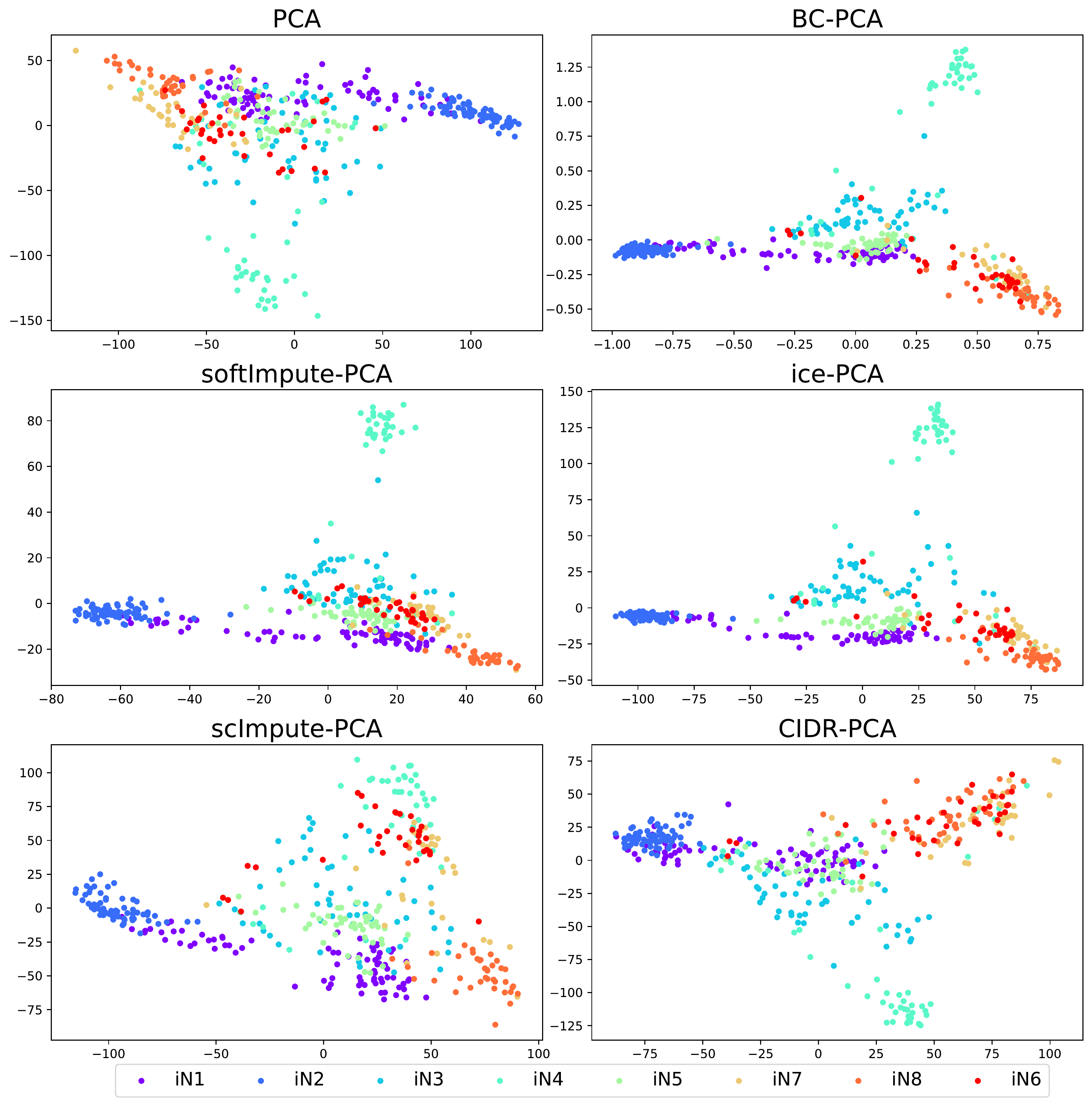}
    \caption{Visualisation of the Treutlein dataset obtained by PCA and its variants integrated with the bias correction or imputations.}
    \label{figs:Treutlein_PCA_vis}
\end{figure}

\begin{figure}[htbp]
    \centering
    \vspace*{-0.13cm}\includegraphics[width=1\linewidth]{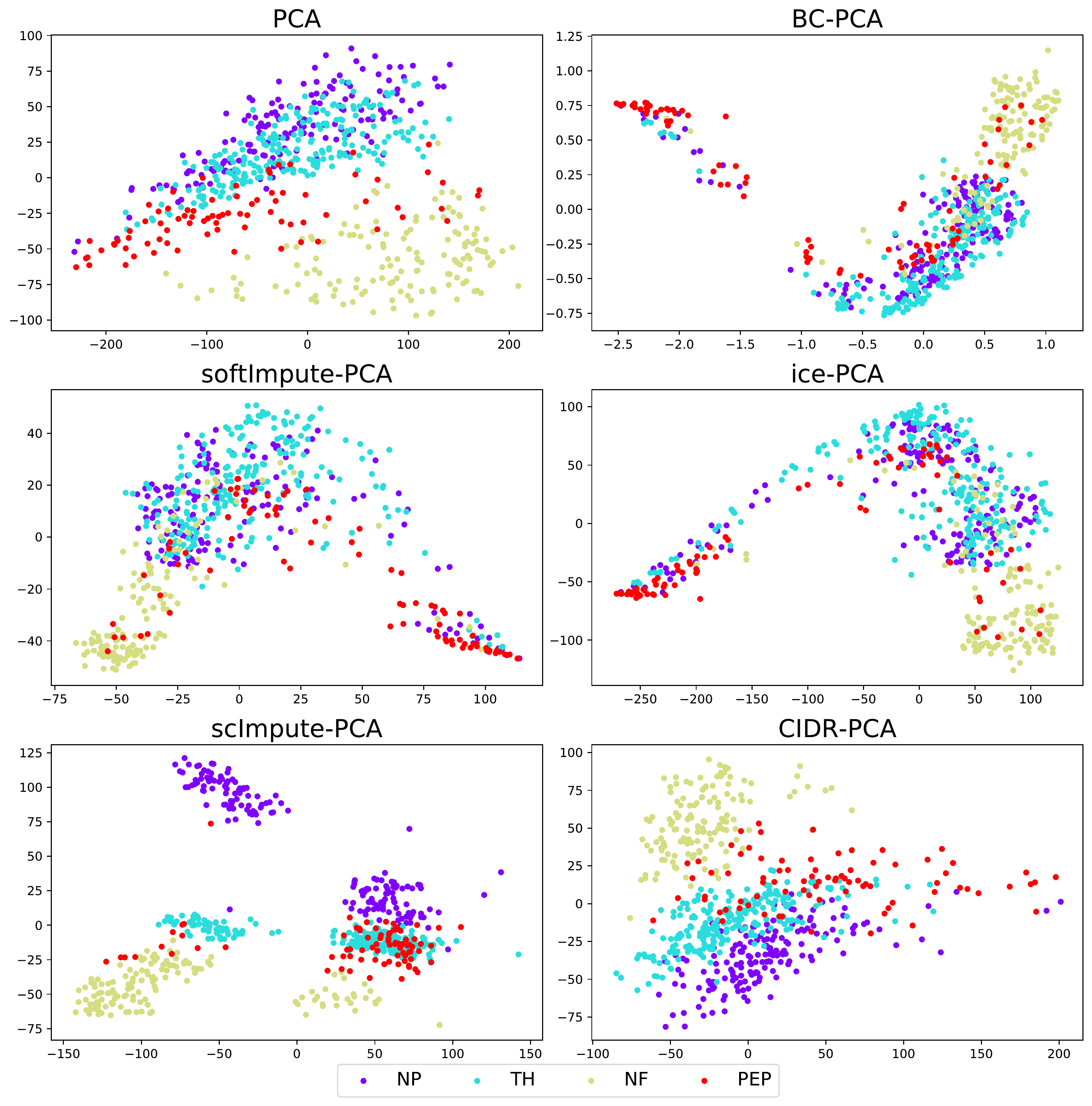}
    \caption{Visualisation of the Usoskin dataset obtained by PCA and its variants integrated with the bias correction or imputations.}
    \label{figs:Usoskin_PCA_vis}
\end{figure}

\begin{figure}[htbp]
    \centering
    \includegraphics[width=1\linewidth]{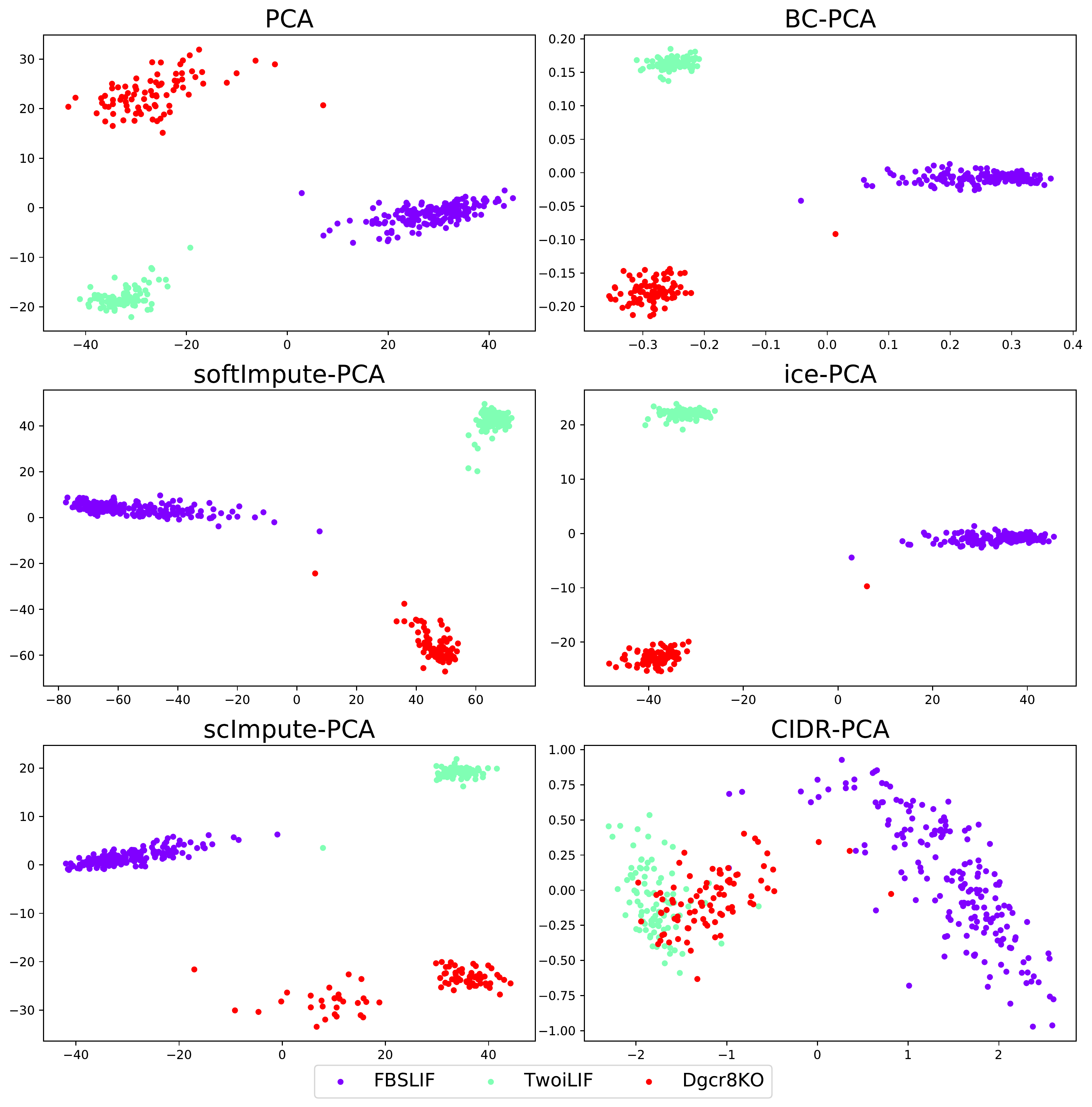}
    \caption{Visualisation of the Kumar dataset obtained by PCA and its variants integrated with the bias correction or imputations.}
    \label{figs:Kumar_PCA_vis}
\end{figure}

\clearpage
\subsubsection{Visualisation of GPLVM results on the real datasets}
\begin{figure}[H]
    \centering
    \includegraphics[width=1\linewidth]{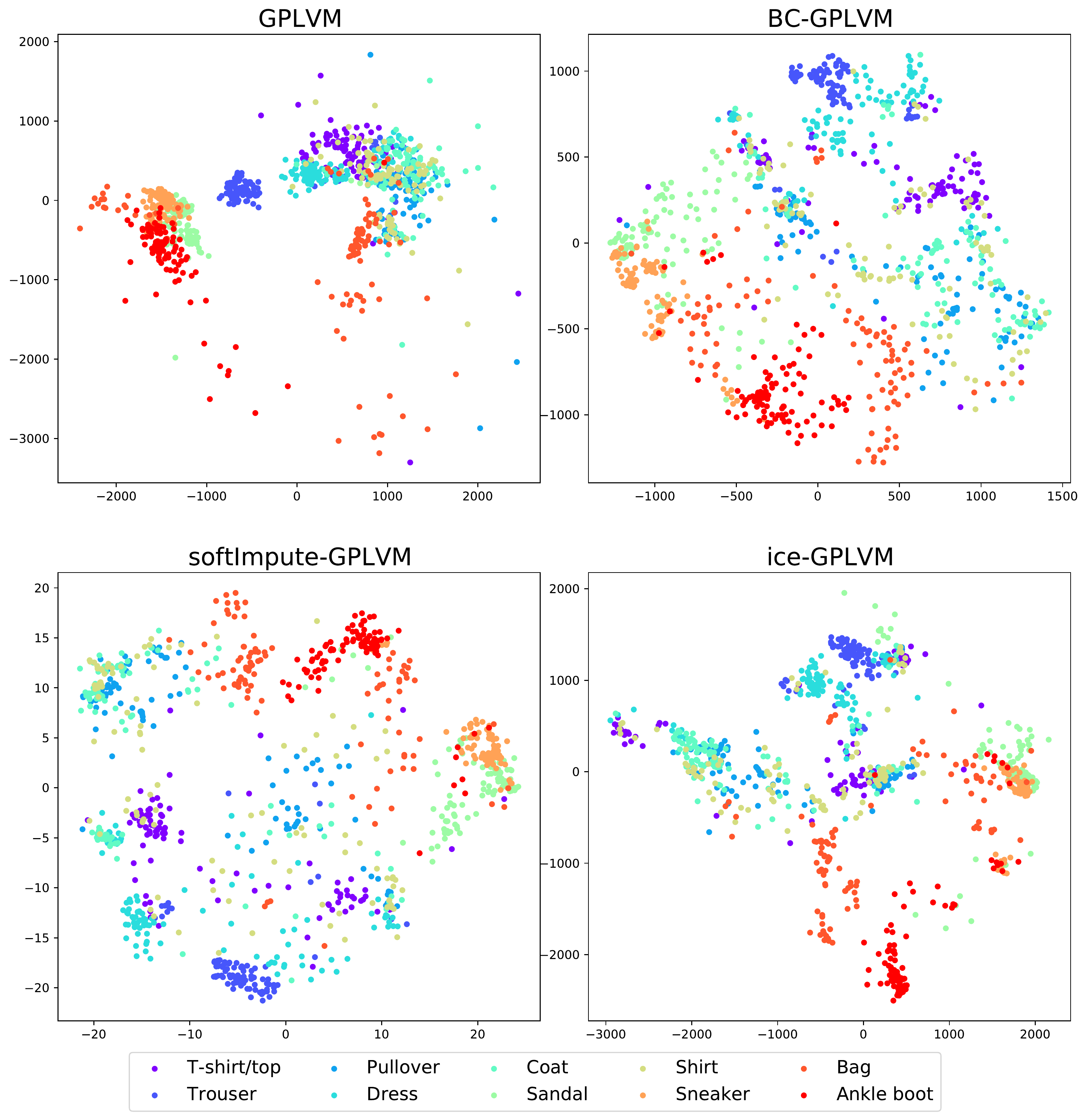}
    \caption{Visualisation of the fashion MNIST dataset obtained by GPLVM and its variants integrated with the bias correction or imputations.}
    \label{figs:fashion_MNIST_GPLVM_vis}
\end{figure}

\begin{figure}[H]
    \centering
    \includegraphics[width=1\linewidth]{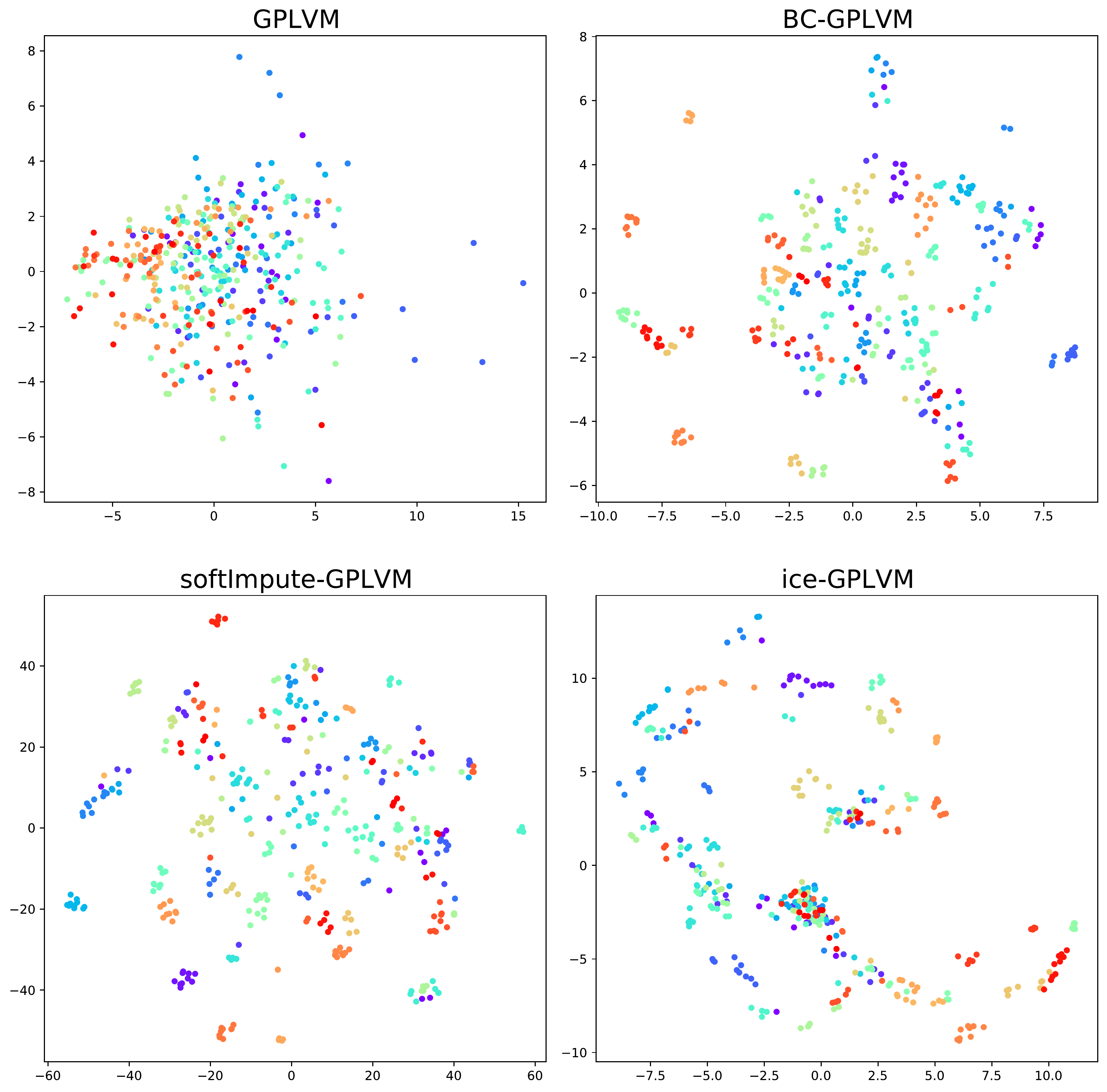}
    \caption{Visualisation of the Olivetti faces dataset obtained by GPLVM and its variants integrated with the bias correction or imputations.}
    \label{figs:Olivetti_faces_GPLVM_vis}
\end{figure}

\begin{figure}[H]
    \centering
    \includegraphics[width=1\linewidth]{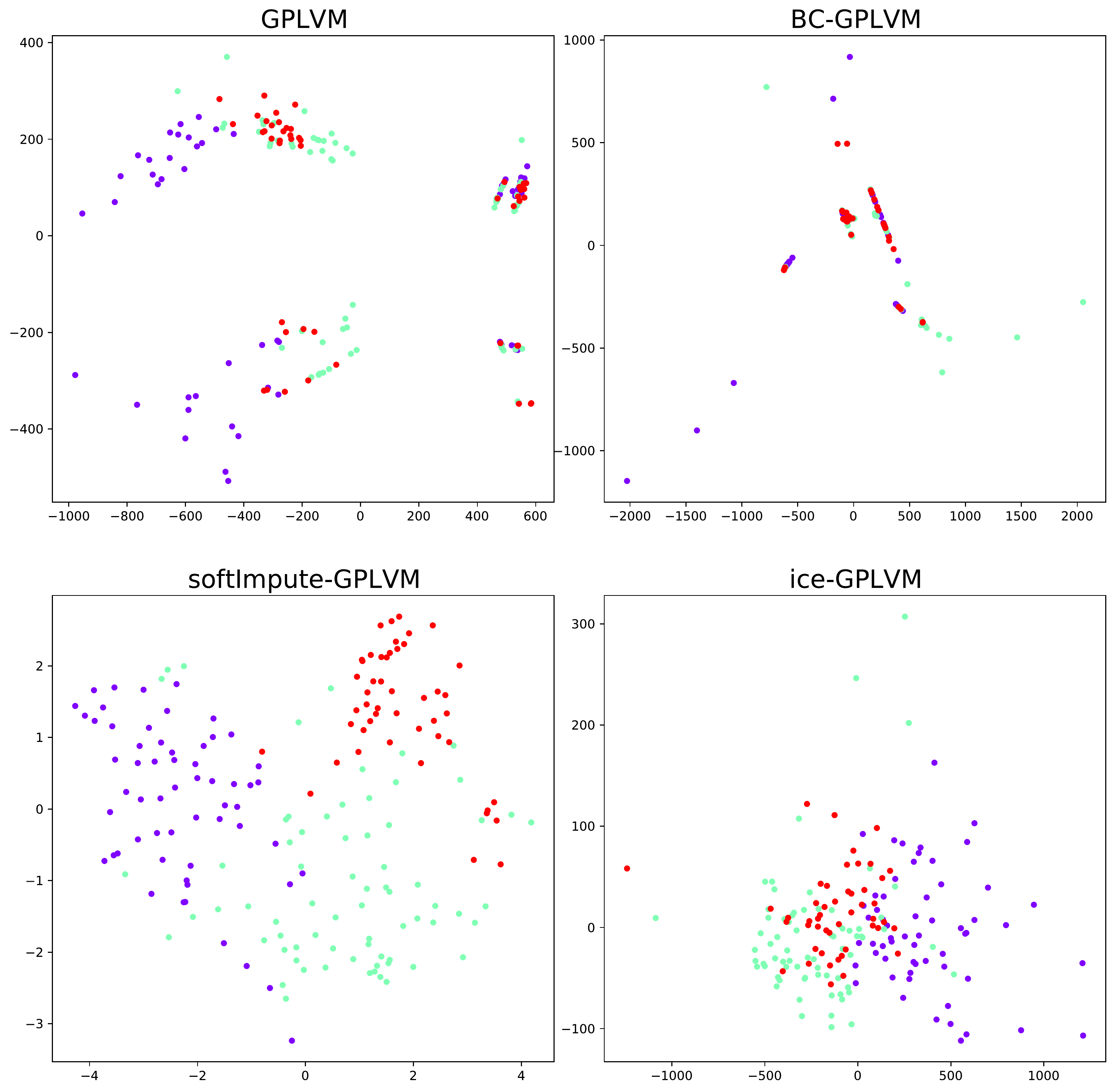}
    \caption{Visualisation of the wine dataset obtained by GPLVM and its variants integrated with the bias correction or imputations.}
    \label{figs:wine_GPLVM_vis}
\end{figure}

\begin{figure}[h]
    \centering
    \includegraphics[width=1\linewidth]{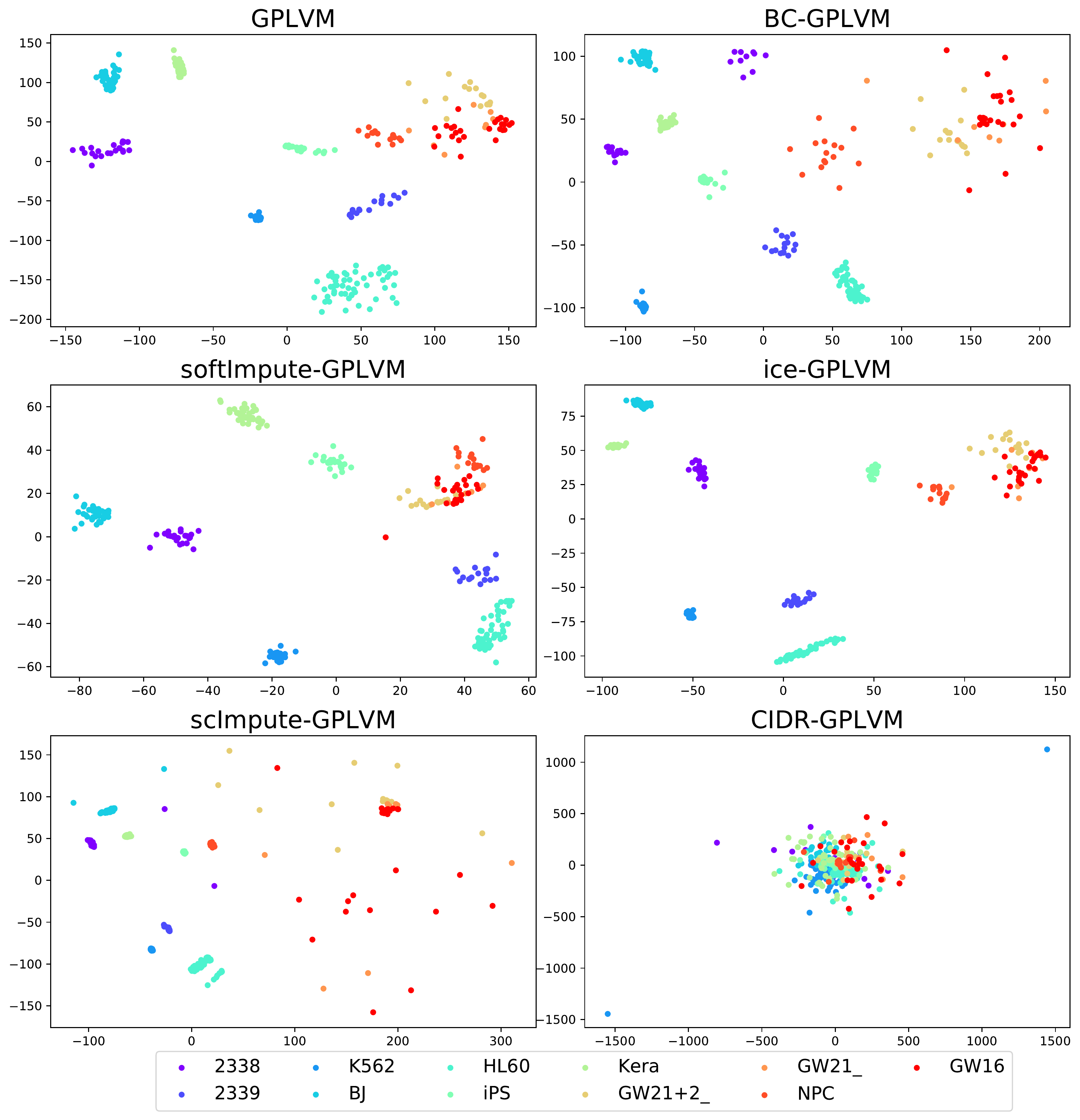}
    \caption{Visualisation of the Pollen dataset obtained by GPLVM and its variants integrated with the bias correction or imputations.}
    \label{figs:Pollen_GPLVM_vis}
\end{figure}

\begin{figure}[h]
    \centering
    \includegraphics[width=1\linewidth]{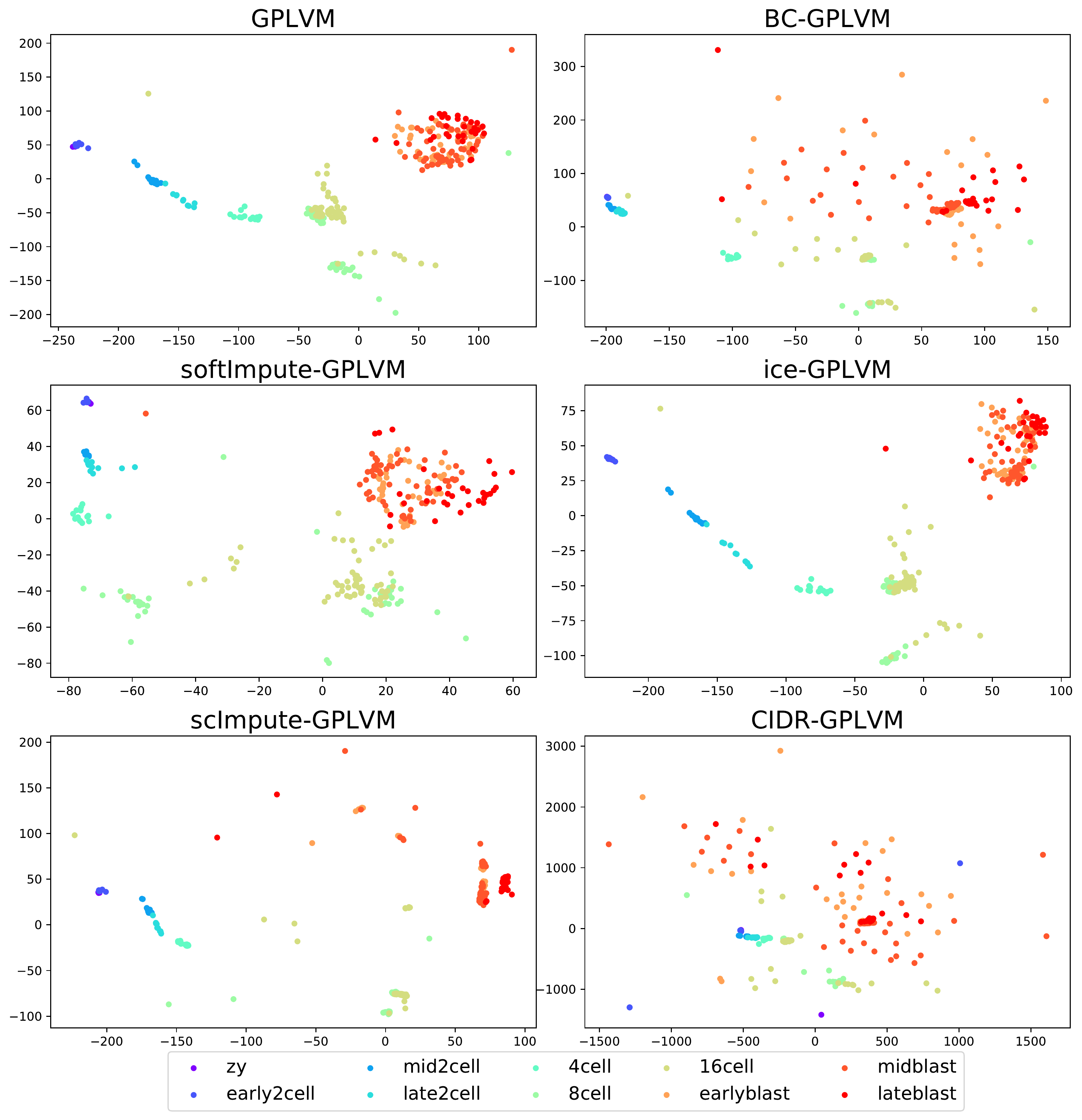}
    \caption{Visualisation of the Deng dataset obtained by GPLVM and its variants integrated with the bias correction or imputations.}
    \label{figs:Deng_GPLVM_vis}
\end{figure}

\begin{figure}[h]
    \centering
    \includegraphics[width=1\linewidth]{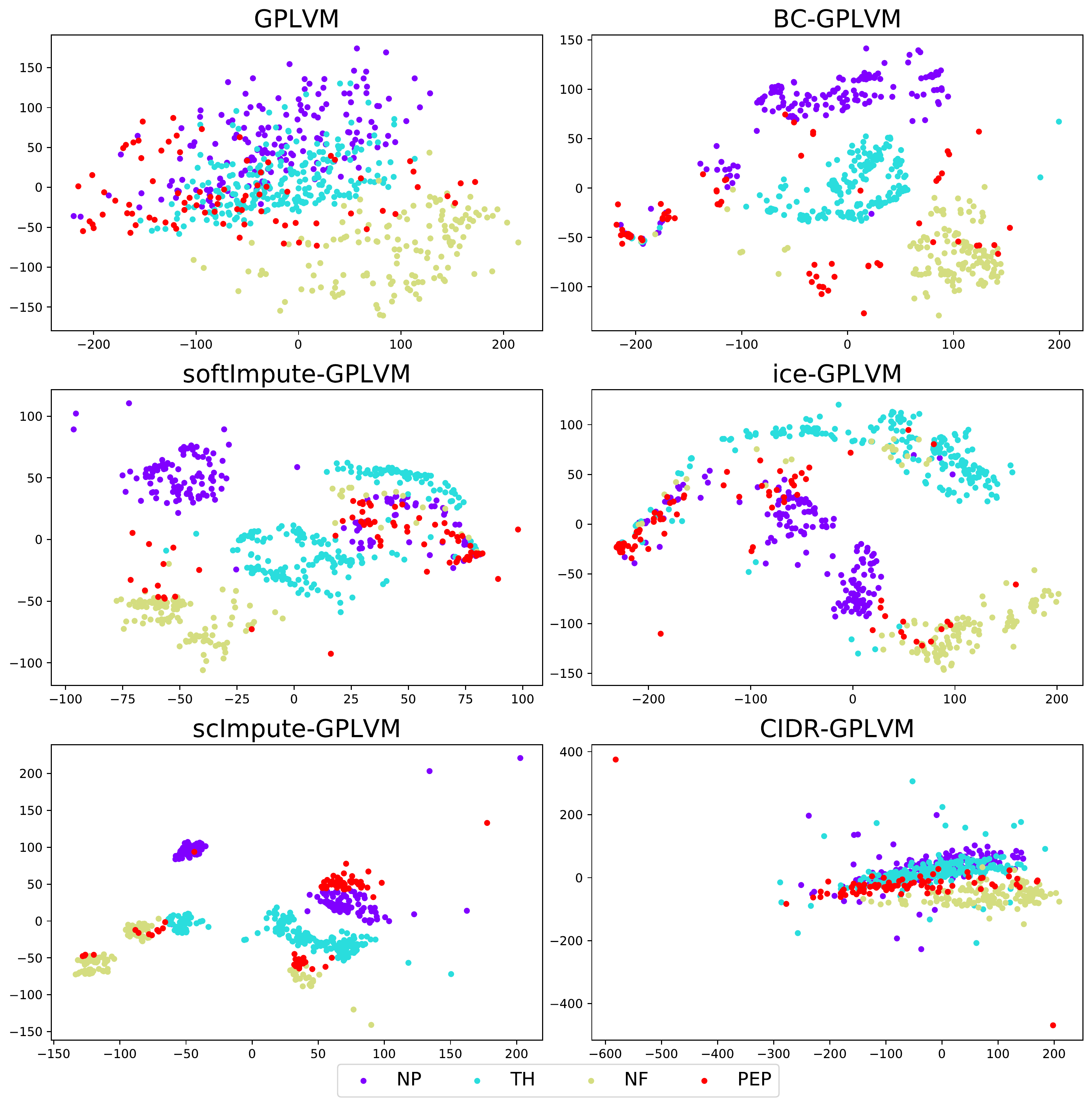}
    \caption{Visualisation of the Usoskin dataset obtained by GPLVM and its variants integrated with the bias correction or imputations.}
    \label{figs:Usoskin_GPLVM_vis}
\end{figure}

\newpage
\begin{figure}[t]
    \centering
    \vspace*{0.12cm}\includegraphics[width=1\linewidth]{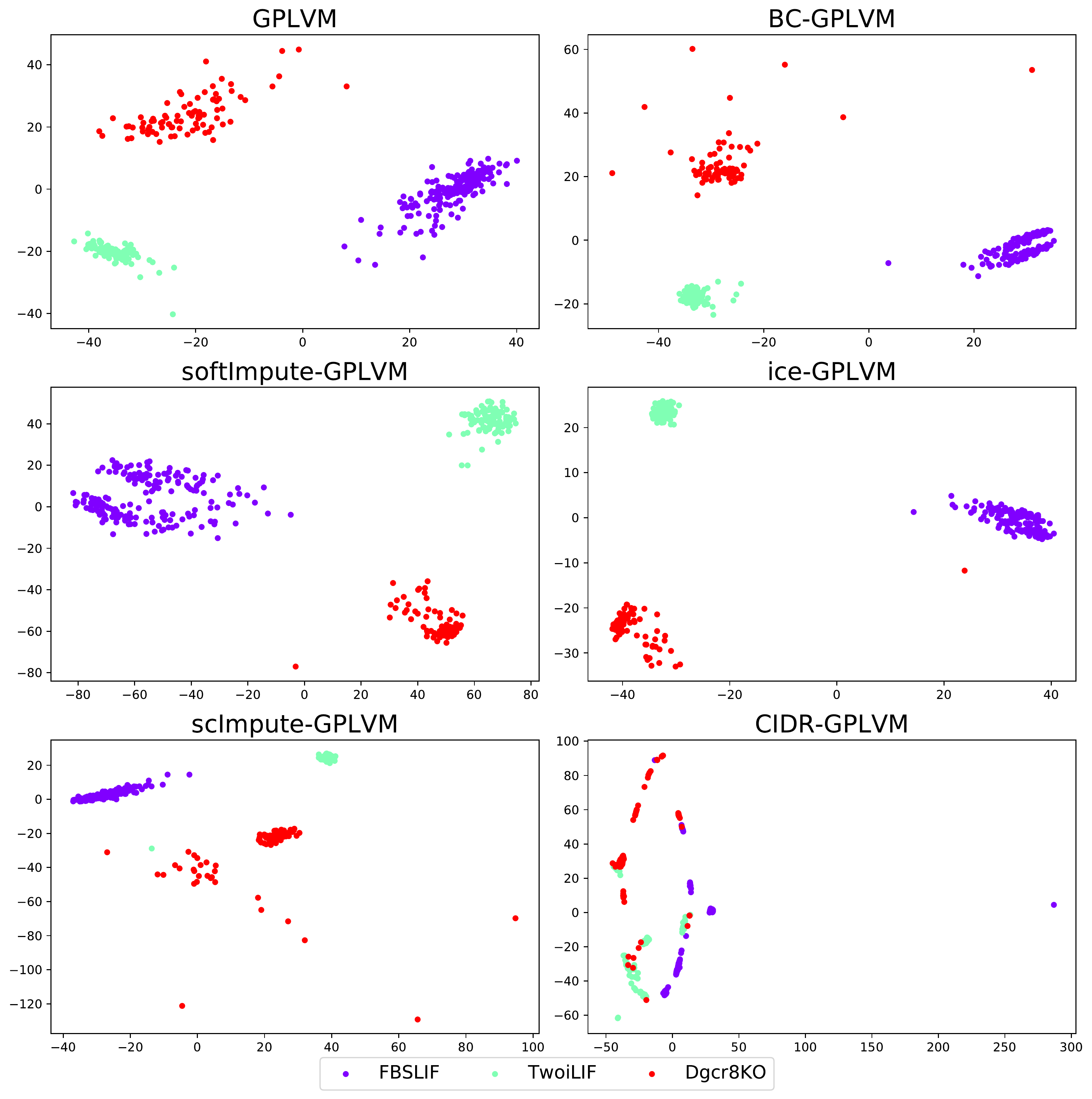}
    \caption{Visualisation of the Kumar dataset obtained by GPLVM and its variants integrated with the bias correction or imputations.}
    \label{figs:Kumar_GPLVM_vis}
\end{figure}

\begin{figure}[h]
    \centering
    \includegraphics[width=1\linewidth]{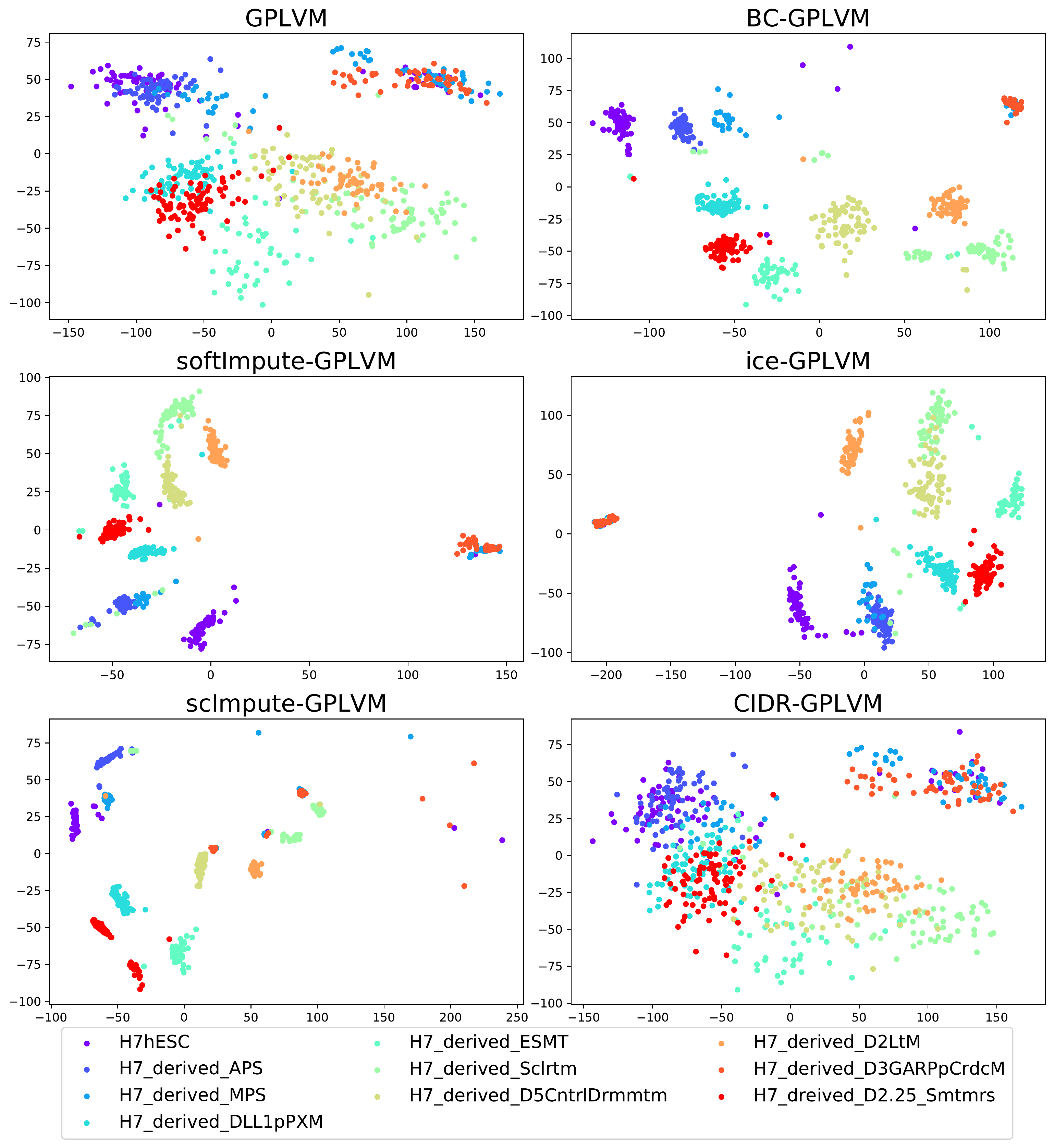}
    \caption{Visualisation of the Koh dataset obtained by GPLVM and its variants integrated with the bias correction or imputations.}
    \label{figs:Koh_GPLVM_vis}
\end{figure}

\clearpage
\subsubsection{Visualisation of tSNE results on the real datasets}
\begin{figure}[htbp]
    \centering
    \includegraphics[width=1\linewidth]{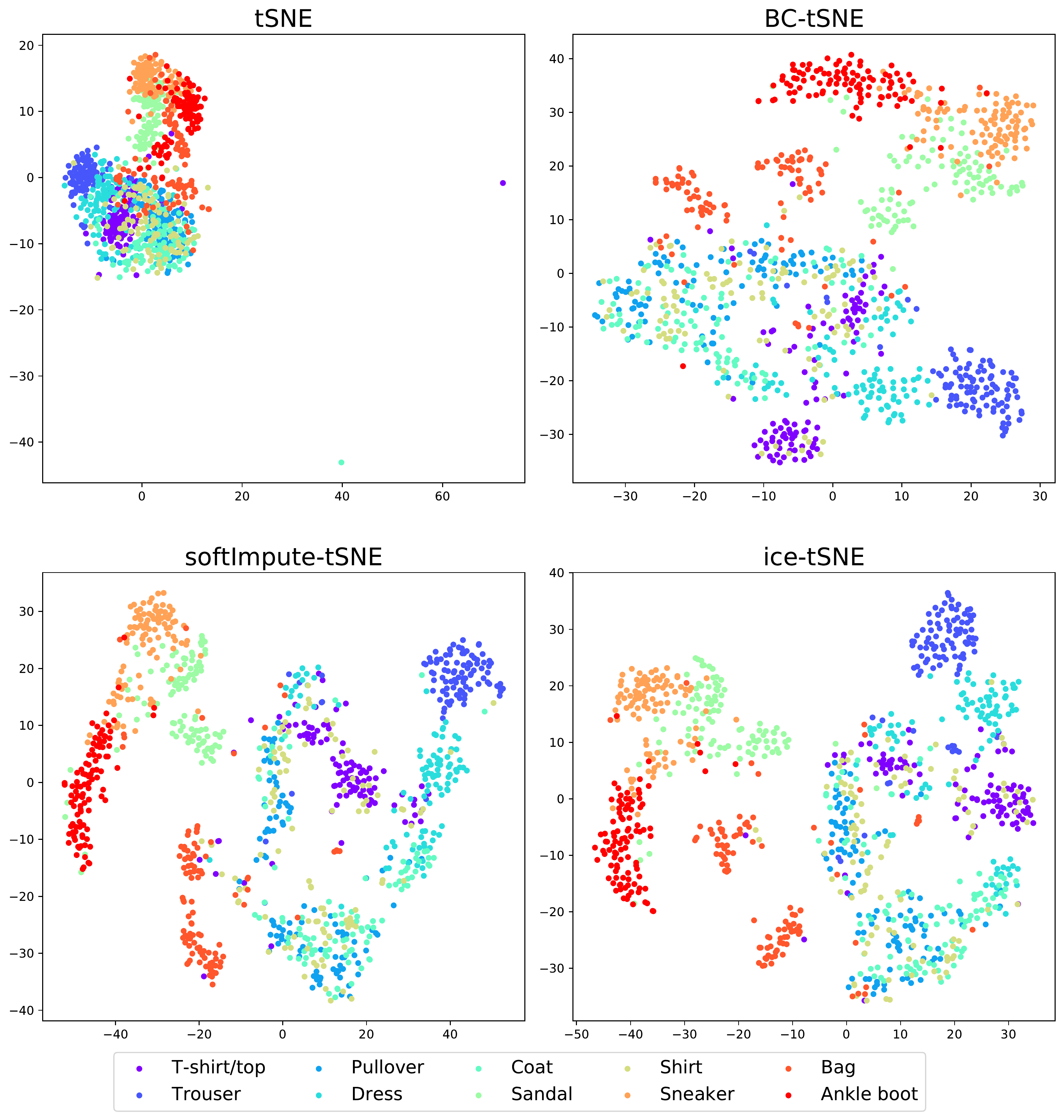}
    \caption{Visualisation of the fashion MNIST dataset obtained by tSNE and its variants integrated with the bias correction or imputations.}
    \label{figs:fashion_MNIST_tSNE_vis}
\end{figure}

\begin{figure}[H]
    \centering
    \includegraphics[width=1\linewidth]{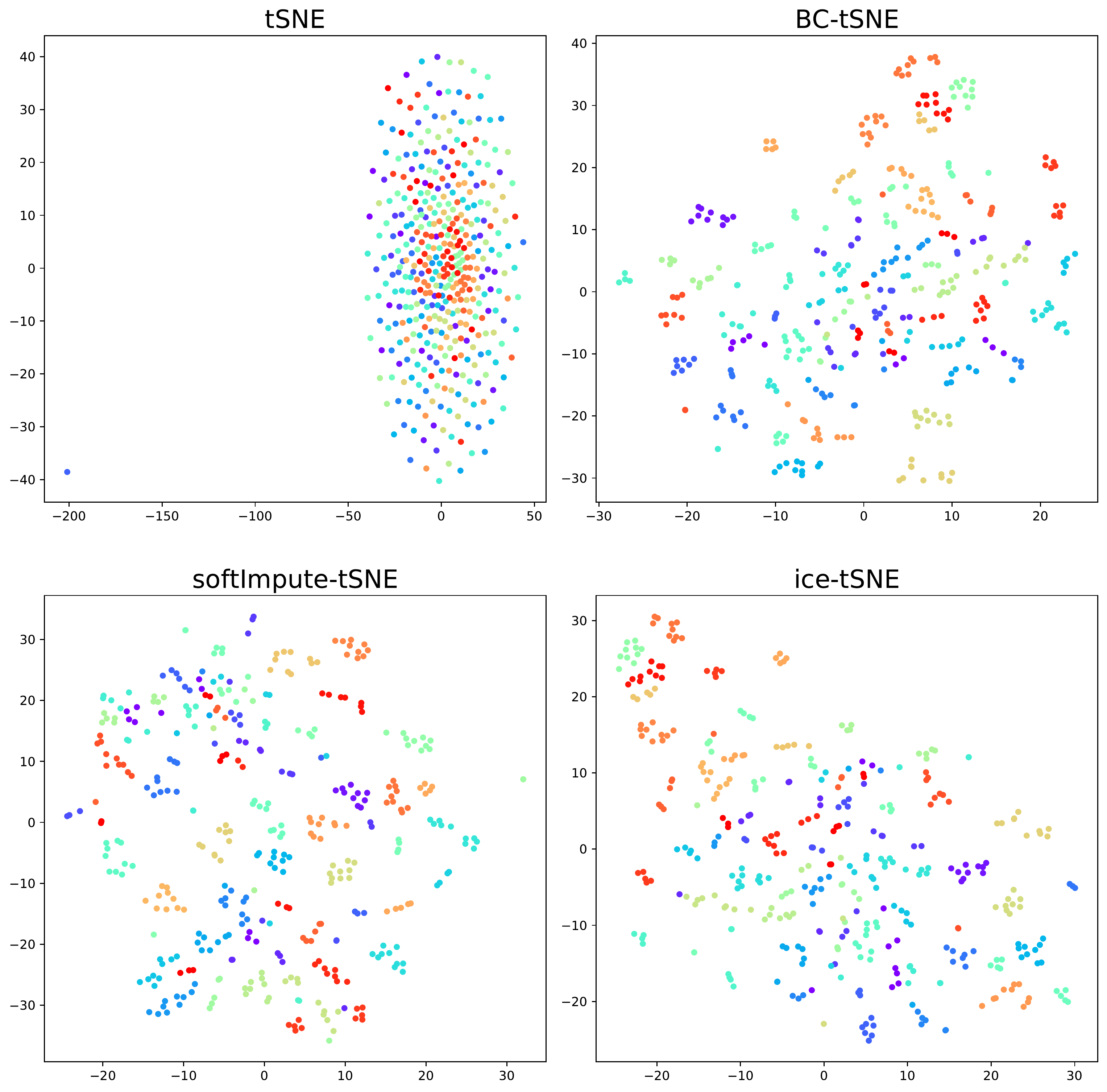}
    \caption{Visualisation of the Olivetti faces dataset obtained by tSNE and its variants integrated with the bias correction or imputations.}
    \label{figs:Olivetti_faces_tSNE_vis}
\end{figure}

\begin{figure}[H]
    \centering
    \includegraphics[width=1\linewidth]{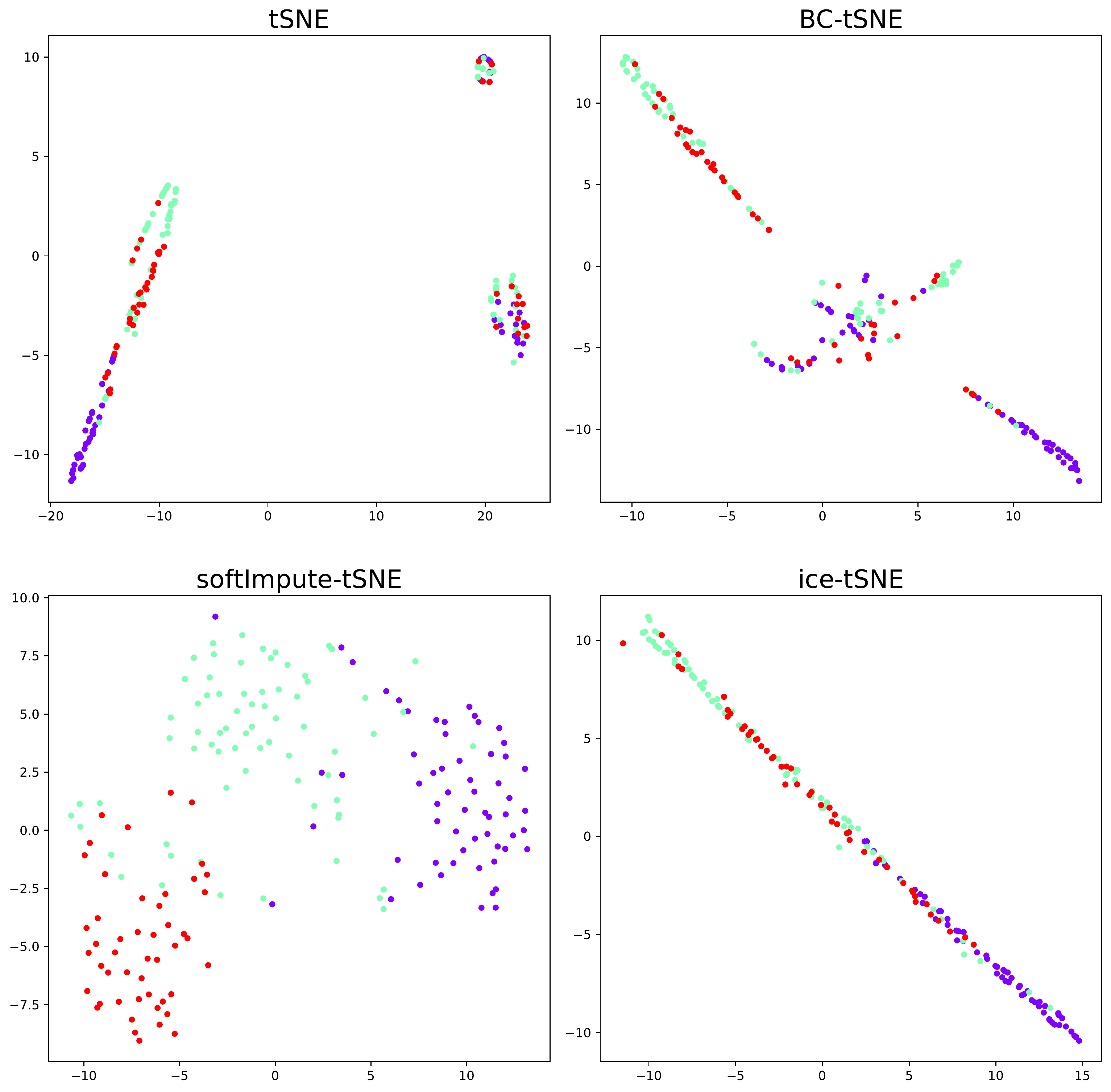}
    \caption{Visualisation of the wine dataset obtained by tSNE and its variants integrated with the bias correction or imputations.}
    \label{figs:wine_tSNE_vis}
\end{figure}

\clearpage
\begin{figure}[t]
    \centering
    \includegraphics[width=1\linewidth]{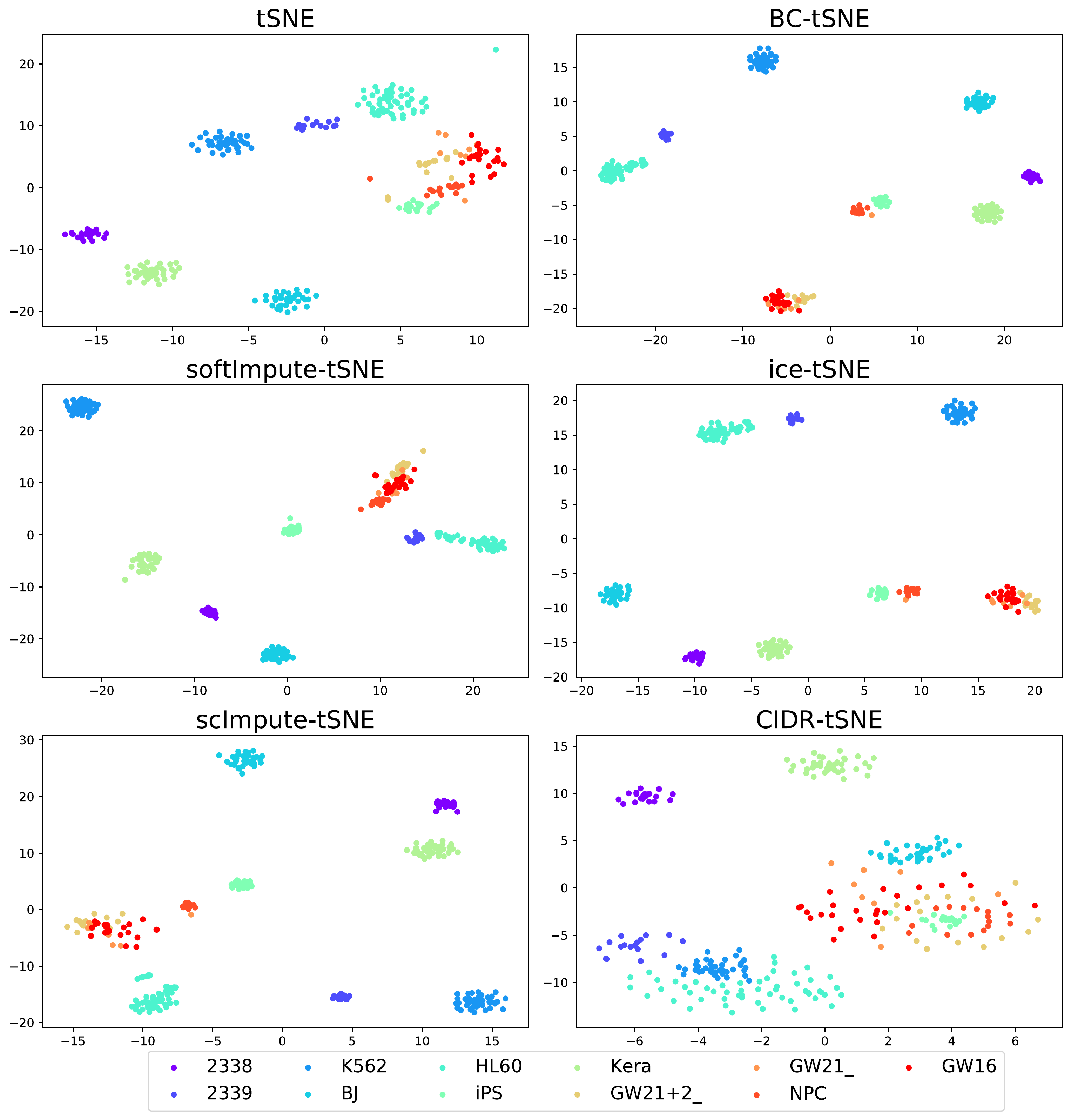}
    \caption{Visualisation of the Pollen dataset obtained by tSNE and its variants integrated with the bias correction or imputations.}
    \label{figs:Pollen_tSNE_vis}
\end{figure}

\begin{figure}[H]
    \centering
    \includegraphics[width=1\linewidth]{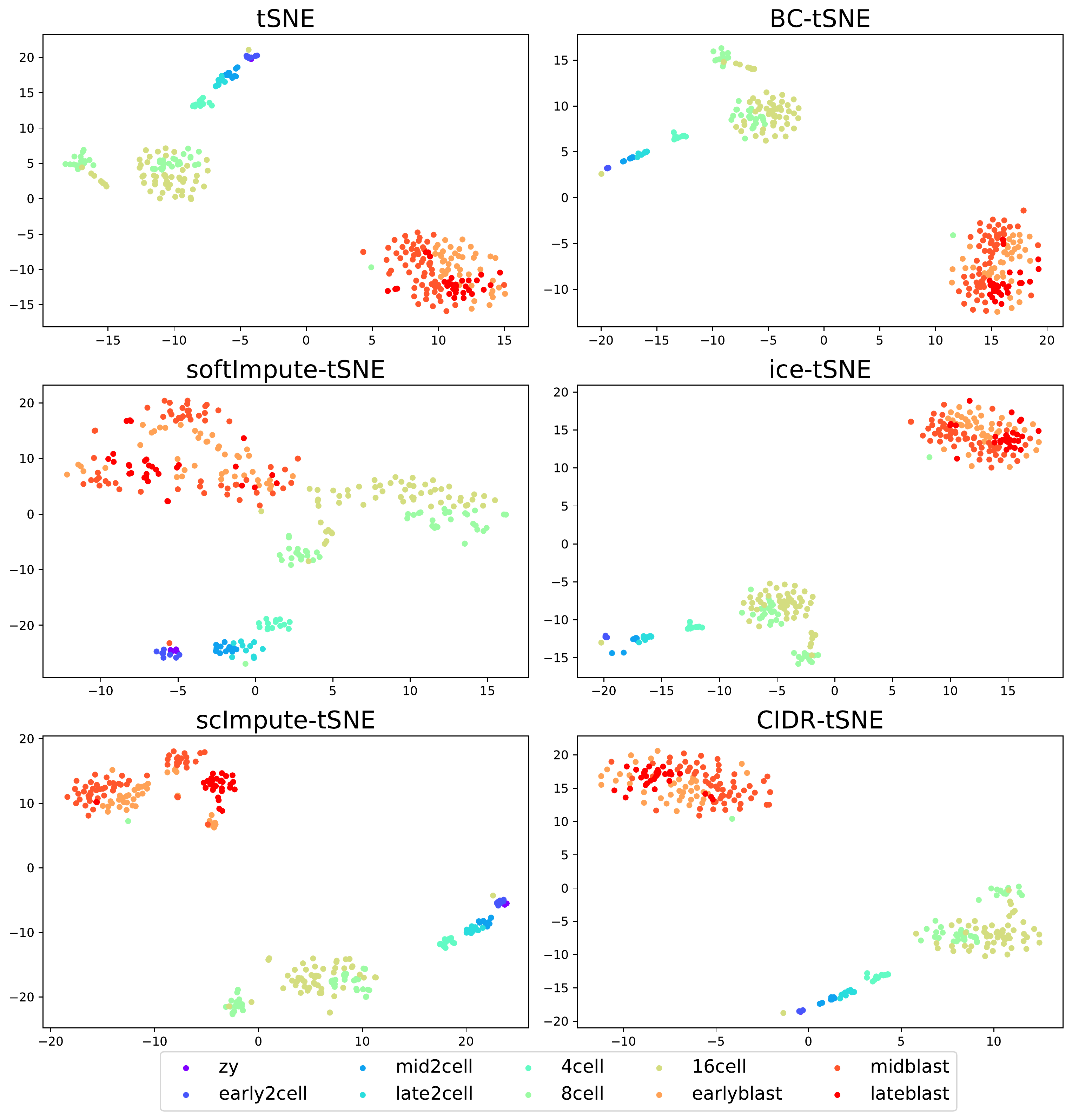}
    \caption{Visualisation of the Deng dataset obtained by tSNE and its variants integrated with the bias correction or imputations.}
    \label{figs:Deng_tSNE_vis}
\end{figure}

\begin{figure}[htbp]
    \centering
    \vspace*{-0.8cm}\includegraphics[width=1\linewidth]{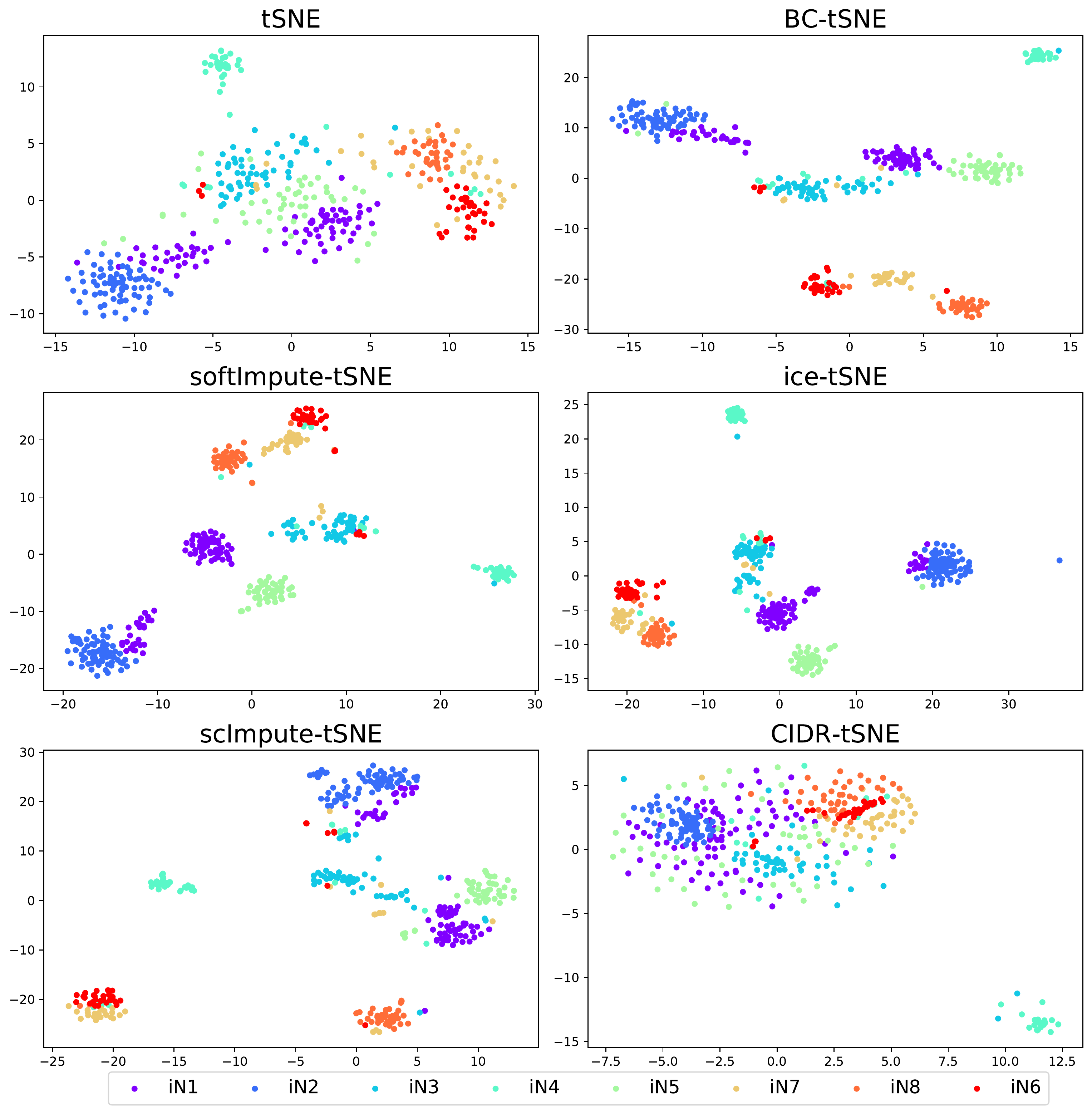}
    \caption{Visualisation of the Treutlein dataset obtained by tSNE and its variants integrated with the bias correction or imputations.}
    \label{figs:Treutlein_tSNE_vis}
\end{figure}

\begin{figure}[htbp]
    \centering
    \includegraphics[width=1\linewidth]{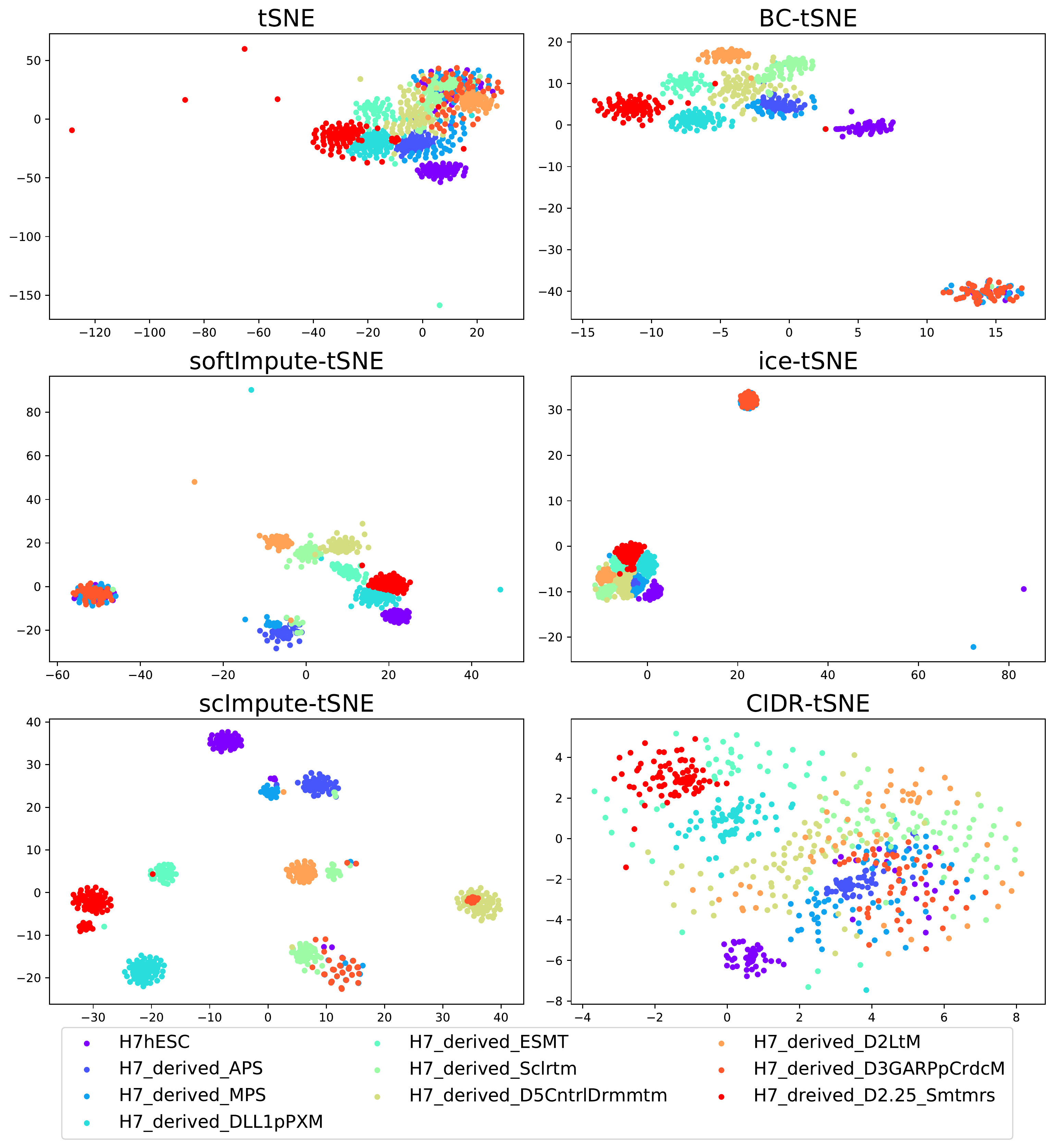}
    \caption{Visualisation of the Koh dataset obtained by tSNE and its variants integrated with the bias correction or imputations.}
    \label{figs:Koh_tSNE_vis}
\end{figure}

\begin{figure}[H]
    \centering
    \includegraphics[width=1\linewidth]{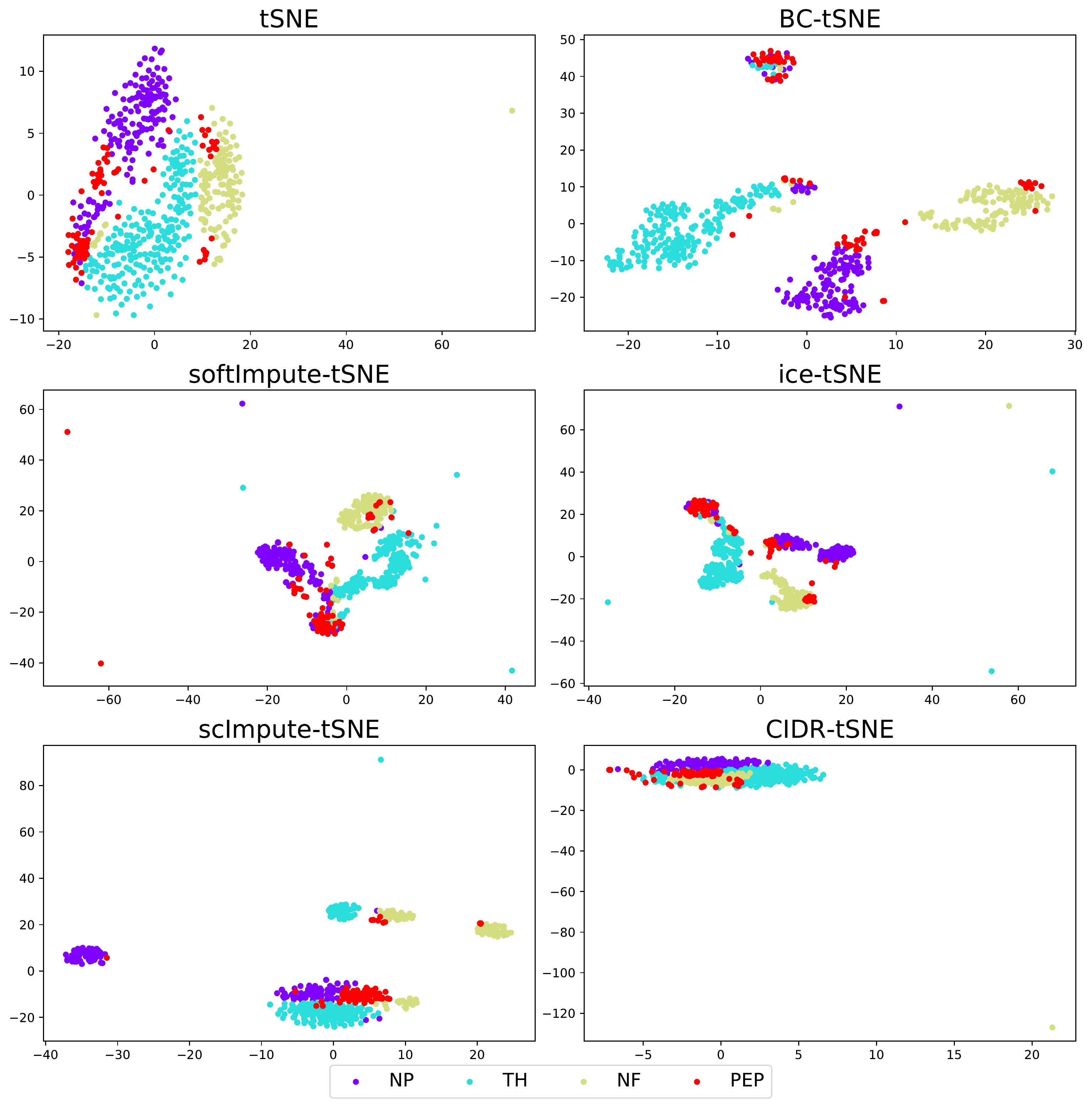}
    \caption{Visualisation of the Usoskin dataset obtained by tSNE and its variants integrated with the bias correction or imputations.}
    \label{figs:Usoskin_tSNE_vis}
\end{figure}

\begin{figure}[htbp]
    \centering
    \includegraphics[width=1\linewidth]{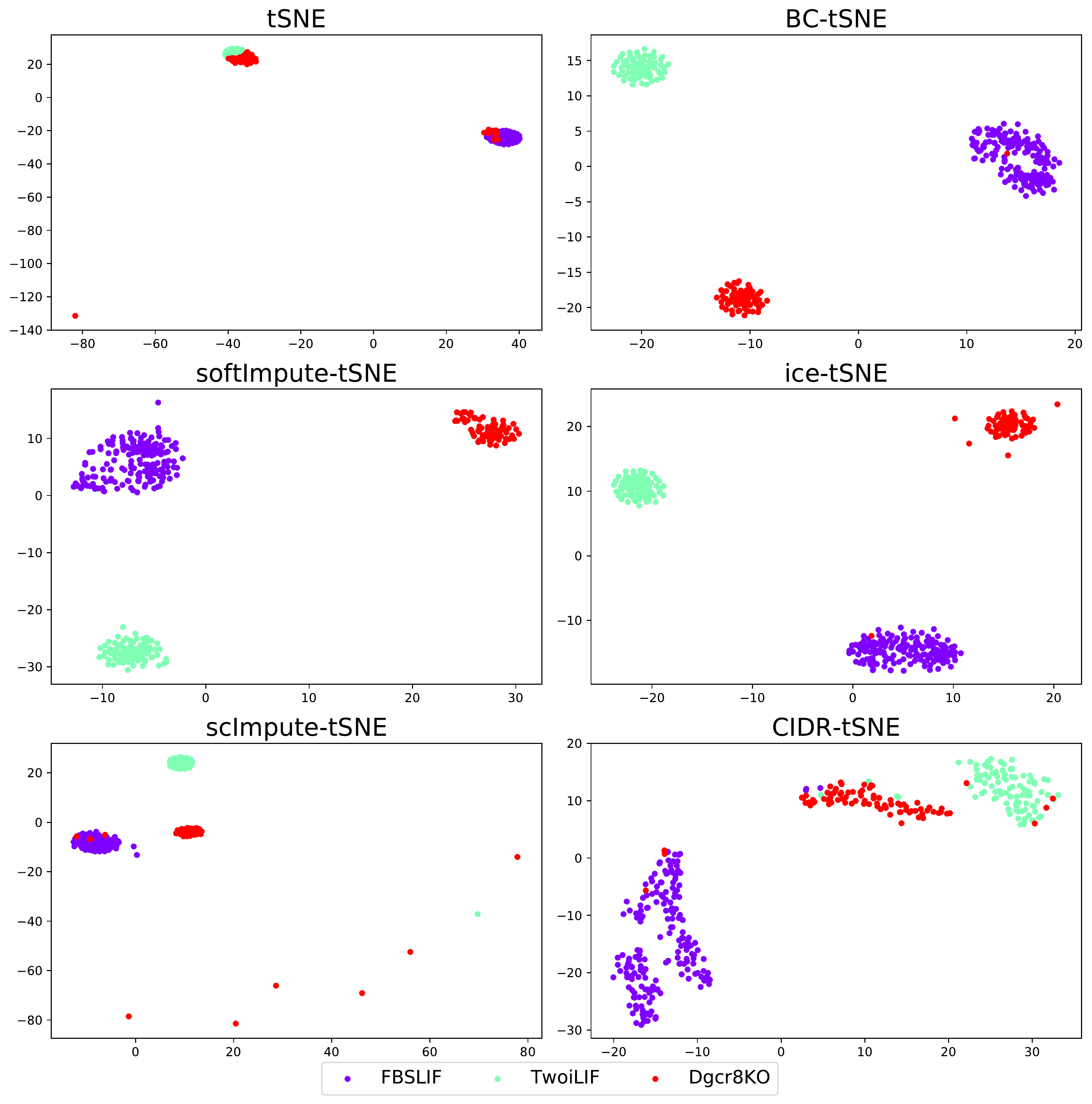}
    \caption{Visualisation of the Kumar dataset obtained by tSNE and its variants integrated with the bias correction or imputations.}
    \label{figs:Kumar_tSNE_vis}
\end{figure}

\subsubsection{Visualisation of UMAP results on the real datasets}

\begin{figure}[H]
    \centering
    \includegraphics[width=1\linewidth]{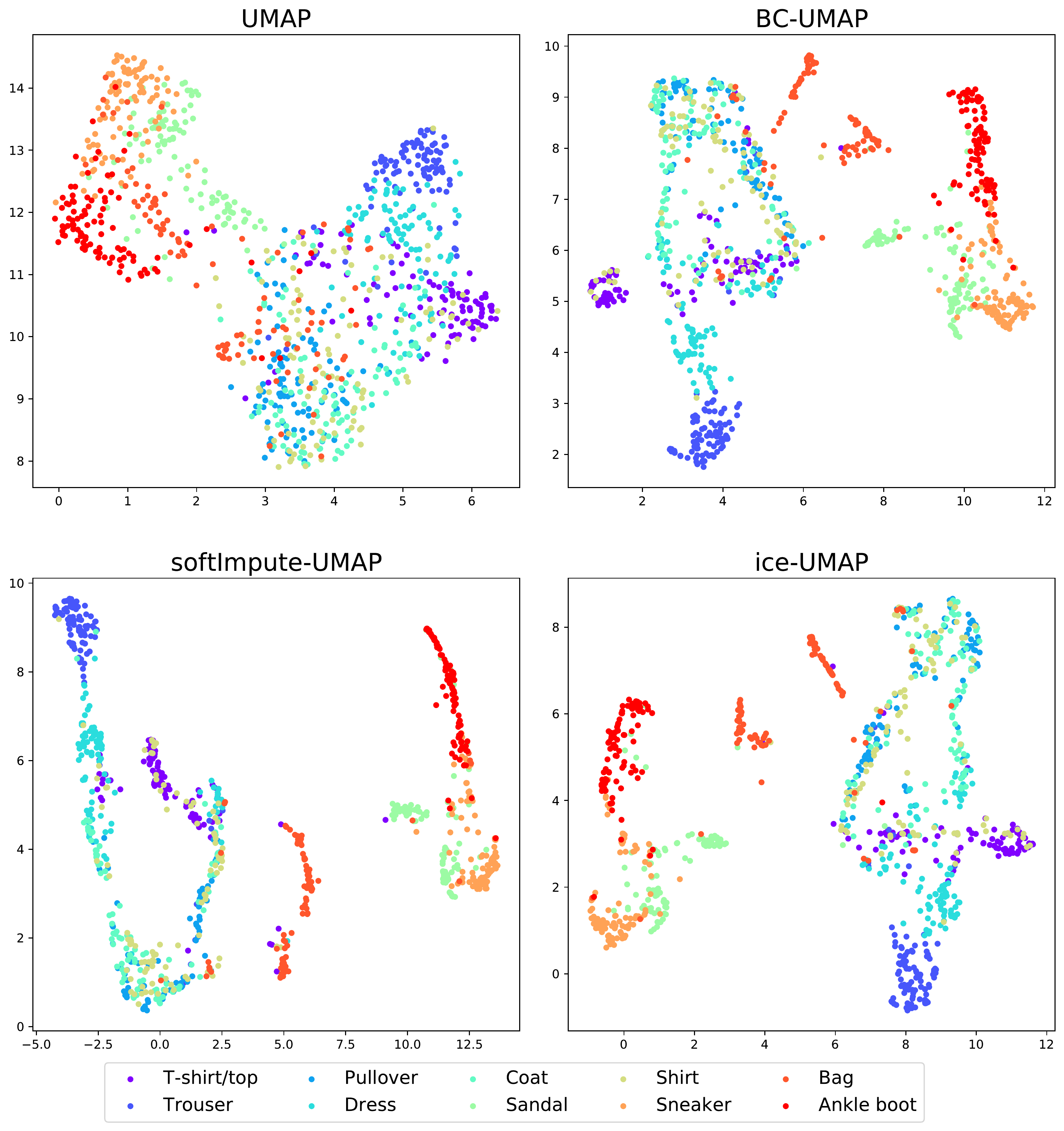}
    \caption{Visualisation of the fashion MNIST dataset obtained by UMAP and its variants integrated with the bias correction or imputations.}
    \label{figs:fashion_MNIST_UMAP_vis}
\end{figure}

\begin{figure}[H]
    \centering
    \includegraphics[width=1\linewidth]{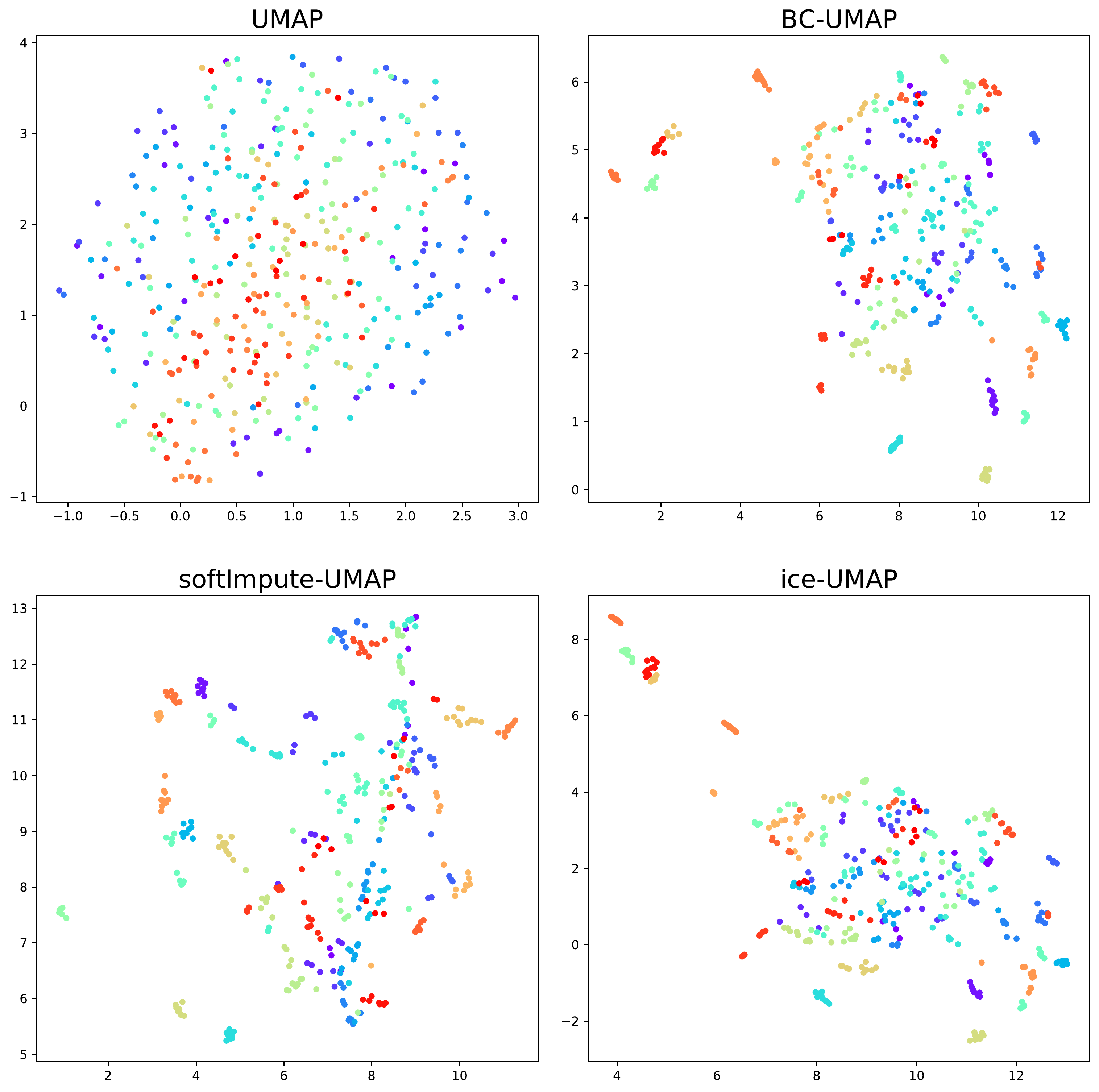}
    \caption{Visualisation of the Olivetti faces dataset obtained by UMAP and its variants integrated with the bias correction or imputations.}
    \label{figs:Olivetti_faces_UMAP_vis}
\end{figure}

\begin{figure}[htbp]
    \centering
    \vspace*{-0.6cm}\includegraphics[width=1\linewidth]{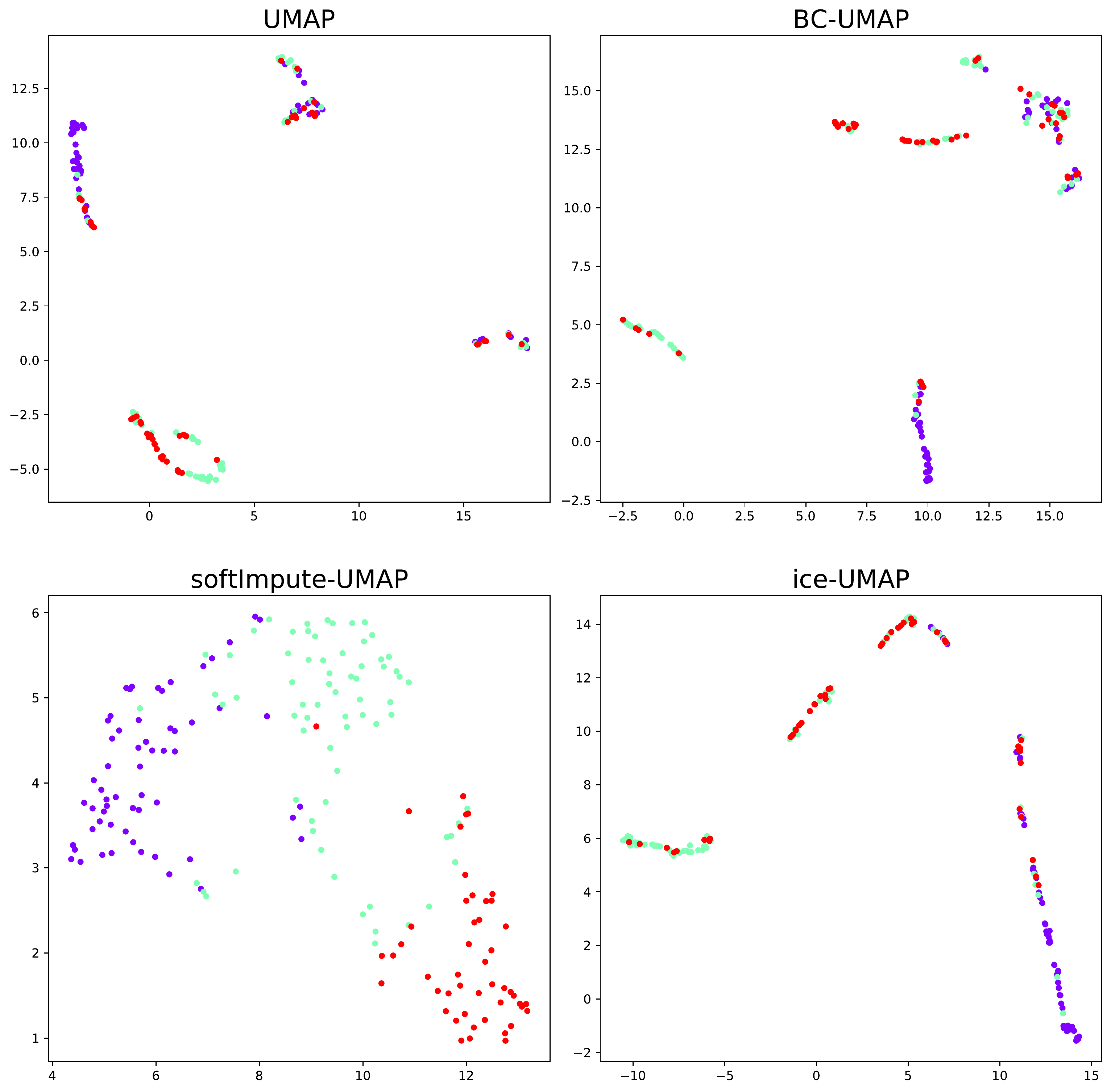}
    \caption{Visualisation of the wine dataset obtained by UMAP and its variants integrated with the bias correction or imputations.}
    \label{figs:wine_UMAP_vis}
\end{figure}

\begin{figure}[htbp]
    \centering
    \vspace*{-1cm}\includegraphics[width=1\linewidth]{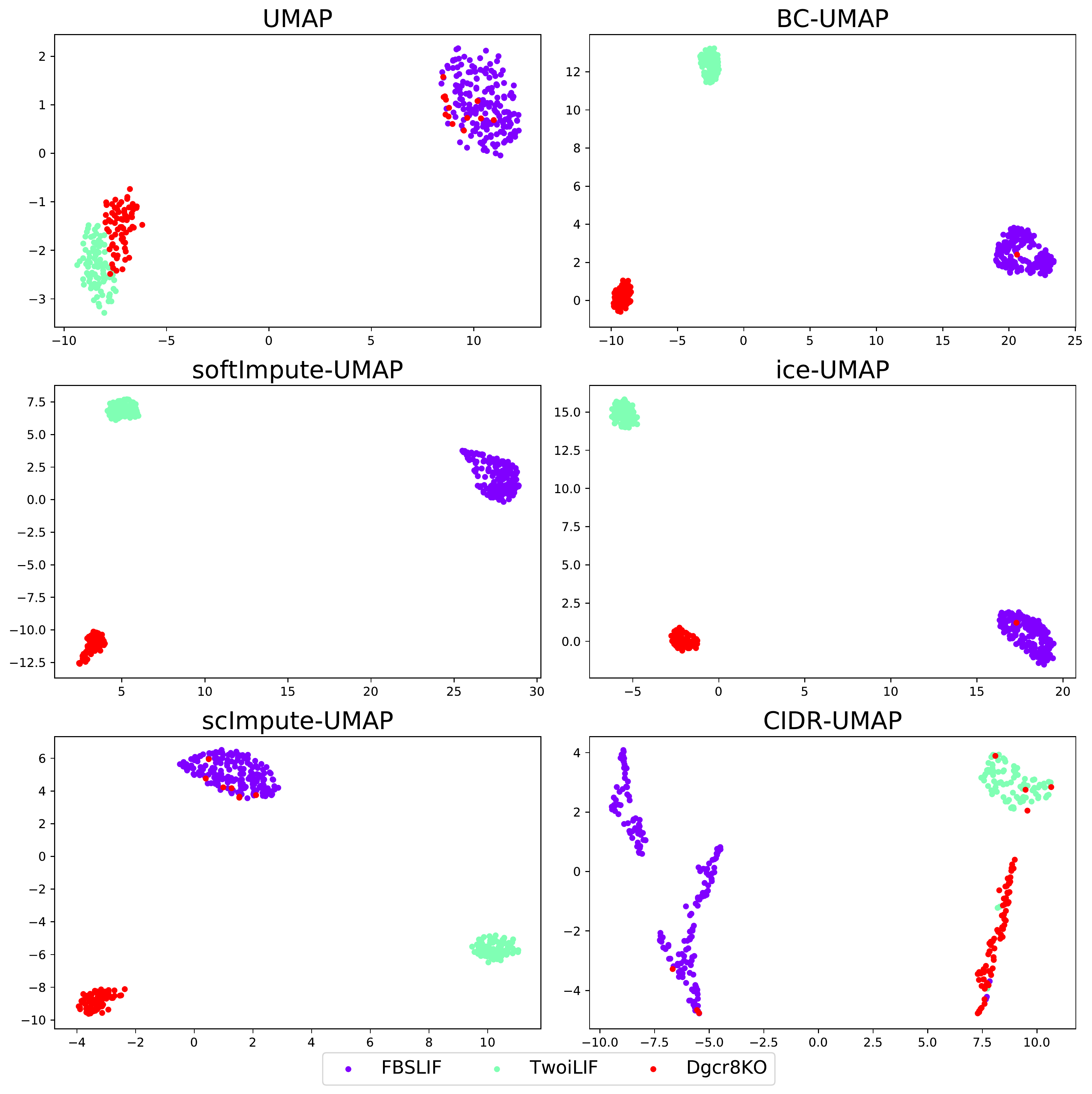}
    \caption{Visualisation of the Kumar dataset obtained by UMAP and its variants integrated with the bias correction or imputations.}
    \label{figs:Kumar_UMAP_vis}
\end{figure}

\begin{figure}[htbp]
    \centering
     \includegraphics[width=1\linewidth]{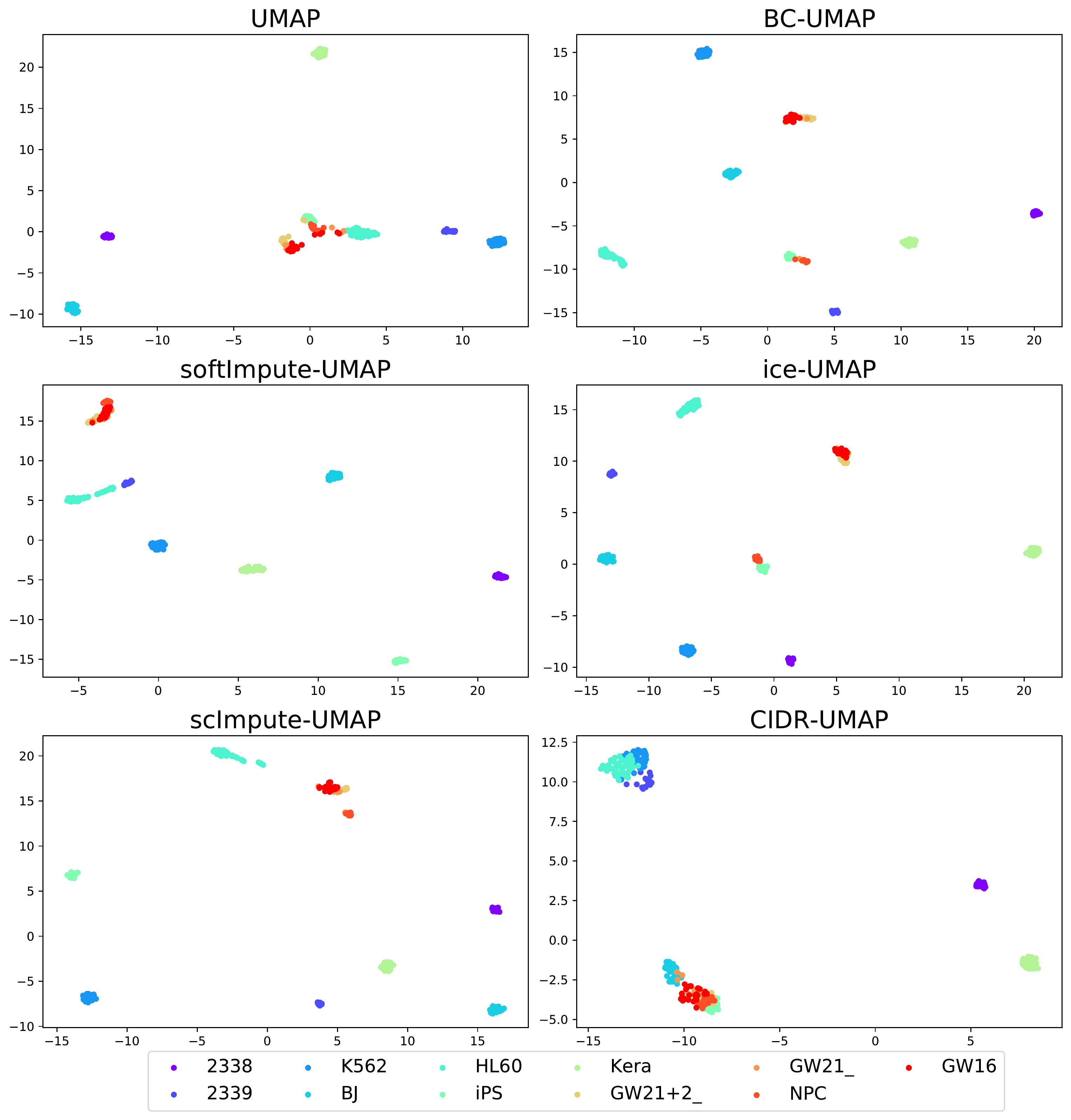}
    \caption{Visualisation of the Pollen dataset obtained by UMAP and its variants integrated with the bias correction or imputations.}
    \label{figs:Pollen_UMAP_vis}
\end{figure}

\begin{figure}[htbp]
    \centering
    \includegraphics[width=1\linewidth]{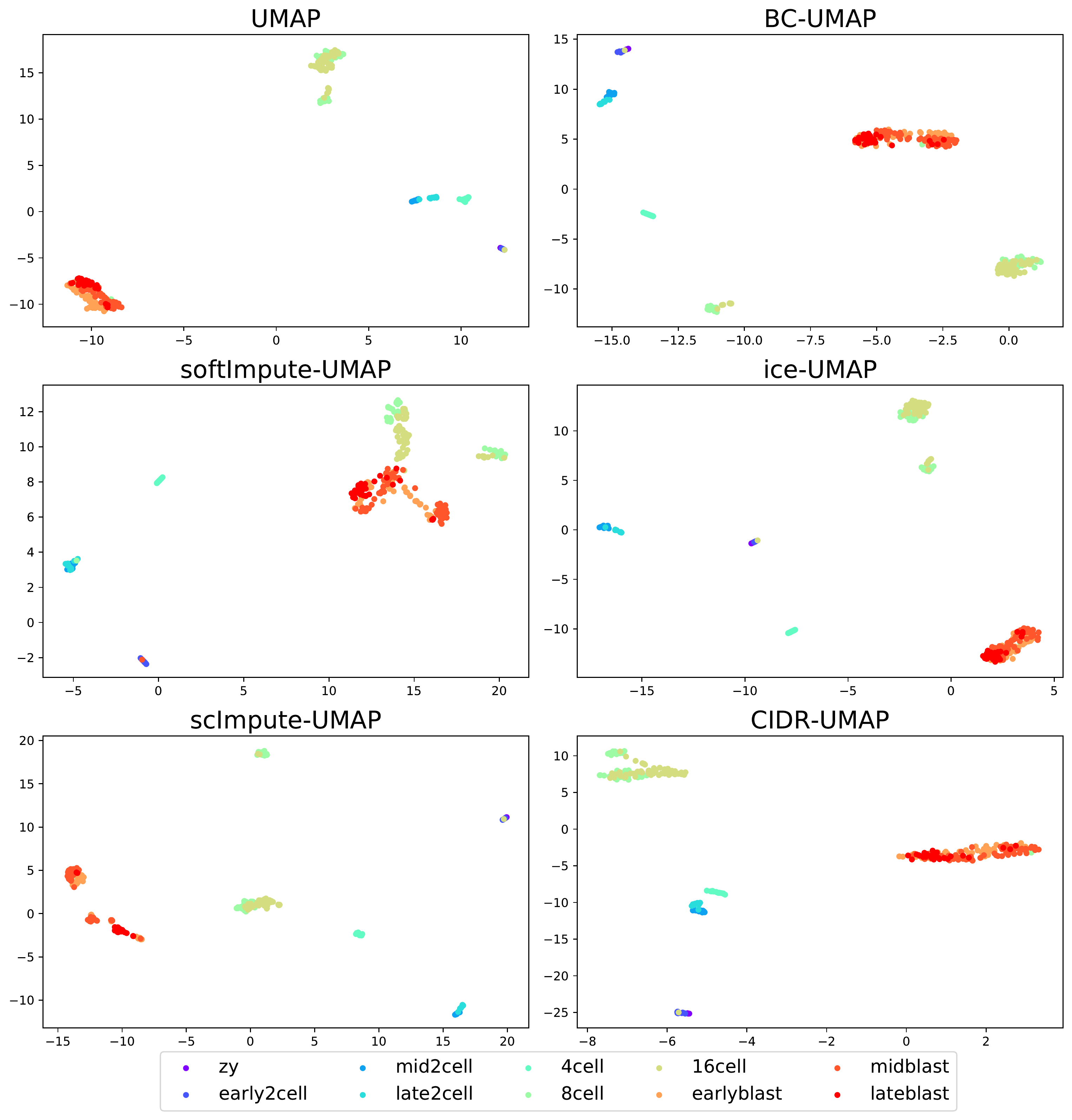}
    \caption{Visualisation of the Deng dataset obtained by UMAP and its variants integrated with the bias correction or imputations.}
    \label{figs:Deng_UMAP_vis}
\end{figure}

\begin{figure}[t]
    \centering
    \includegraphics[width=1\linewidth]{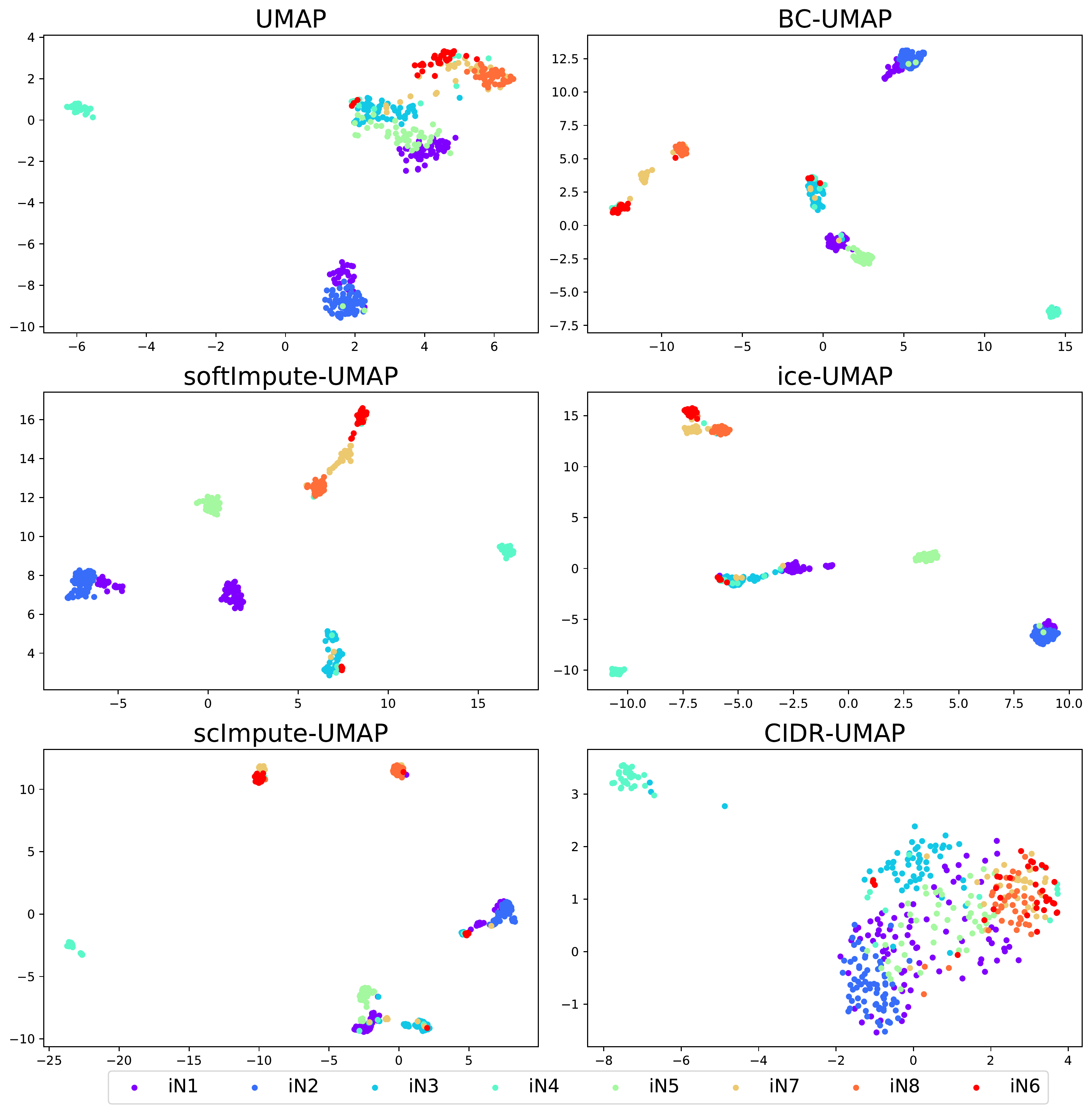}
    \caption{Visualisation of the Treutlein dataset obtained by UMAP and its variants integrated with the bias correction or imputations.}
    \label{figs:Treutlein_UMAP_vis}
\end{figure}

\begin{figure}[htbp]
    \centering
    \includegraphics[width=1\linewidth]{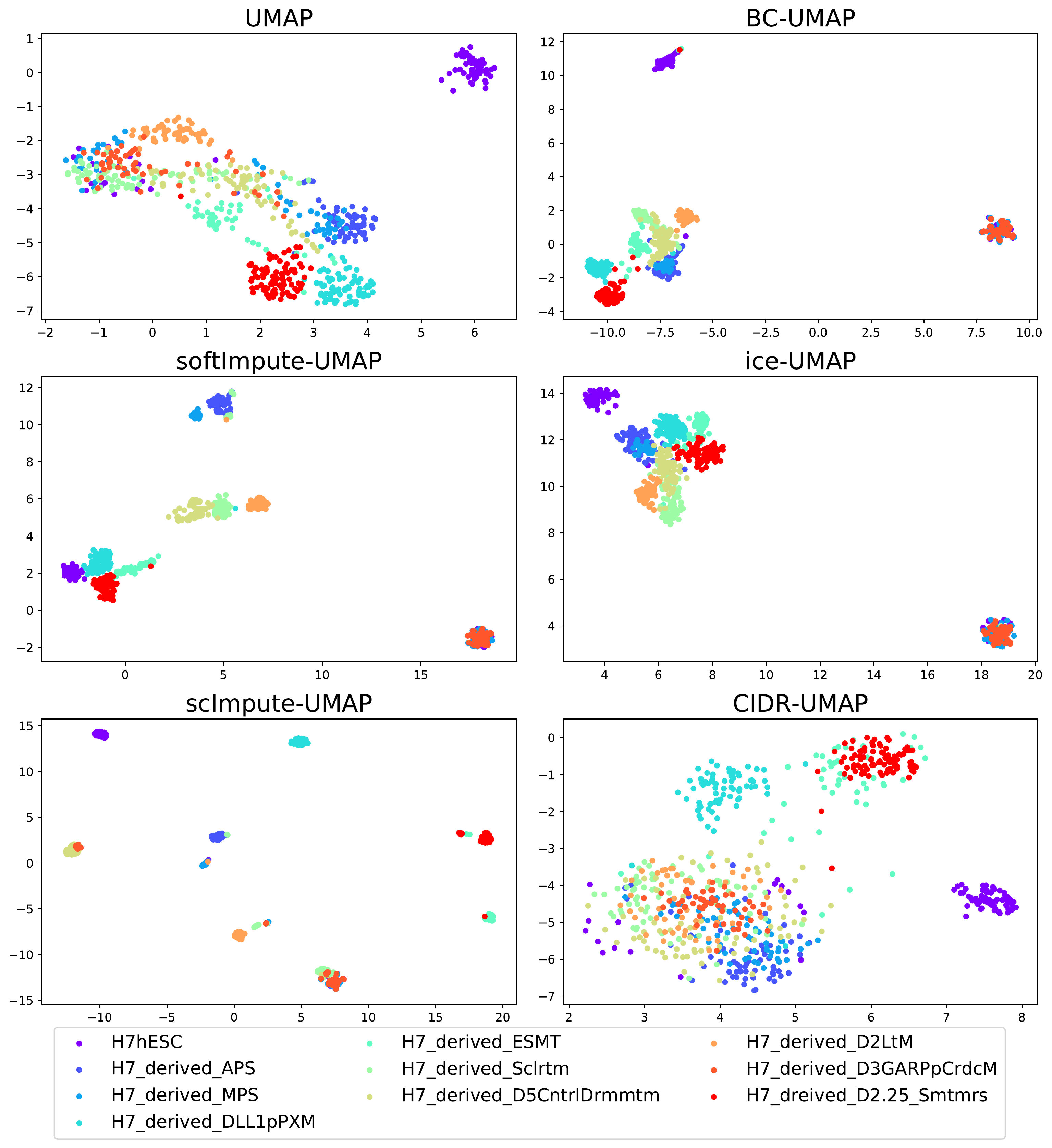}
    \caption{Visualisation of the Koh dataset obtained by UMAP and its variants integrated with the bias correction or imputations.}
    \label{figs:Koh_UMAP_vis}
\end{figure}

\begin{figure}[htbp]
    \centering
    \includegraphics[width=1\linewidth]{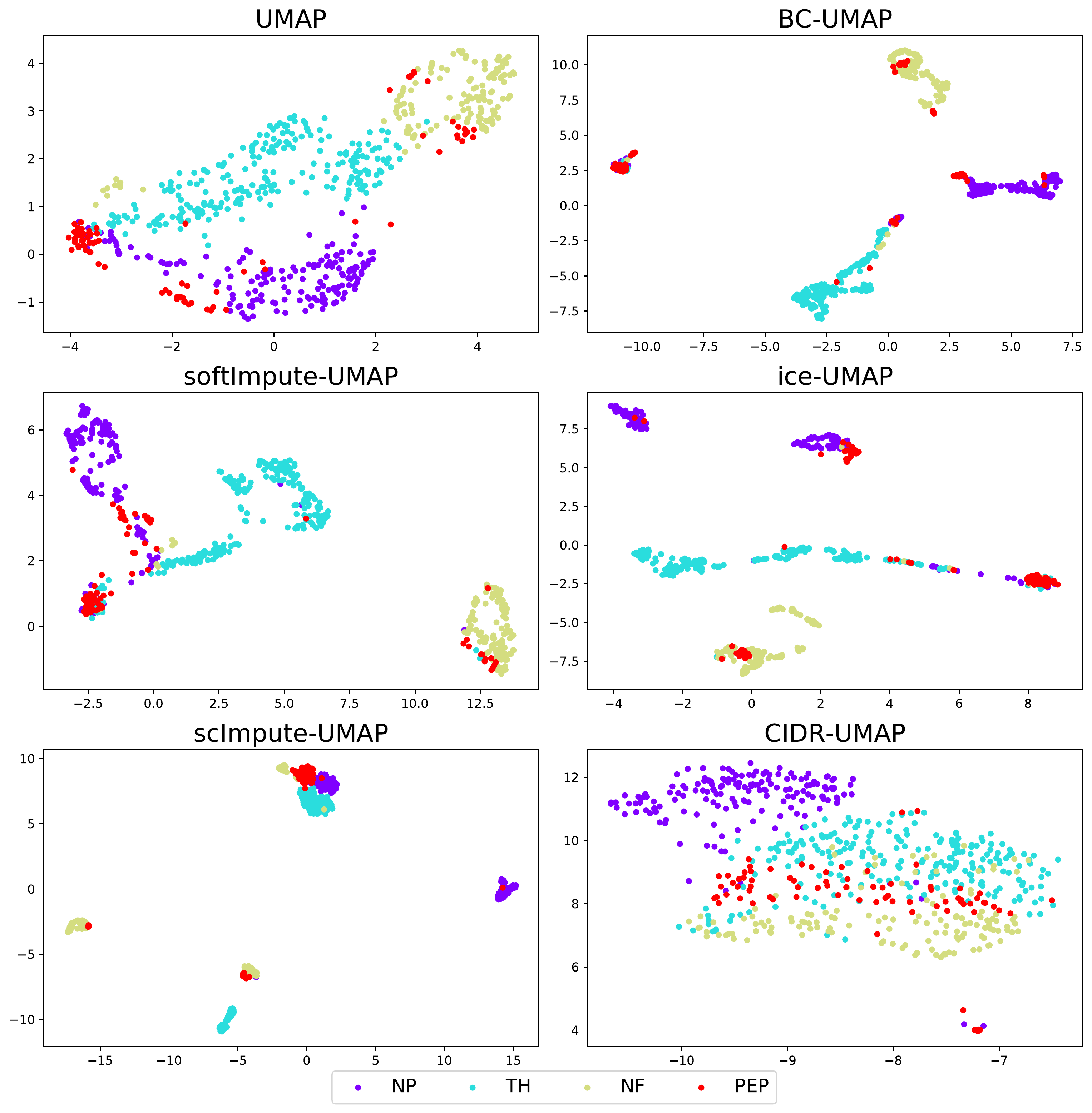}
    \caption{Visualisation of the Usoskin dataset obtained by UMAP and its variants integrated with the bias correction or imputations.}
    \label{figs:Usoskin_UMAP_vis}
\end{figure}

\clearpage
\subsubsection{K-means results on the real datasets}
\textbf{Results on the scRNA-seq datasets}
\begin{figure}[htbp]
    \centering
     \subfloat[]{\includegraphics[width=1\linewidth]{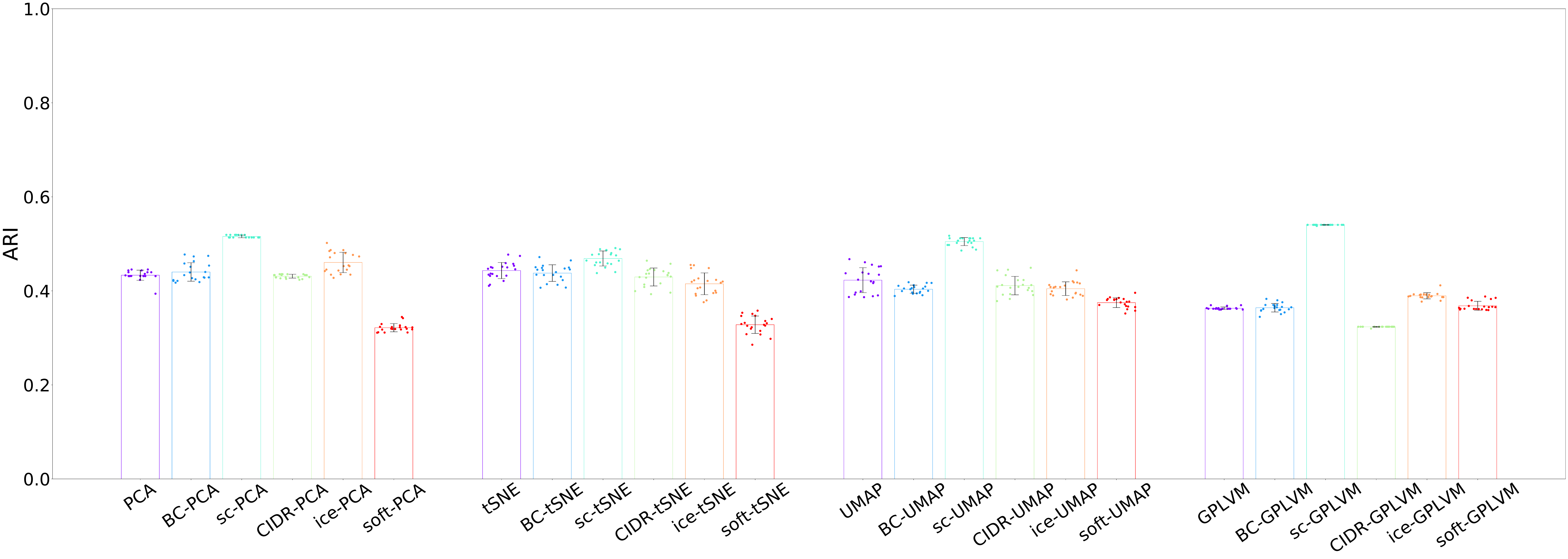}} \\
     \subfloat[]{\includegraphics[width=1\linewidth]{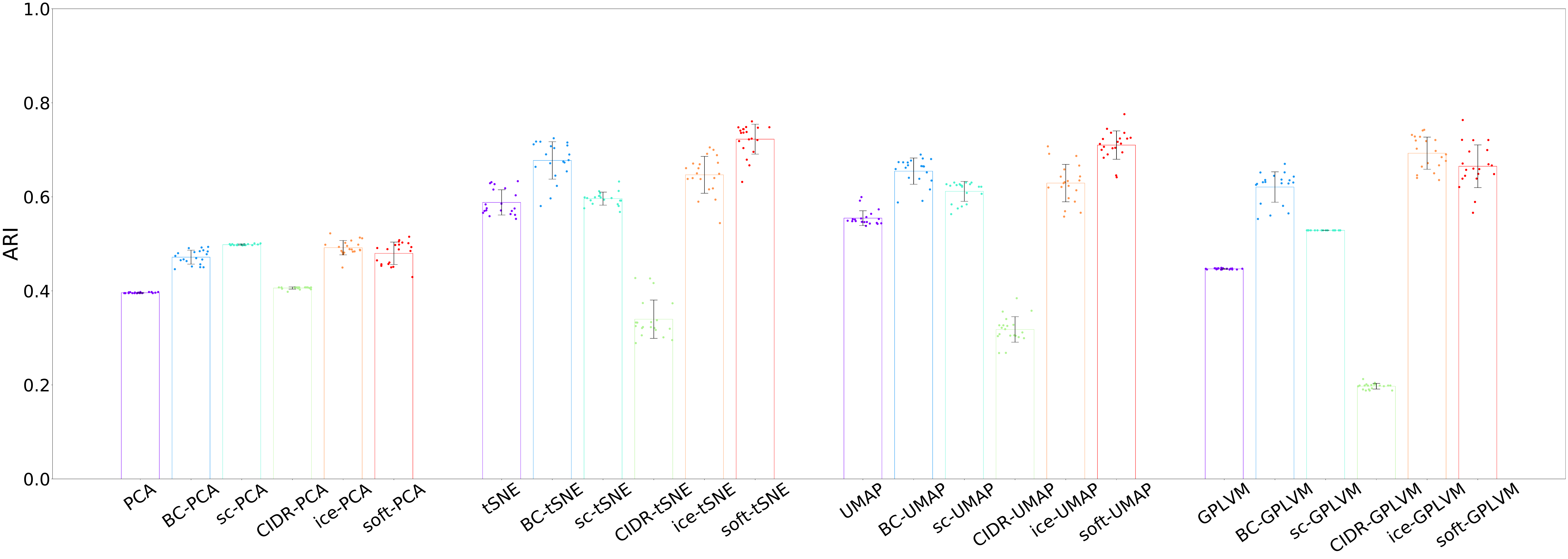}} \\
     \subfloat[]{\includegraphics[width=1\linewidth]{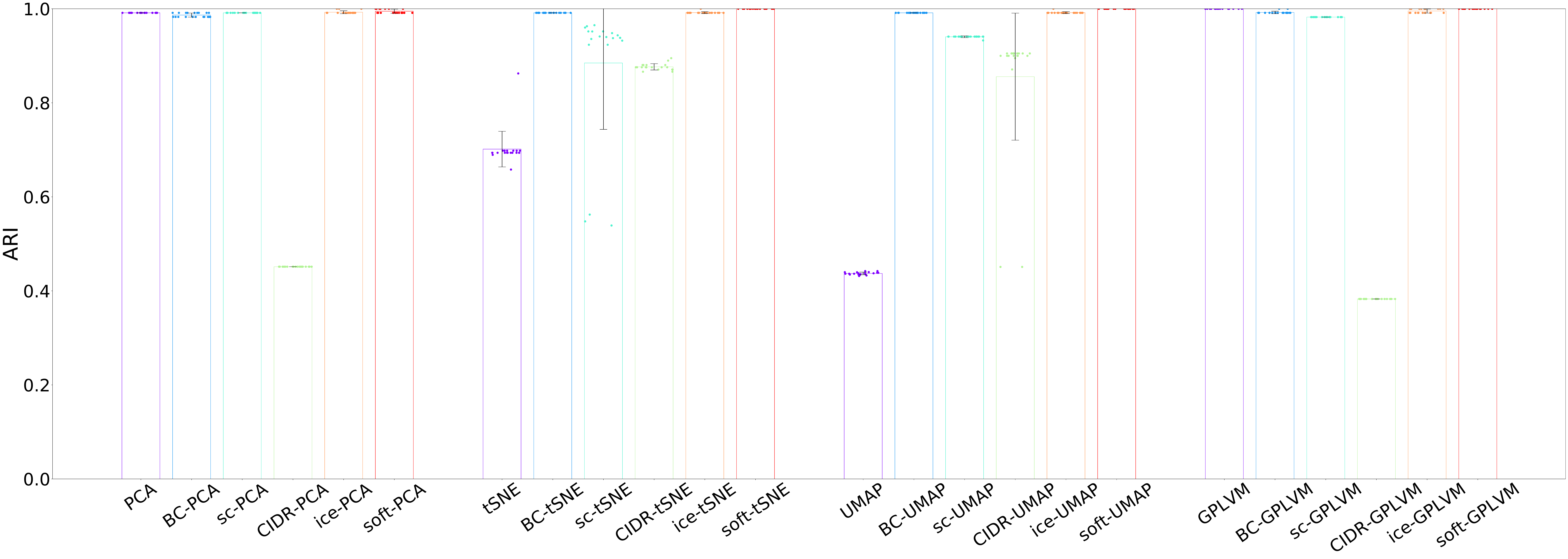}} \\
     \subfloat[]{\includegraphics[width=1\linewidth]{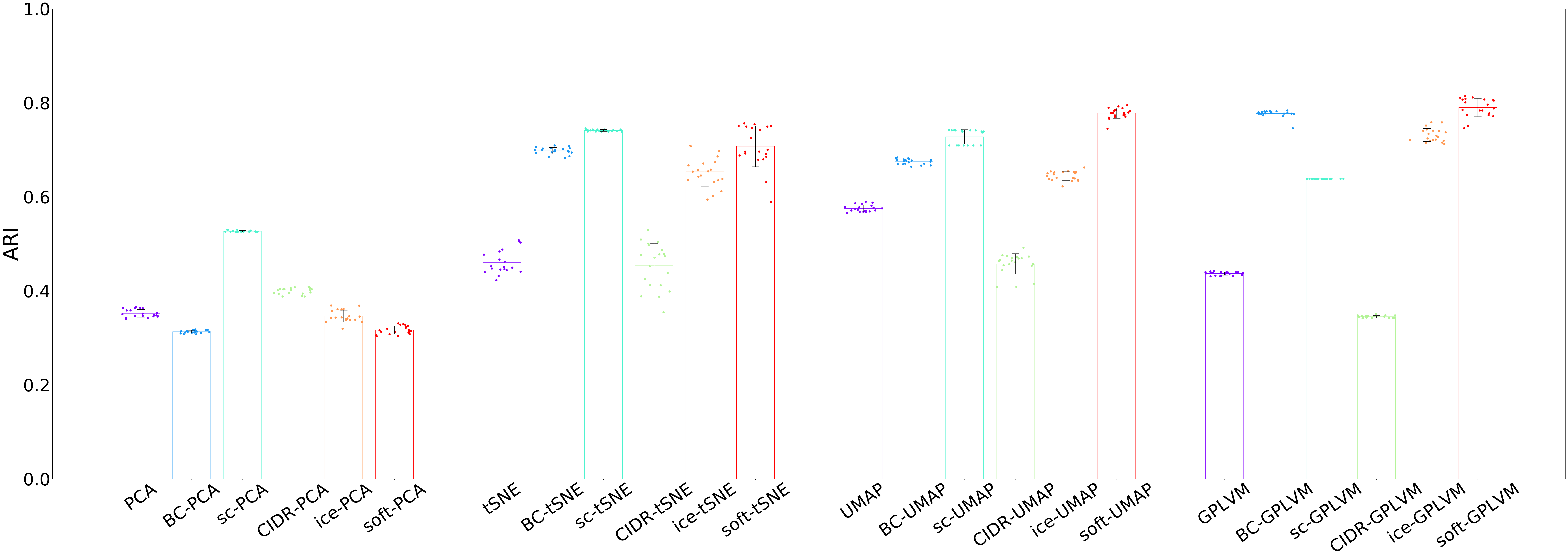}} \\
    \caption{ARI of $k$-means with different DR approaches and their variants on the scRNA-seq datasets (a) Deng, (b) Treutlein, (c) Kumar, and (d) Koh.}
    \label{fig:Kmeans_all_2}
\end{figure}

\newpage
\textbf{Results on the other datasets}
\begin{figure}[htbp]
    \centering
     \subfloat[]{\includegraphics[width=1\linewidth]{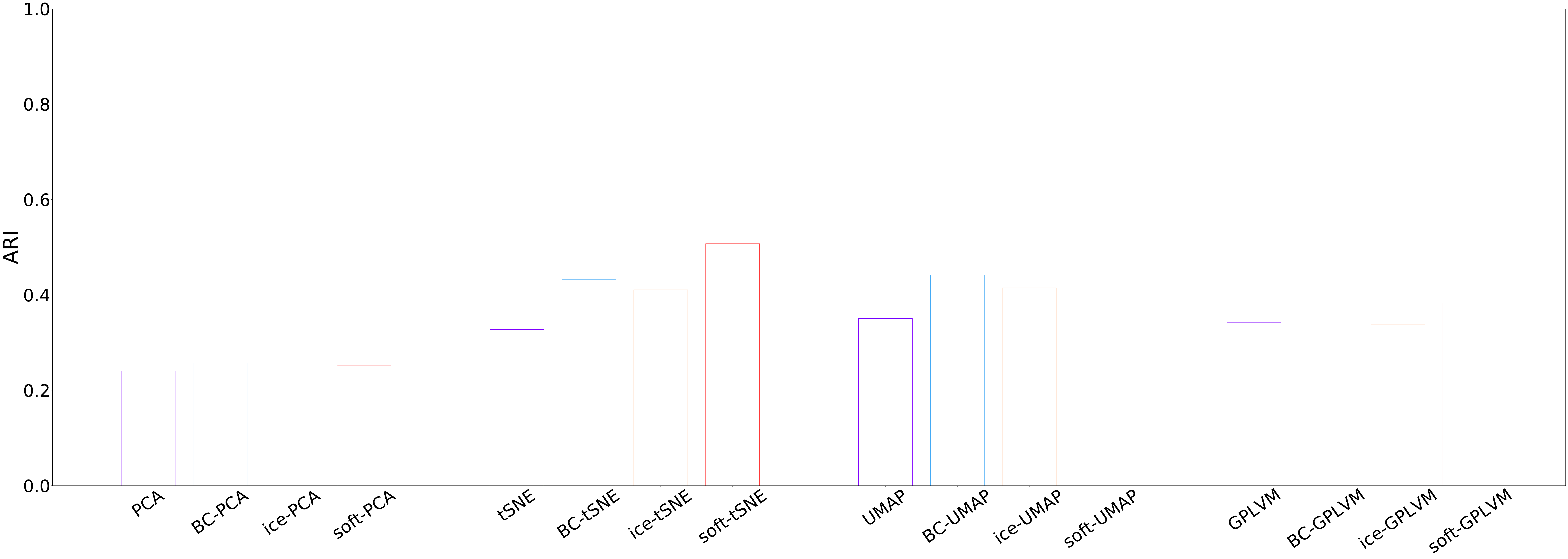}} \\
     \subfloat[]{\includegraphics[width=1\linewidth]{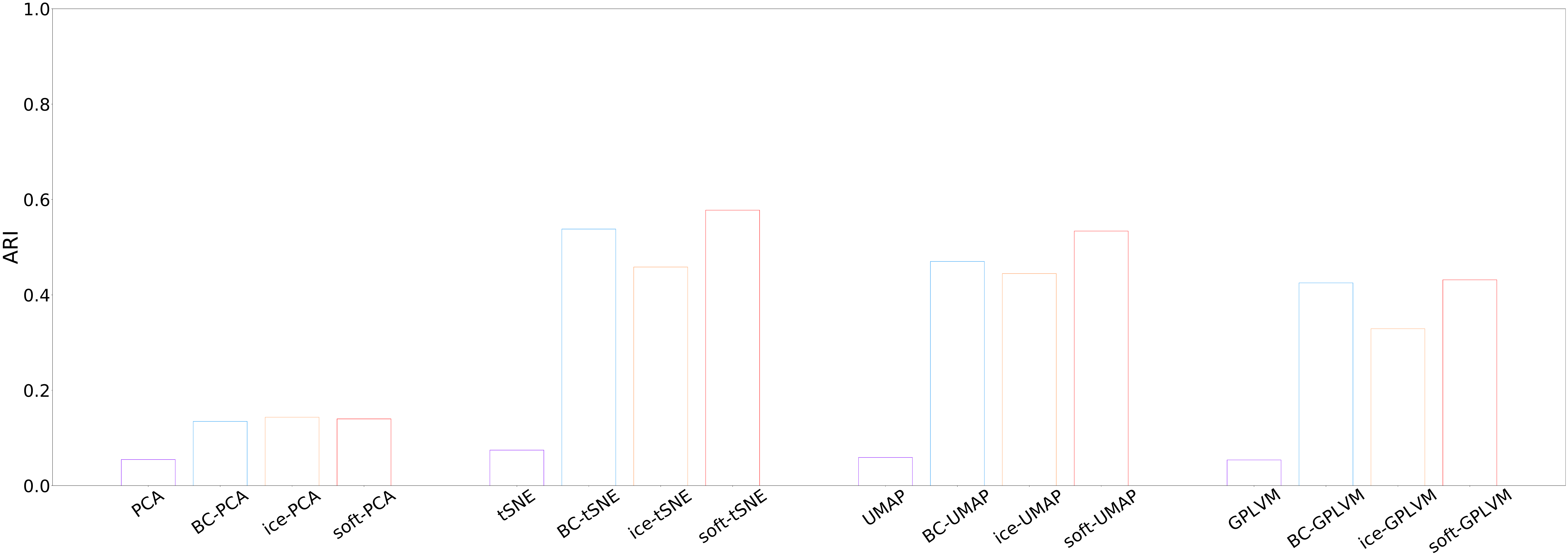}} \\
      \subfloat[]{\includegraphics[width=1\linewidth]{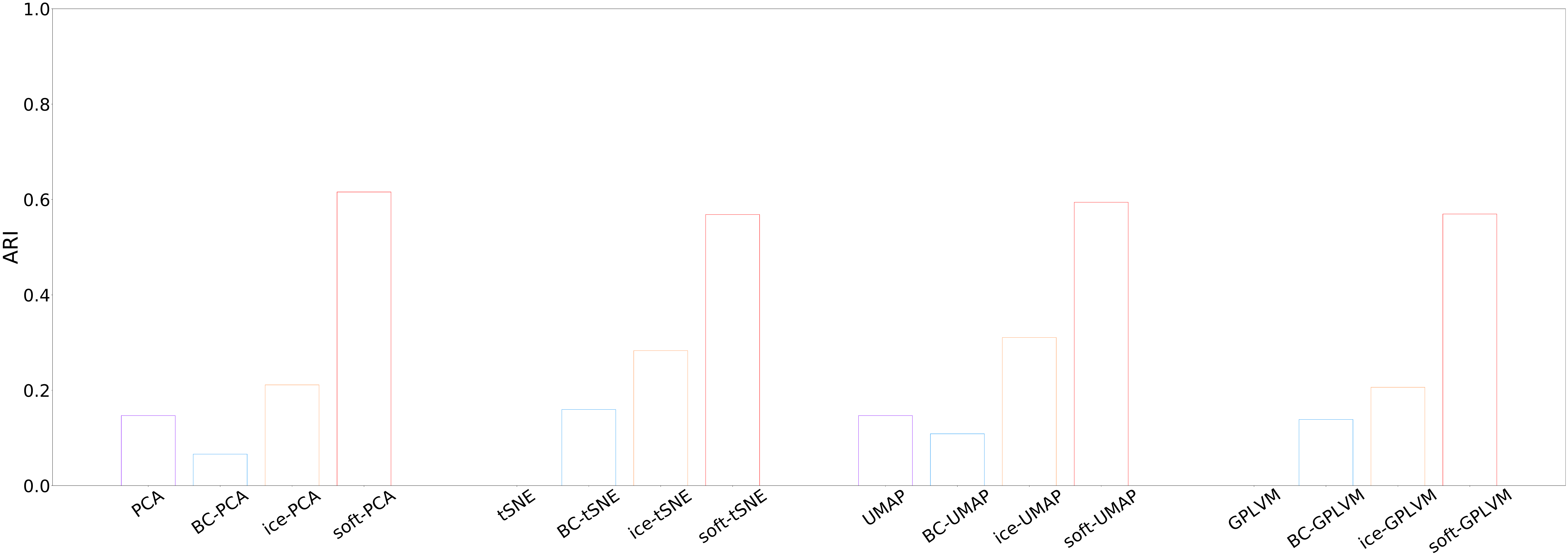}} 
    \caption{ARI of $k$-means with different DR approaches and their variants on the real datasets (a) fashion MNIST, (b) Olivetti faces, and (c) wine.}
    \label{fig:Kmeans_all_3}
\end{figure}

\end{document}